\documentclass[sn-apa, iicol]{sn-jnl}

\usepackage{graphicx}
\usepackage{multirow}%
\usepackage{amsmath,amssymb,amsfonts}%
\usepackage{amsthm}%
\usepackage{mathrsfs}%
\usepackage[title]{appendix}%
\usepackage{xcolor}%
\usepackage{textcomp}%
\usepackage{manyfoot}%
\usepackage{booktabs}%
\usepackage{algorithm}%
\usepackage{algorithmicx}%
\usepackage{algpseudocode}%
\usepackage{listings}%
\usepackage{enumitem}
\usepackage{colortbl}
\usepackage{overpic}
\usepackage{xspace}
\usepackage{capt-of}

\newcommand{\ourmethod}{NOPS\xspace}
\newcommand{\newmethod}{SNOPS\xspace}
\newcommand{\CC}[1]{\cellcolor{#1}}
\definecolor{novelcolor}{rgb}{0.8, 1., 0.9}

\usepackage[capitalize]{cleveref}
\crefname{section}{Sec.}{Secs.}
\Crefname{section}{Section}{Sections}
\Crefname{table}{Table}{Tables}
\crefname{table}{Tab.}{Tabs.}

\usepackage{todonotes}
\usepackage{soul}

\newcommand\blfootnote[1]{%
  \begingroup
  \renewcommand\thefootnote{}\footnote{#1}%
  \addtocounter{footnote}{-1}%
  \endgroup
}

\begin{document}

\title{Novel class discovery meets foundation models for 3D semantic segmentation}

\author*[1]{\fnm{Luigi} \sur{Riz}}\email{luriz@fbk.eu}

\author[2]{\fnm{Cristiano} \sur{Saltori}}

\author[1]{\fnm{Yiming} \sur{Wang}}

\author[1,2]{\fnm{Elisa} \sur{Ricci}}

\author[1]{\fnm{Fabio} \sur{Poiesi}}

\affil[1]{\orgname{Fondazione Bruno Kessler}, \orgaddress{\city{Trento}, \country{Italy}}}

\affil[2]{\orgname{University of Trento}, \orgaddress{\city{Trento}, \country{Italy}}}

\abstract{
The task of Novel Class Discovery (NCD) in semantic segmentation involves training a model to accurately segment unlabelled (novel) classes, using the supervision available from annotated (base) classes. 
The NCD task within the 3D point cloud domain is novel, and it is characterised by assumptions and challenges absent in its 2D counterpart. 
This paper advances the analysis of point cloud data in four directions. 
Firstly, it introduces the novel task of NCD for point cloud semantic segmentation. 
Secondly, it demonstrates that directly applying an existing NCD method for 2D image semantic segmentation to 3D data yields limited results. 
Thirdly, it presents a new NCD approach based on online clustering, uncertainty estimation, and semantic distillation. 
Lastly, it proposes a novel evaluation protocol to rigorously assess the performance of NCD in point cloud semantic segmentation. 
Through comprehensive evaluations on the SemanticKITTI, SemanticPOSS, and S3DIS datasets, our approach show superior performance compared to the considered baselines.
}


\maketitle

\blfootnote{Project page: \url{https://luigiriz.github.io/SNOPS_website/}.}
\blfootnote{This project has received funding from the European Union’s Horizon Europe research and innovation programme under the projects AI-PRISM (grant agreement No.~101058589) and FEROX (grant agreement No.~101070440).
This work was also partially sponsored by the PRIN project LEGO-AI (Prot.~2020TA3K9N), EU ISFP PRECRISIS (ISFP-2022-TFI-AG-PROTECT-02-101100539) , PNRR ICSC National Research Centre for HPC, Big Data and Quantum Computing (CN00000013) and the FAIR - Future AI Research (PE00000013), funded by NextGeneration EU. It was carried out in the Vision and Learning joint laboratory of FBK and UNITN.}

\begin{figure*}
    \centering
    \includegraphics[width=\textwidth]{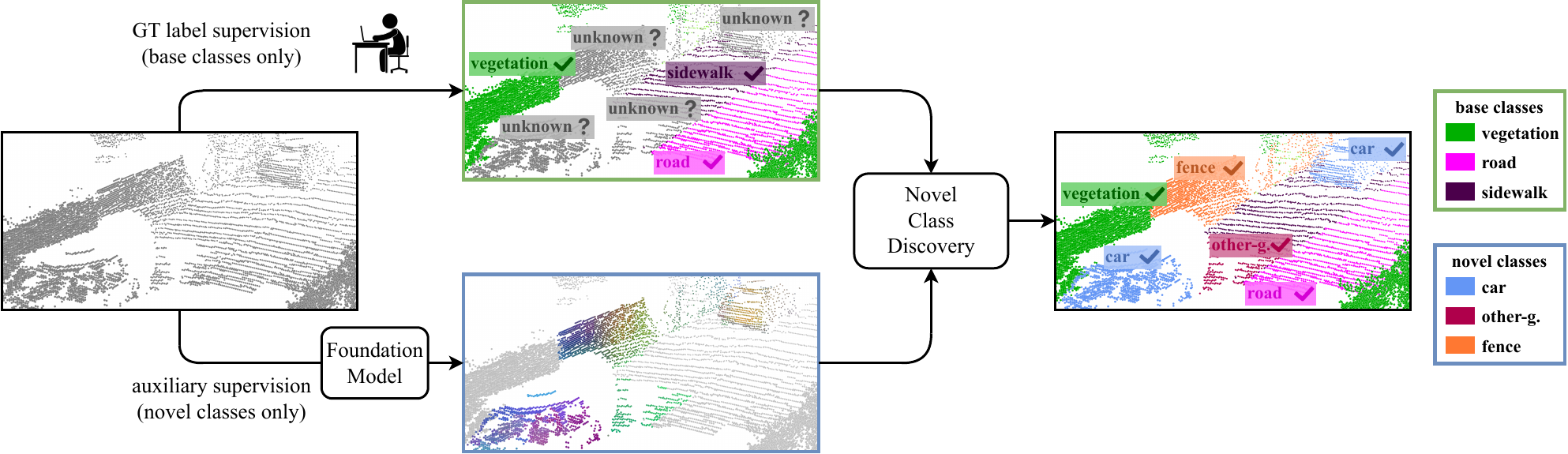}
    \vspace{-3mm}
    \caption{\newmethod addresses the novel class discovery task in 3D point cloud semantic segmentation by leveraging the knowledge of ground-truth labels (for base classes) and the auxiliary supervision from a foundation model (for novel classes) to learn the correct semantic segmentation of both base and novel points.}
    \label{fig:teaser}
\end{figure*}

\section{Introduction}\label{sec:intro}

Humans possess a remarkable ability to categorise new information (or novelties) into homogeneous groups, even when they are unfamiliar with what is observed. 
In contrast, machines can hardly achieve this without guidance. 
The primary challenges of machine vision lie in crafting discriminative latent representations of the real world and in quantifying uncertainty when faced with novelties \citep{han2019learning,zhong2021neighborhood,zhao2022novel}. 
\cite{han2019learning} pioneered the formulation of the Novel Class Discovery (NCD) problem. 
They defined it as the endeavor to categorise samples from an unlabelled dataset, termed \textit{novel samples}, into distinct classes by leveraging the insights from a set of labelled samples, known as the \textit{base samples}. 
Note that the classes in the labelled and unlabelled datasets are disjoint.

NCD has been explored in the 2D image domain for classification \citep{han2019learning,fini2021unified,zhong2021neighborhood}, and subsequently, for semantic segmentation \citep{zhao2022novel}. 
Specifically, \cite{zhao2022novel} introduced the first approach to address NCD in the 2D semantic segmentation task. 
The authors posited two key assumptions: first, each image contains only one novel class; and second, the novel class corresponds to a foreground object detectable through saliency detection (e.g., a man on a bicycle, with the bicycle being the novel class). 
Leveraging these assumptions, the authors were able to pool the features of each image into a single latent representation and group the representations of the entire dataset to identify clusters of novel classes. 
However, we argue that these assumptions impose significant constraints that are difficult to meet with generic 3D data, especially point clouds obtained from LiDAR sensors in large-scale settings. 
A single point cloud may contain multiple novel classes, and the concept of saliency in 3D data does not directly translate from its 2D counterpart. 
While both concepts relate to the focus of human attention, 3D saliency is more about the regional significance of 3D surfaces rather than a simple foreground/background distinction \citep{Ran2021}.

Our previous work, NOvel Point Segmentation (NOPS)~\citep{riz2023novel} pioneers in NCD for 3D semantic segmentation, with the primary focus on addressing the above discussed limitations. 
NOPS has shown promising performance in tackling 3D NCD, yet, the recent emergence of 3D foundation models \citep{peng2023openscene} offers us new opportunities in terms of the methodology design in NCD for their strong performance in zero-shot recognition.
However we empirically show that the accuracy of using the foundation model alone in a zero-shot manner for the NCD task is significantly lower than combining it with method that is specifically designed for NCD on multiple benchmark datasets~\citep{behley2019semantickitti,pan2020semanticposs,armeni20163d}, as shown in Tabs. \ref{tab:results_poss}-\ref{tab:results_s3dis}. 

In this work, we present Semantically-aligned Novel Point Segmentation (SNOPS), a method that extends NOPS~\citep{riz2023novel} by utilising an additional, unsupervised source of semantic knowledge in the form of a foundation model, such as CLIP~\citep{radford2021clip} (Fig.~\ref{fig:teaser}). 
SNOPS, given a dataset partially annotated by humans, concurrently learns base and novel semantic classes by clustering unlabelled points based on their semantic similarities. 
We have adapted the methodology of \cite{zhao2022novel}, termed Entropy-based Uncertainty Modelling and Self-training (EUMS), to accommodate point cloud data, thereby establishing it as our baseline. 
We move beyond their framework and, drawing inspiration from~\cite{caron2020unsupervised} and \cite{peng2023openscene}, incorporate batch-level (online) clustering and distillation from a foundation model.
Batch-level clustering generates prototypes that we utilise to manage large-cardinality 3D point clouds, while distillation is essential to leverage the intrinsic semantic knowledge contained within foundation models. 
We update prototypes during training to make clustering computationally feasible and introduce a method based on uncertainty to enhance prototype quality. 
We establish point-cluster assignment to produce pseudo-labels for self-training and also employ over-clustering to ensure precision. 
Given the diverse semantic classes within point clouds, it is inevitable that not all classes are represented in every batch. 
To address this issue, we have developed a queuing approach to maintain representative features throughout the training process. 
These features act as proxies for missing categories during the generation of pseudo-labels, facilitating a more balanced clustering of the novel classes.
Lastly, we generate two augmented perspectives of a singular point cloud and enforce consistency in pseudo-labels between them.
Our methodology is assessed on SemanticKITTI \citep{behley2019semantickitti, geiger2012cvpr, behley2021ijrr}, SemanticPOSS \citep{pan2020semanticposs}, S3DIS~\citep{armeni20163d}, and nuScenes\cite{caesar2020nuscenes}.
We establish an evaluation protocol for NCD and point cloud segmentation, serving as a potential benchmark for subsequent research. 
Empirical evidence suggests that our method significantly surpasses our baseline and predecessor version of our method \citep{riz2023novel} across all datasets.
Additionally, we undertake a comprehensive ablation study to underscore the significance of our method's diverse components.

\vspace{1mm}
\noindent To summarise, our contributions are:
\setlist{nolistsep}
\begin{itemize}
    \item Tackling NCD for 3D semantic segmentation, addressing unfit assumptions that are originally imposed on NCD for 2D semantic segmentation;
    \item Providing empirical proof that zero-shot semantic segmentation with 3D foundation model is not a good enough solution for NCD.
    \item Presenting a novel method \newmethod that effectively synergises NCD method with semantic distillation through foundation model, advancing the state of the art in NCD for 3D semantic segmentation;
    \item Introducing a new evaluation protocol to assess the performance of NCD for 3D semantic segmentation.
\end{itemize}

This paper extends our earlier work \citep{riz2023novel} in several aspects.
We extend the original NOPS by leveraging a foundation model \citep{peng2023openscene} to improve the accuracy of novel classes.
We empirically show that the zero-shot accuracy of the foundation model alone is significantly lower than that achieved by using it in combination of our novel class discovery method.
Then, we significantly extend our experimental evaluation and analysis by adding new experiments, new datasets, new comparisons, and new ablation studies to evaluate this new setup.
Lastly, we expand the related work by reviewing additional state-of-the-art approaches.

\section{Related work}
In this section, we thoroughly discuss recent works on three relevant topics, including point cloud semantic segmentation, 3D representation learning and novel class discovery. 
\nocite{Giugliari2022}

\noindent \textbf{Point cloud semantic segmentation} can be performed at point level~\citep{qi2017pointnet++}, on range view maps~\citep{ronneberger2015u}, or by voxelising the input points~\citep{zhou2018voxelnet}. 
Point-level networks process the input without intermediate representations. Examples of these include PointNet~\citep{qi2017pointnet}, PointNet++~\citep{qi2017pointnet++}, RandLA-Net~\citep{hu2020randla}, and KPConv~\citep{thomas2019kpconv}.
PointNet~\citep{qi2017pointnet} and PointNet++~\citep{qi2017pointnet++} are based on a series of multi-layer perceptron where PointNet++ introduces global and local feature aggregation at multiple scales. 
RandLA-Net~\citep{hu2020randla} uses random sampling, attentive pooling, and local spatial encoding. 
KPConv~\citep{thomas2019kpconv} employs flexible and deformable convolutions in a continuous input space. 
Point-level networks are computationally inefficient when large-scale point clouds are processed. 
Range view architectures~\citep{milioto2019rangenet++} and voxel-based approaches~\citep{choy20194d} are more computationally efficient than their point-level counterpart. 
The former requires projecting the input points on a 2D dense map, processing input maps with 2D convolutional filters~\citep{ronneberger2015u}, and re-projecting predictions to the initial 3D space. 
SqueezeSeg networks~\citep{wu2018squeezeseg, wu2019squeezesegv2}, 3D-MiniNet~\citep{alonso2020MiniNet3D}, RangeNet++~\citep{milioto2019rangenet++}, and PolarNet~\citep{zhang2020polarnet} are examples of this category. 
Although they are more efficient, these approaches tend to lose information during projection and re-projection.
The latter includes 3D quantisation-based approaches that discretise the input points into a 3D voxel grid and employ 3D convolutions~\citep{zhou2018voxelnet} or 3D sparse convolutions~\citep{SubmanifoldSparseConvNet, choy20194d} to predict per-voxel classes. VoxelNet~\citep{zhou2018voxelnet}, SparseConv~\citep{SubmanifoldSparseConvNet, 3DSemanticSegmentationWithSubmanifoldSparseConvNet}, MinkowskiNet~\citep{choy20194d}, Cylinder3D~\citep{zhu2021cylindrical}, and (AF)$^2$-S3Net~\citep{ran2021af2s3net} are architectures belonging to this category. 
The above-mentioned approaches usually tackle point cloud segmentation in a supervised setting, whereas we address novel class discovery with both labelled base classes and unlabelled novel classes.

\noindent \textbf{3D representation learning} refers to learn general and useful point cloud representations from unlabelled point cloud data~\citep{achlioptas2018learning, xiao2023tpami}.
Existing methods can be grouped into generative, context similarity based, local descriptor based, and multi-modal approaches.
Generative approaches involve the generation of a point cloud as unsupervised task~\citep{yang2018foldingnet, yang2021progressive, yang2019pointflow}. FoldingNet~\citep{yang2018foldingnet}, PSG-Net~\citep{yang2021progressive} and PointFlow~\citep{yang2019pointflow} follow the autoencoder~\citep{hinton2006reducing} paradigm and learn to self-reconstruct the input point cloud.
Differently, LatentGAN~\citep{achlioptas2018learning}, Tree-GAN~\citep{shu20193d} and 3D-GAN~\citep{wu2016learning} follow a generative adversarial strategy and learn to generate point cloud instances from a sampled vector or a latent embedding.
PU-GAN~\citep{li2019pu} and PU-GCN~\citep{qian2021pu} learn the underlying geometries of point clouds by generating a denser point cloud with similar geometries.
On the other hand, PCN~\citep{yuan2018pcn}, SA-Net~\citep{wen2020point}, Point-BERT~\citep{yu2022point} and Point-MAE~\citep{pang2022masked} learn to complete the input point cloud by predicting the arbitrary missing parts.
Context similarity based approaches learn discriminative 3D representations through the underlying similarities between point samples.
PointContrast~\citep{xie2020pointcontrast}, DepthContrast~\citep{zhang2021self}, ACD~\citep{gadelha2020label} and STRL~\citep{huang2021spatio} enforce the network to group feature representations through contrastive learning between positive and negative point cloud pairs.
Another similarity based technique makes use of coordinate sorting as a unsupervised task. For example, Jigsaw3D~\citep{sauder2019self} and Rotation3D~\citep{poursaeed2020self} follow this idea and train the network to predict either the re-organised version or the rotation angle of the input point clouds. Local descriptor approaches focus on learning to encode per-point informative features by solving low-level tasks, \textit{e.g.} point cloud registration. PPF-FoldNet~\citep{deng2018ppf} and CEM~\citep{jiang2021sampling} learn compact descriptors by solving the task of point cloud matching and registration, respectively. Differently, GeDi~\citep{poiesi2022learning} employs constrastive learning between canonical point cloud patches to learn compact and generalisable descriptors.
Multi-modal approaches follow the recent success from the 2D literature~\citep{radford2021learning, dong2023maskclip} and learn robust and comprehensive representations by modeling the relationships across modalities.
Language grounding~\citep{rozenberszki2022language} maps per-point features to text CLIP~\citep{radford2021learning} embeddings, providing a robust pre-training for semantic tasks.
ConceptFusion~\citep{jatavallabhula2023conceptfusion} leverages the open-set capabilities of foundation models~\citep{guzhov2022audioclip, radford2021clip, kirillov2023segment} from multiple modalities and fuses their features into a 3D map via traditional integration approaches.
More recently, OpenScene~\citep{peng2023openscene} learns a feature space where text and multi-view image pixels are co-embedded in the CLIP feature space.
In this work, we tackle NCD for 3D segmentation and extend our previous method \ourmethod~\citep{riz2023novel} by leveraging the powerful representations of OpenScene~\citep{peng2023openscene} in our novel class discovery network. 

\noindent \textbf{Novel class discovery} (NCD) is initially explored for 2D classification~\citep{han2019learning, zhong2021neighborhood, fini2021unified, joseph2022novel, roy2022class, jia2021joint, zhong2021openmix, vaze2022generalized, yang2022divide} and 2D segmentation~\citep{zhao2022novel}. 
NCD is formulated in a different way compared to standard semi-supervised learning~\citep{souly2017semi, zhang2020wcp, tang2016large}. In semi-supervised learning, labelled and unlabelled samples belong to the same classes, while in NCD, novel and base samples belong to disjoint classes. 
Han et al.~\citep{han2019learning} pioneered the NCD problem for 2D image classification. 
A classification model is pre-trained on a set of base classes and used as feature extractor for the novel classes. 
They then train a classifier for the novel classes using the pseudo-labels produced by the pre-trained model. 
Zhong et al.~\citep{zhong2021neighborhood} introduced neighbourhood contrastive learning to generate discriminative representations for clustering. 
They retrieve and aggregate pseudo-positive pairs with contrastive learning, encouraging the model to learn more discriminative representations. 
Hard negatives are obtained by mixing labelled and unlabelled samples in the feature space.
UNO~\citep{fini2021unified} unifies the two previous works by using a unique classification loss function for both base and novel classes, where pseudo-labels are processed together with ground-truth labels. 
NCD without Forgetting~\citep{joseph2022novel} and FRoST~\citep{roy2022class} further extend NCD to the incremental learning setting. 
EUMS~\citep{zhao2022novel} is the only approach analysing NCD for 2D semantic segmentation. Unlike image classification, the model has to classify each pixel and handle multiple classes in each image. EUMS consists of a multi-stage pipeline using a saliency model to cluster the latent representations of novel classes to produce pseudo-labels. Moreover, entropy-based uncertainty and self-training are used to overcome noisy pseudo-labels while improving the model performance on the novel classes.

In this work, we focus on NCD for 3D point cloud semantic segmentation. Unlike previous works, our problem inherits the challenges from the fields of 2D semantic segmentation~\citep{deeplabv3plus2018, chen2017deeplab} and 3D point cloud segmentation~\citep{choy20194d, saltori2022cosmix, milioto2019rangenet++}. From 2D semantic segmentation, the main challenges are multiple novel classes in the same image and the strong class unbalance. From 3D point cloud segmentation, we have to tackle the sparsity of input data, the different density of point cloud regions and the inability to identify foreground and background, which are not present in 2D segmentation~\citep{zhao2022novel}.
From related fields in 3D scene understanding, the previous effort REAL~\citep{cen2022open} tackles open-world 3D semantic segmentation by classifying all the unknown points into a single class. Novel classes are then labelled by a human annotator and used for learning incrementally novel classes.
Instead, NOPS~\citep{riz2023novel} is the first work tackling NCD for 3D semantic segmentation.
Unlike \citep{zhao2022novel} that uses K-Means, \cite{riz2023novel} formulate clustering as an optimal transport problem to avoid degenerate solutions, i.e.~all data points may be assigned to the same label and learn a constant representation~\citep{Asano2020, mei2022data}. On top of NOPS~\citep{riz2023novel}, this work incorporates the unsupervised semantic knowledge distilled from a 3D foundation model~\citep{peng2023openscene}. We show that the unsupervised knowledge distilled from a 3D foundation model significantly improves NCD performance.


\begin{figure*}[ht]
    \centering
    \includegraphics[width=0.95\textwidth]{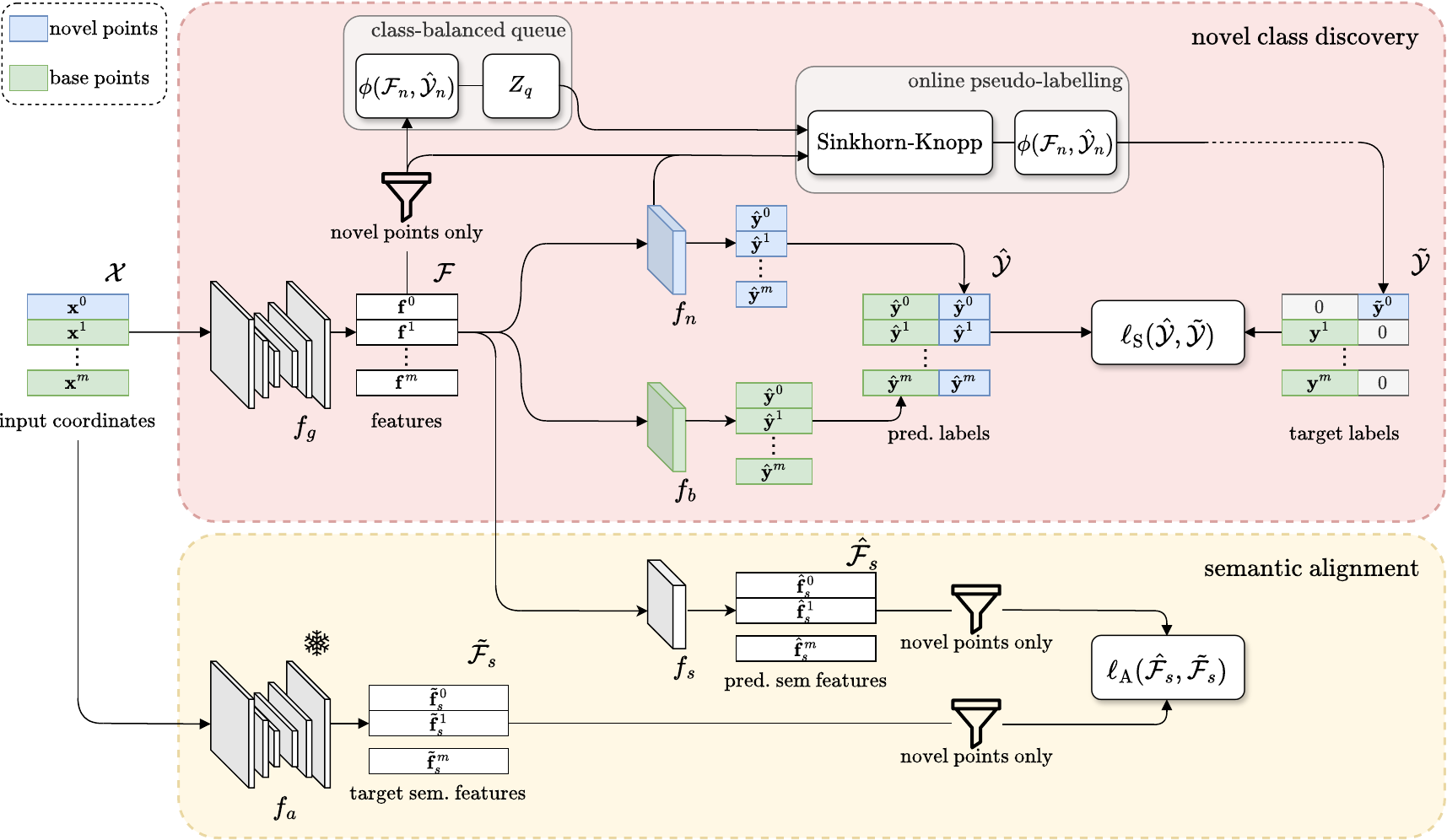}
    \caption{Overview of \newmethod.
    We extract point-level features $\mathcal{F}$ with the shared backbone $f_g$.
    $\mathcal{F}$ are used to obtain pseudo-labels in the online pseudo-labelling block.
    We forward $\mathcal{F}$ through a novel $f_n$ and a base $f_b$ segmentation head to obtain point-wise predictions.
    We also pass $\mathcal{F}$ through a projection layer $f_s$ that produces point-wise features for novel points. We align such point descriptors to the ones output by a frozen auxiliary network $f_a$, a large 3D vision model.
    The network is optimised by minimising the sum of a segmentation loss and an alignment loss.}
    \label{fig:main_chart}
\end{figure*}

\section{Our approach}\label{sec:approach}

\subsection{Overview}

We use two UNet-like deep neural networks optimised for 3D data to extract point-level features from an input point cloud. 
The primary network starts untrained, serving as our target for training to concurrently segment both base and novel classes. 
The secondary network is auxiliary and pre-trained for task-agnostic open-vocabulary 3D scene understanding.
For base class points, we use traditional supervised training, leveraging the available human annotations (ground truth). 
The training for novel classes pursues two distinct objectives.
Firstly, we aim to align the features with the semantic knowledge of the auxiliary network (Sec.~\ref{sec:semantic_distillation}). 
Secondly, we adopt a self-supervised approach that generates pseudo-labels based on our \textit{online pseudo-labelling} method through the Sinkhorn-Knopp algorithm~\citep{cuturi2013sinkhorn} (Sec.~\ref{sec:pseudo_labelling}).
To enable each processed batch to maintain an equal number of novel classes, even if some are absent in point clouds being processed, we use a \textit{class-balanced queue} that stores features during training (Sec.~\ref{sec:class_balanced_queue}). 
We harness the pseudo-label confidences (class probabilities) to sift out uncertain points, thus populating the queue solely with high-quality points (Sec.~\ref{sec:unc_train}).
Specifically, our optimisation objective is
\begin{equation}
    \mathcal{L} = \ell_\text{S} + \gamma \ell_\text{A},
\end{equation}
where $\ell_\text{S}$ is the segmentation loss involving ground-truth labels and pseudo-labels (Sec.~\ref{sec:pseudo_labelling}), $\ell_\text{A}$ is the alignment loss that considers the semantic features extracted with the auxiliary network (Sec.~\ref{sec:semantic_distillation}) and $\gamma$ is a weighting factor.
Fig.~\ref{fig:main_chart} shows the block diagram of \newmethod.

\subsection{Problem formulation}\label{sec:problem_formulation}

Let $\mathrm{X} = \{\mathcal{X}\}$ be a dataset of 3D point clouds captured in different scenes.
The point cloud $\mathcal{X}$ is a set composed of a base set $\mathcal{X}_b$ and a novel set $\mathcal{X}_n$, s.t.~$\mathcal{X} = \mathcal{X}_b \cup \mathcal{X}_n$.
The semantic categories that can be present in our point clouds are $\mathcal{C} = \mathcal{C}_b \cup \mathcal{C}_n$, where $\mathcal{C}_b$ is the set of base classes and $\mathcal{C}_n$ is the set of novel classes, s.t.~$\mathcal{C}_b \cap \mathcal{C}_n = \emptyset$.
Each $\mathcal{X} \in \mathrm{X}$ is composed of a finite but unknown number of 3D points $\mathcal{X} = \{(\mathbf{x}, c)\}$, where $\mathbf{x} \in \mathbb{R}^3$ is the coordinate of the a point and $c$ is its semantic class.
We know the class of the point $(\mathbf{x}, c)$, s.t.~$\mathbf{x} \in \mathcal{X}_b$ and $c \in \mathcal{C}_b$, but we do not know the class of the point $(\mathbf{x}, c)$, s.t.~$x \in \mathcal{X}_n$ and $c \in \mathcal{C}_n$.
No points in $\mathcal{X}_n$ belong to one of the base classes $\mathcal{C}_b$.
As in \citep{han2019learning, zhong2021neighborhood, zhao2022novel}, we assume that the number of classes to discover is known, i.e.~$|\mathcal{C}_n| = C_n$.
We aim to train a deep neural network $f_{\mathbf{\Theta}}$ that can segment all the points of a given point cloud, thus learning to jointly segment base classes $\mathcal{C}_b$ and novel classes $\mathcal{C}_n$.
$\mathbf{\Theta}$ are the weights of our deep neural network.
$f_\mathbf{\Theta}$ is composed of a feature extractor network $f_g$, two segmentation heads $f_n$ and $f_b$ (for novel and base classes, respectively) and a feature projector $f_s$. $f_\mathbf{\Theta} = f_g \circ \{f_b, f_n, f_s\}$, where $\circ$ is the composition operator (Fig.~\ref{fig:main_chart}).

\subsection{Online pseudo-labelling}\label{sec:pseudo_labelling}

We formulate pseudo-labelling as the assignment of novel points to the class-prototypes learnt during training~\citep{caron2020unsupervised}.
Let $\mathtt{P} \in \mathbb{R}^{D \times \rho}$ be the class prototypes, where $D$ is the size of the output features from $f_g$ and $\rho$ is the number of prototypes.
Let $\mathtt{Z} \in \mathbb{R}^{D \times m_n}$ be the normalised output features for novel points extracted from $f_g$, i.e.~$\mathtt{Z} = f_g(\mathcal{X}_n)$, where $m_n$ is the number of novel points of the point cloud. 
$m_n$ is unknown a priori and it can differ across point clouds.
We define $\mathtt{Q} \in \mathbb{R}^{\rho \times m_n}$ as the assignment between the $\rho$ prototypes and the $m_n$ novel points that equally partitions the novel points in the point cloud across the available prototypes.
This equipartition ensures that the feature representations of the points belonging to different novel classes are well separated, thus preventing the case in which the novel class feature representations collapse into a unique solution.
\cite{caron2020unsupervised} employs an arbitrary large number of prototypes $\rho$ to effectively organise the feature space produced by $f_g$. 
They discard $\mathtt{P}$ after training. 
In contrast, we learn exactly $\rho = C_n$ class prototypes and use $\mathtt{P}$ as the weights for our new class segmentation head $f_n$, which outputs the $C_n$ logits for the new classes. 
In order to optimise the assignment $\mathtt{Q}$, we maximise the similarity between the features of the new points and the learnt prototypes as
\begin{equation}
    \label{eq:sk_problem}
    \max_{\mathtt{Q} \in \mathcal{Q}} \,\, \text{Tr}(\mathtt{Q}^\top \mathtt{P}^\top \mathtt{Z}) + \epsilon H(\mathtt{Q})  \rightarrow \mathtt{Q}^*,
\end{equation}
where $H$ is the entropy function, $\epsilon$ is the parameter that determines the smoothness of the assignment and $\mathtt{Q}^*$ is our sought solution. 
\cite{Asano2020} enforce the equipartioning constraint by requiring $\mathtt{Q}$ to belong to a transportation polytope and perform this optimisation on the whole dataset at once (offline).
This operation with point cloud data is computationally impractical.
Therefore, we formulate the transportation polytope such that the optimisation is performed online, which consist of considering only the points within the point cloud being processed
\begin{equation}
\resizebox{\linewidth}{!}{$
    \mathcal{Q} = \left\{ \mathtt{Q} \in \mathbb{R}^{C_n \times m_n}_+ | \mathtt{Q} \mathtt{1}_{m_n} = \frac{1}{C_n} \mathtt{1}_{C_n}, \mathtt{Q}^\top \mathtt{1}_{C_n} = \frac{1}{m_n} \mathtt{1}_{m_n} \right\},
$}
\end{equation}
where $\mathtt{1}_\star$ represents a vector of ones of dimension $\star$.
These constraints ensure that each class prototype is selected on average at least $m_n / C_n$ times in each point cloud. 
The solution $\mathtt{Q}^*$ can take the form of a normalised exponential matrix
\begin{equation}
    \mathtt{Q}^* = \text{diag}(\alpha) \exp \left( {\frac{\mathtt{P}^\top \mathtt{Z}}{\epsilon}} \right) \text{diag}(\beta),
\end{equation}
where $\alpha$ and $\beta$ are renormalisation vectors that are computed iteratively with the Sinkhorn-Knopp algorithm \citep{cuturi2013sinkhorn, mei2023overlap}.
We then transpose the optimised soft assignment $\mathtt{Q}^* \in \mathbb{R}^{C_n \times m_n}_+$ to obtain the soft pseudo-labels for each of the $m_n$ novel points being processed within each point cloud.
For simplicity, the procedure described here takes into account only batches composed of a single point cloud. However, the same algorithm can be applied when two or more point clouds are concatenated into a single batch.

We empirically found that training can be more effective if pseudo-labels are smoother in the first training epochs and peaked in the last training epochs.
Therefore, we introduce a linear decay of $\epsilon$ during training.

\begin{figure}[t]
    \centering
    \includegraphics[width=0.9\linewidth]{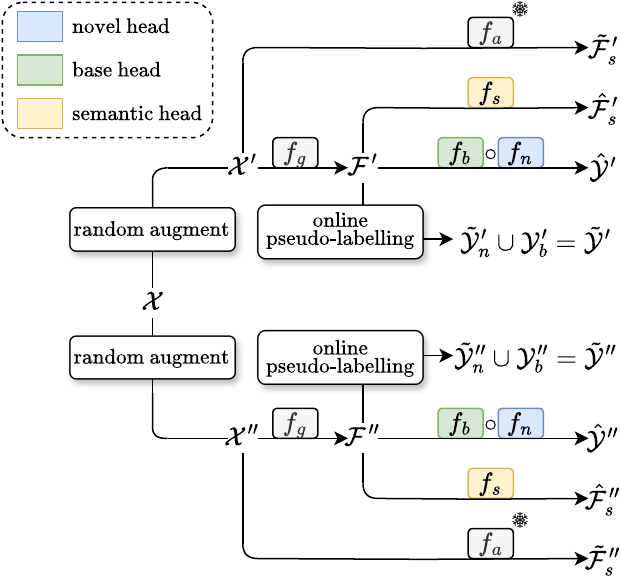}
    \caption{Overview of the different outputs after the input point cloud $\mathcal{X}$ undergoes two different random augmentations, required for the generation of self-supervised pseudo-labels.}
    \label{fig:augmentations}
\end{figure}

\noindent \textbf{Segmentation objective:}
The segmentation objective $\ell_\text{S}$ is formulated as the weighted Cross Entropy loss and it is based on the ground-truth labels $\mathcal{Y}_b$ for base points and on the pseudo-labels $\tilde{\mathcal{Y}}_n$ for novel points.
We formulate a swapped prediction task based on these pseudo-labels \citep{caron2020unsupervised}.
We begin by generating two different augmentations of the original point cloud $\mathcal{X}$ that we define as $\mathcal{X}^\prime$ and $\mathcal{X}^{\prime\prime}$ (Fig.~\ref{fig:augmentations}).
For the augmentation $\mathcal{X}^\prime$ we define the segmentation predictions $\hat{\mathcal{Y}}^\prime = f_b(f_g(\mathcal{X^\prime})) \oplus f_n(f_g(\mathcal{X^\prime})) $, where $\oplus$ is the concatenation operator. 
Analogously, we define $\hat{\mathcal{Y}}^{\prime\prime}$ as the network output for $\mathcal{X}^{\prime\prime}$.
The segmentation targets are defined as $\tilde{\mathcal{Y}}^\prime = \tilde{\mathcal{Y}}^\prime_n \cup \mathcal{Y}^\prime_b$, where $\tilde{\mathcal{Y}}^\prime_n$ are the pseudo-labels predicted with our approach and $\mathcal{Y}^\prime_b$ are the available targets for base classes (same for $\tilde{\mathcal{Y}}_n^{\prime\prime}$).
\\
At this point, we enforce prediction consistency between the swapped pseudo-labels of the two augmentations as:
\begin{equation}
    \label{eq:swapped_pred_task}
    \ell_\text{S}(\mathcal{X}) = \ell_\text{wCE}(\hat{\mathcal{Y}}^\prime, \tilde{\mathcal{Y}}^{\prime\prime}) + \ell_\text{wCE}(\hat{\mathcal{Y}}^{\prime\prime}, \tilde{\mathcal{Y}}^{\prime}),
\end{equation}
where $\ell_\text{wCE}$ is the weighted Cross Entropy loss.
The weights of the loss for the base classes are computed based on their occurrence frequency in the training set.
The weights of the loss for the novel classes are all set equally as their occurrence frequency in the dataset is unknown.

\noindent \textbf{Multi-headed segmentation:}
A single segmentation head may converge to a suboptimal feature space, thus producing suboptimal prototype solutions.
To further improve the segmentation quality, we use multiple novel class segmentation heads to optimise $f_\Theta$ based on different training solutions.
Different solutions increase the likelihood of producing a diverse partitioning of the feature space as they regularise with each other (they share the same backbone)~\citep{ji2019invariant}.
In practise, we concatenate the logits of the base class segmentation head with the outputs of each novel class segmentation head and we separately evaluate their loss for each novel class segmentation head at training time.

We task our network to over-cluster novel points, using segmentation heads that output $o\cdot C_n$ logits, where $o$ is the over-clustering factor. 
Previous studies empirically showed that this is beneficial to learn more informative features \citep{caron2020unsupervised, fini2021unified, mei2022data, ji2019invariant}. We observed the same and concur that over-clustering can be useful for increasing expressivity of the feature representations. 
The over-clustering heads are then discarded at inference time.

\subsection{Class-balanced queuing}\label{sec:class_balanced_queue}

Soft pseudo-labelling described in Sec.~\ref{sec:pseudo_labelling} produces an equipartite matching between the novel points and the class centroids.
However, it is likely that batches are sampled with point clouds containing novel classes with different cardinalities when dealing with 3D data.
In addition, some scenes may contain only a subset of the novel classes. 
Therefore, enforcing the equipartitioning constraint for each batch of the dataset could affect the learning of less frequent (long-tail) classes.
As a solution, we introduce a queue $\mathtt{Z}_q$ containing a randomly extracted portion of the features of the novel points from the previous iterations.
We use this additional data to mitigate the potential class imbalance that may occur during training. 
We compute $\mathtt{Z} \leftarrow \mathtt{Z} \oplus \mathtt{Z}_q$, where $\oplus$ is the concatenation operator, and execute the Sinkhorn-Knopp algorithm on this augmented version of $\mathtt{Z}$.
The obtained $\mathtt{Q}^* \in \mathbb{R}^{\mathcal{C}_n \times (m_c + |\mathtt{Z}_q|)}$ represents the assignment between the class prototypes and all the points in the augmented version of $\mathtt{Z}$. 
Being interested only in the pseudo-labels for the points in the actual batch, we retain only the first $m_c$ columns of $\mathtt{Q}^*$, discarding the additional information related to the points contained in $\mathtt{Z}_q$.

\subsection{Uncertainty-aware training and queuing}\label{sec:unc_train}

The optimisation of $f_\Theta$ through pseudo-labels and the insertion of the novel points into the queue $\mathtt{Z}_q$ can both benefit from the selection of novel points that are considered reliable by the network. 
We perform this selection by considering the class assignment probability $\hat{\mathcal{Y}}_n$ for the novel points.
In particular, we propose to apply a different threshold $\tau_c$ for each novel class $c \in \mathcal{C}_n$.
All the novel points predicted by the network as belonging to novel class $c$ with the confidence above $\tau_c$ are used during optimisation, and are kept as candidates for the insertion in the queue. All the other novel points are instead discarded.
We found that it is impractical to seek a fixed threshold for all the novel classes, while being also compatible with the variations of the class probabilities during training.
Therefore, we employ an adaptive threshold based on the class probabilities within each batch.

Our adaptive selection strategy operates as follows.
Firstly, we extract the novel points that have been predicted as part of novel class $c$ by the network.
Secondly, we compute $\tau_c$ as the $p$-th percentile of the class probabilities of these novel points.
Lastly, we retain only the novel points of class $c$ whose class probability is above the threshold $\tau_c$.
We define this selection strategy as the function
\begin{equation}
    \phi : (\mathcal{F}_n, \hat{\mathcal{Y}_n}) \times p \mapsto (\bar{\mathcal{F}}_n),
\end{equation}
where $\mathcal{F}_n$ is the set of feature vectors extracted from $f_g$ and $\hat{\mathcal{Y}}_n$ is the set of class probabilities predicted by the network for these points.
The selected features $\bar{\mathcal{F}}_n$ for the reliable novel points are both processed by the Sinkhorn-Knopp algorithm to generate the pseudo-labels and added to $\mathtt{Z}_q$ to make it more effective.

\begin{figure}[t]
    \centering
    \includegraphics[width=\linewidth]{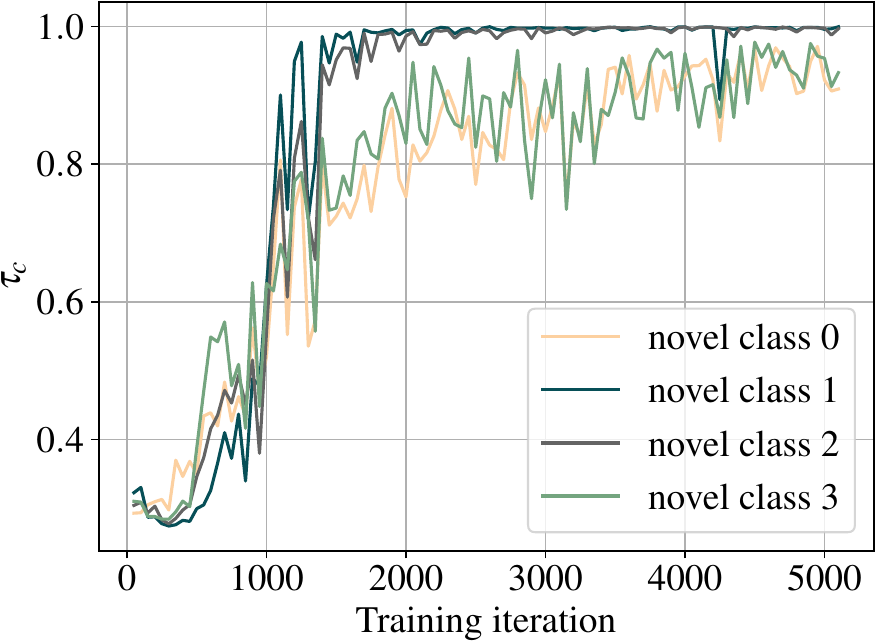}
    \vspace{-3mm}
    \caption{Evolution of the adaptive selection threshold $\tau_c$ when discovering four novel classes on S3DIS.}
    \label{fig:adaptive_thr}
\end{figure}

At the first optimisation iterations, the threshold $\tau_c$ is low for all the novel classes $c \in \mathcal{C}_n$ due to the network's random initialisation. 
However, each novel class is discovered during training, each threshold $\tau_c$ is expected to increase in an adaptive way to select novel points that are more and more reliable, resulting in a better optimisation of $f_\Theta$.
Fig.~\ref{fig:adaptive_thr} shows the evolution of the adaptive threshold $\tau_c$ when discovering four novel classes. The behaviour of the four different thresholds indicates that our method progressively selects more reliable novel points for training, thereby enhancing the optimisation process of $f_\Theta$, leading to effective discovery of the four novel classes.

\subsection{Incorporating semantic knowledge}\label{sec:semantic_distillation}
The optimisation of $f_\Theta$ through the ground-truth labels and the pseudo-labels generated as described in Sec.~\ref{sec:pseudo_labelling} arranges the feature space output by $f_g$ so to effectively separate representations of novel and base classes.
However, the supervision provided by ground-truth targets is significantly stronger than the self-supervision of the pseudo-labels. This unbalance could result into a sub-optimal organisation of the feature space, in which base class representations are compact and well-separated while novel class features are poorly clustered with more noise and less compactness.
To address this issue, we incorporate additional supervision for novel categories.
We employ an auxiliary neural network $f_a$ that is able to output point-level semantic-aligned features, such as a 3D foundation model~\citep{peng2023openscene}. Such architectures have shown great performance in multiple 3D scene understanding tasks, being able to reach the results of models specifically tailored for each single task. This good generalisation capability suggests that the feature spaces of 3D foundation model are well partitioned and organised according to semantics.
So, by guiding our network $f_\Theta$ to mimic the point-level features output by $f_a$, we can enhance the semantic organisation of its feature space.

The most natural choice when aligning the features of our network to the ones of $f_a$ would be to consider the feature space output by our backbone $f_g$. However, there is the risk that the distillation procedure from $f_a$ interferes with the Sinkhorn-Knopp algorithm in organising the feature space output by $f_g$.
So, we attach an additional projection head $f_s$ on top of the feature extractor $f_g$. The knowledge distillation from the foundation model is performed on the features $\mathcal{F}_s$ output by $f_s$, while the features $\mathcal{F}$ are reserved for the Sinkhorn-Knopp algorithm.
A well-separated representation of novel classes in the feature space of $f_s$, achieved through knowledge distillation from the foundation model, can in turn lead to improved separation of novel classes in the feature space of $f_g$. In fact, the feature space $f_s$ is on top of $f_g$ and they share the same underlying architecture.
The distillation procedure is performed by minimising the alignment objective:
\begin{equation}
    \ell_\text{A}(\mathcal{X}) = \ell_\text{cos}(\hat{\mathcal{F}}^\prime_s, \tilde{\mathcal{F}}^\prime_s) + \ell_\text{cos}(\hat{\mathcal{F}}^{\prime\prime}_s, \tilde{\mathcal{F}}^{\prime\prime}_s),
\end{equation}
where $\ell_\text{cos}$ is the cosine loss, $\hat{\mathcal{F}}^\prime_s = f_s(f_g(\mathcal{{X}^\prime}))$ and $\tilde{\mathcal{F}}^\prime_s = f_a(\mathcal{X}^\prime)$. The same applies for $\hat{\mathcal{F}}^{\prime\prime}_s$ and $\tilde{\mathcal{F}}^{\prime\prime}_s$ (Fig.~\ref{fig:augmentations}). Differently from $\ell_\text{S}$, in this case we do not use a swapped prediction task.

The projection head $f_s$ and features $\mathcal{F}_s$ are only considered during training and we ignore this branch of $f_\Theta$ at test time.

\begin{figure*}[t]
    \centering
    \includegraphics[width=\linewidth]{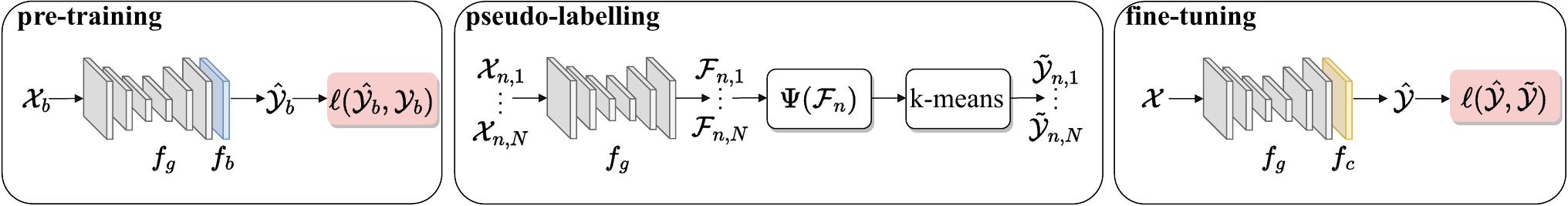}
    \vspace{-.4cm}
    \caption{Overview of EUMS$^\dag$, our adaptation of the method proposed by~\citep{zhao2022novel}. We first pre-train $f_g$ and $f_b$ considering only the base points in each point cloud. Using $f_g$, we extract the features of the novel points in each scene, that are filtered with the selection function $\Psi(\cdot)$. Then, we produce the pseudo-labels for the selected novel points by using the k-means algorithm. Lastly, we plug a new segmentation head $f_c$ into $f_g$ and fine-tune the complete model on both novel and base points, considering pseudo-labels and ground-truth labels respectively.}
    \label{fig:eums}
\end{figure*}

\section{Baseline methods for 3D Novel Class Discovery}

\newmethod and our earlier method \ourmethod~\citep{riz2023novel} are the first architectures proposed to tackle the task of Novel Class Discovery in point cloud semantic segmentation. So, in this work we also present two baseline methods related to 3D novel class discovery we can compare \newmethod to: the adaptation of EUMS from the image domain to the 3D point cloud domain (referred to EUMS$^\dag$) and the zero-shot testing of the OpenScene~\citep{peng2023openscene} model. These approaches hold significant importance in the relatively unexplored domain of 3D NCD, as they offer valuable insights into the challenges of such task. In particular, EUMS$^\dag$ serves as a baseline to highlight the challenges in naively adapting 2D methods to the 3D domain. The zero-shot testing with OpenScene provides instead a reference point, demonstrating the deep scene understanding capabilities of 3D Vision-Language Models and highlighting also the difficulties encountered in their application.


\subsection{Adapting NCD for 2D images to 3D point clouds} \label{sec:adaptation_Zhao}

One of the contributions of this work is to adapt the method proposed by \citep{zhao2022novel} for NCD in 2D semantic segmentation (EUMS) to 3D data. 
Our empirical evaluation (see Sec.~\ref{sec:experiments}) shows that the transposition of EUMS to the 3D domain has some limitations.
In particular, as described in Sec.~\ref{sec:intro}, EUMS uses two assumptions: \textbf{I}) the novel classes belong to the foreground and \textbf{II}) each image can contain at most one novel class.
This allows EUMS to leverage a saliency detection model to produce a foreground mask and a segmentation model pre-trained on the base classes to determine which portion of the image is background.
The portion of the image that belongs to both the foreground mask and the background mask is where features are then pooled.
EUMS computes a feature representation for each image by average pooling the features of the pixels belonging the unknown portion.
The feature representations of all the images in the dataset are clustered with K-Means by using the number of classes to discover as the target number of clusters.
EUMS shows that overclustering and entropy-based modelling can be exploited to improve the results.
The affiliation of a point to its cluster is used to produce hard pseudo-labels that are in turn used along with the ground-truth labels to fine-tune the pre-trained model.

With 3D point clouds, there is no concept of foreground and background (in contrast with \textbf{I}). Our adaptation is designed to discover the classes of all the unlabelled points (in contrast with \textbf{II}).
Therefore, given the unlabelled points of each point cloud, we randomly extract a subset of these by setting a ratio (e.g.~30\%) with upper bound (e.g.~1K) on the number of points to select.
We compute and collect their features for all the point clouds in the dataset and apply K-Means on the whole set of features.
Note that this clustering step is computationally expensive, and we had to use High Performance Computing to execute it.
The subsampling of the points was necessary to fit the data in the RAM (see Sec.~\ref{sec:experiments} for a detailed analysis).
Once the cluster prototypes are computed, we produce the hard pseudo-labels.
To enrich the set of pseudo-labels, we propagate the pseudo-label of each point to its nearest neighbour in the coordinate space.
This allows us to expand the subset of pseudo-labelled randomly selected points.
We also implement the other steps of overclustering and entropy-based modelling to boost the results.
Lastly, we fine-tune our model with these pseudo-labels.
We name our transposition of EUMS as EUMS$^\dag$ and report its block diagram in Fig.~\ref{fig:eums}.

\begin{figure}[t]
    \centering
    \includegraphics[width=0.9\linewidth]{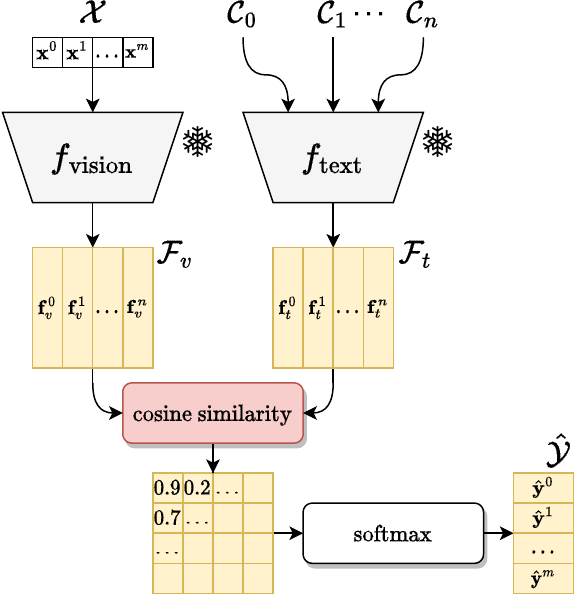}
    \caption{Zero-shot 3D semantic segmentation.
    We extract point-level features from the 3D vision encoder $f_{vision}$ and text embedding from the text encoder $f_{text}$. We assume class vocabularies to be known. Class predictions are assigned by similarity matching between point-level features and text embedding.}
    \label{fig:zero_shot}
\end{figure}

\subsection{Zero-shot testing with OpenScene}\label{sec:zero-shot}

3D Vision-Language Models (3D-VLMs) have shown promising generalization capabilities in scene understanding ~\citep{peng2023openscene}, especially in the context of 3D semantic segmentation. Their capabilities can be used off-the-shelf to recognize new objects within scenes by providing the correct text prompt. In this section, we assess the zero-shot semantic segmentation capabilities of the OpenScene models~\citep{peng2023openscene}. Our objective is to evaluate these 3D-VLMs on datasets that differ from the ones they were distilled from.
We consider this testing to be our lower bound and with \newmethod we aim to improve over it.
\\
We report in Fig.~\ref{fig:zero_shot} the implementation of our zero-shot experiment. The input point cloud $\mathcal{X}$ is forwarded into the frozen $f_\text{vision}$, resulting in point-wise CLIP-aligned features $\mathcal{F}_{\text{v}}$. Subsequently, the frozen text encoder $f_\text{text}$ is provided with classical prompts following the template \texttt{An image of a <CLASS>} for all the classes present in the dataset. This process yields the output $\mathcal{F}_{\text{t}}$.
Finally, we use the pairwise cosine similarity between each point and class name embedding to assign the predicted classes $\hat{\mathcal{Y}}$.
\\
In its naive version, this experiment would imply the use of dataset class names as input prompts with a simple text template,\textit{e.g.}, \texttt{An image of a <CLASS>}. However, we argue that the simple use of the given dataset class names may limit zero-shot capabilities of OpenScene. Dataset names are selected from a (large) set of synonyms with the same meaning, e.g., \textit{person} can also be indicated with the words \textit{pedestrian} or \textit{walker}.
Oppositely, the CLIP text encoder has been trained with descriptors that contain different terms for the same concept and consequently it has learnt slightly different embeddings for each of the synonyms of the same concept. So, deriving a single text embedding for each class (e.g.~for the dataset class name \textit{person}) does not ensure the coverage of all the variations inside such class (e.g.~for \textit{walker} and \textit{pedestrian}). 
To this extent, we propose to map each class name to four additional synonyms, which capture different meanings under the same class name. Similarly to what proposed for the text templates, we use the five different embeddings for each class as an ensemble. This enable a better coverage ot the CLIP feature space and increases the zero-shot capabilities of OpenScene. 
The synonyms for each class are extracted by querying the WordNet dataset~\citep{miller1995wordnet}. When the returned words are not enough, we also make use of \textit{Thesaurus.com} for additional synonyms.
To further exploit the sensitivity of $f_{text}$ to text input~\citep{radford2021clip}, we use an ensemble of text templates to produce robust and stable predictions. In particular, we use the 80 templates originally proposed in \citep{radford2021clip}.

Tab.~\ref{tab:results_poss}, \ref{tab:results_semantickitti} \& \ref{tab:results_s3dis} show the OpenScene zero-shot performance on SemanticPOSS, SemanticKITTI, and S3DIS respectively.
We report such results as ``\textit{OpenScene$^\star$}'' to highlight the usage of the proposed pipeline involving synonyms and templates.
On SemanticPOSS, \textit{OpenScene$^\star$} using ensembles of 3 synonyms reaches $21.18\%$ mIoU, surpassing the standard testing with single class words by around $2.4\%$. The testing with 5 synonyms shows even better results, showing an overall $23.05\%$ mIoU, with an improvement over the baseline of around $4.3\%$.
On SemanticKITTI, \textit{OpenScene$^\star$} reaches $19.28\%$ mIoU with 5 synonyms, improving over the testing with 3 synonyms by around $0.7\%$ and gaining $1.0\%$ mIoU over the standard testing with single class words.
On S3DIS, we report the results without using ensembles of synonyms, since the furniture class names (e.g.~\textit{chair}) of this dataset are too specific to find suitable synonyms for each class label. The zero-shot testing of \textit{OpenScene$^\star$} on the S3DIS dataset results in an overall $36.76$ mIoU.

\section{Experimental results}\label{sec:experiments}
\subsection{Experiments}\label{sec:experiments_sub}

\noindent \textbf{Datasets.} 
We evaluate our approach on SemanticKITTI~\citep{behley2019semantickitti, geiger2012cvpr, behley2021ijrr}, SemanticPOSS~\citep{pan2020semanticposs} and Stanford 3D Indoor Scene Dataset (S3DIS)~\citep{armeni20163d}.
SemanticKITTI~\citep{behley2019semantickitti} consists of 43,552 point cloud acquisitions with point-level annotations of 19 semantic classes.
Based on the conventional benchmark guidelines~\citep{behley2019semantickitti}, we use sequence $08$ for validation and the other sequences for training.
SemanticPOSS~\citep{pan2020semanticposs} consists of 2,988 real-world point cloud acquisitions with point-level annotations of 13 semantic classes.
Based on the conventional benchmark guidelines~\citep{pan2020semanticposs}, we use sequence $03$ for validation and the other sequences for training. 
S3DIS~\citep{armeni20163d} consists of 271 indoor RGB-D scans with point-level annotations of 13 semantic classes.
We follow the official split~\citep{armeni20163d} and use Area\_5 for validation and the other areas for training.

\noindent \textbf{Experimental protocol for 3D NCD.}
Similarly to what proposed by \citep{zhao2022novel} in the 2D domain, we create different splits of each dataset to validate the NCD performance with point cloud data.
We create four splits for SemanticKITTI, SemanticPOSS, and S3DIS.
We refer to these splits as SemanticKITTI-$n^i$, SemanticPOSS-$n^i$, and S3DIS-$n^i$, where $i$ indexes the split.
In each set, the novel classes and the base classes correspond to unlabelled and labelled points, respectively.
Tabs.~\ref{tab:KITTI_folds}, \ref{tab:POSS_folds} \& \ref{tab:S3DIS_folds} detail the splits of our datasets.
These splits are selected based on their class distribution in the dataset and on the semantic relationship between novel and base classes, \textit{e.g.}~in KITTI-$4^3$ the base class \textit{motorcycle} can be helpful to discover the novel class \textit{motorcyclist}.

\begin{table}[t]
    \centering
    \caption{SemanticKITTI splits, is defined as KITTI-$n^i$, where $n$ is the number of novel classes and $i$ is the split index.}
    \label{tab:KITTI_folds}
    \vspace{-.2cm}
    \begin{tabular}{ll}
        \toprule
        Split & Novel Classes \\
        \midrule
        KITTI-$5^0$ & \textit{building}, \textit{road}, \textit{sidewalk}, \textit{terrain}, \textit{veget.} \\
        KITTI-$5^1$ & \textit{car}, \textit{fence}, \textit{other-ground}, \textit{parking}, \textit{trunk} \\
        KITTI-$5^2$ & \textit{motorc.}, \textit{other-v.}, \textit{pole}, \textit{traffic-s.}, \textit{truck} \\
        KITTI-$4^3$ & \textit{bicycle}, \textit{bicyclist}, \textit{motorcyclist}, \textit{person} \\
        \bottomrule
    \end{tabular}
\end{table}
\begin{table}[t]
    \centering
    \caption{SemanticPOSS splits, defined as POSS-$n^i$, where $n$ is the number of novel classes and $i$ is the split index.}
    \label{tab:POSS_folds}
    \vspace{-.2cm}
    \begin{tabular}{ll}
        \toprule
        Split & Novel Classes  \\
        \midrule
        POSS-$4^0$ & \textit{building}, \textit{car}, \textit{ground}, \textit{plants} \\
        POSS-$3^1$ & \textit{bike}, \textit{fence}, \textit{person} \\
        POSS-$3^2$ & \textit{pole}, \textit{traffic-sign}, \textit{trunk} \\
        POSS-$3^3$ & \textit{cone-stone}, \textit{rider}, \textit{trashcan} \\
        \bottomrule
    \end{tabular}
\end{table}
\begin{table}[t]
    \centering
    \caption{S3DIS splits, defined as S3DIS-$n^i$, where $n$ is the number of novel classes and $i$ is the split index.}
    \label{tab:S3DIS_folds}
    \vspace{-.2cm}
    \begin{tabular}{ll}
        \toprule
        Split & Novel Classes  \\
        \midrule
        S3DIS-$4^0$ & \textit{ceiling}, \textit{clutter}, \textit{floor}, \textit{wall} \\
        S3DIS-$3^1$ & \textit{chair}, \textit{door}, \textit{table} \\
        S3DIS-$3^2$ & \textit{beam}, \textit{bookcase}, \textit{column} \\
        S3DIS-$3^3$ & \textit{board}, \textit{sofa}, \textit{window} \\
        \bottomrule
    \end{tabular}
\end{table}

\begin{table*}[t]
    \centering
    \caption{Novel class discovery results on SemanticPOSS. 
    \newmethod outperforms EUMS$^\dag$ and \ourmethod on all the four splits. 
    Full supervision: model trained with labels for base and novel classes. OpenScene$^\star$: reference described in Sec.~\ref{sec:zero-shot} (``\textit{n} Syn'' indicates the number \textit{n} of synonyms used to build the ensembles). EUMS$^\dag$: baseline described in Sec.~\ref{sec:adaptation_Zhao}. Highlighted values are the novel classes in each split.}
    \vspace{-.2cm}
    \label{tab:results_poss}
    \tabcolsep 6pt
    \resizebox{\textwidth}{!}{%
    \begin{tabular}{l|l|ccccccccccccc|ccc}
        \toprule
        \multirow{2}{*}{\textbf{Split}} & \multirow{2}{*}{\textbf{Model}} & \multirow{2}{*}{\rotatebox{45}{\textbf{bike}}} & \multirow{2}{*}{\rotatebox{45}{\textbf{build.}}} & \multirow{2}{*}{\rotatebox{45}{\textbf{car}}} & \multirow{2}{*}{\rotatebox{45}{\textbf{cone.}}} & \multirow{2}{*}{\rotatebox{45}{\textbf{fence}}} & \multirow{2}{*}{\rotatebox{45}{\textbf{grou.}}} & \multirow{2}{*}{\rotatebox{45}{\textbf{pers.}}} & \multirow{2}{*}{\rotatebox{45}{\textbf{plant}}} & \multirow{2}{*}{\rotatebox{45}{\textbf{pole}}} & \multirow{2}{*}{\rotatebox{45}{\textbf{rider}}} & \multirow{2}{*}{\rotatebox{45}{\textbf{traf.}}} & \multirow{2}{*}{\rotatebox{45}{\textbf{trash.}}} & \multirow{2}{*}{\rotatebox{45}{\textbf{trunk}}} &  \multicolumn{3}{c}{\textbf{mIoU}} \\
         &  &  &  &  &  &  &  &  &  &  &  &  &  &  & \textbf{Novel} & \textbf{Base} & \textbf{All}\\
        \midrule
        
         &  Full supervision & 43.20 & 71.30 & 33.00 & 32.50 & 44.60  & 78.50 & 61.80 & 73.90 & 30.90 & 54.70 & 26.70 & 11.00 & 19.30 & - & - & 44.72 \\
         \midrule

         &  OpenScene$^\star$ 1 Syn. & 0.08 & 50.34 & 38.71 & 1.31 & 6.37  & 61.12 & 31.78 & 48.47 & 2.45 & 0.05 & 0.00 & 2.74 & 0.57 & - & - & 18.77 \\
         &  OpenScene$^\star$ 3 Syn. & 0.06 & 52.29 & 39.32 & 0.34 & 5.97  & 61.96 & 41.79 & 61.00 & 3.21 & 0.00 & 0.00 & 4.46 & 4.98 & - & - & 21.18 \\
         &  OpenScene$^\star$ 5 Syn. & 0.06 & 55.60 & 38.62 & 0.12 & 6.63  & 67.04 & 42.00 & 67.81 & 5.72 & 2.83 & 1.61 & 5.37 & 6.28 & - & - & 23.05 \\
        \midrule
        
        \multirow{2}{*}{POSS-$4^0$} & EUMS\dag \citep{zhao2022novel} & 25.67 & \CC{novelcolor}3.98 & \CC{novelcolor}0.56 & 16.44 & 29.40 & \CC{novelcolor}36.76 & 43.84 & \CC{novelcolor}28.46 & 13.13 & 26.75 & 18.18 & 3.34 & 16.91 & \CC{novelcolor}17.44 & 21.52 & 20.26 \\
         & \ourmethod \citep{riz2023novel} & 35.47 & \CC{novelcolor}30.35 & \CC{novelcolor}1.24 & 13.52 & 24.13 & \CC{novelcolor}69.14 & 44.70 & \CC{novelcolor}42.07 & 19.19 & 47.65 & 24.44 & 8.17 & 21.82 & \CC{novelcolor}35.70 & 26.57 & 29.38 \\
         & \newmethod (Ours) & 34.21 & \CC{novelcolor}\textbf{58.80} & \CC{novelcolor}\textbf{10.04} & 13.20 & 18.69 & \CC{novelcolor}\textbf{77.25} & 45.84 & \CC{novelcolor}\textbf{58.62} & 17.27 & 48.35 & 22.61 & 8.72 & 22.85 & \CC{novelcolor}\textbf{51.18} & 25.75 & 33.57 \\
        \midrule
        
        \multirow{2}{*}{POSS-$3^1$} & EUMS\dag \citep{zhao2022novel} & \CC{novelcolor}15.17 & 67.98 & 28.02 & 23.98 & \CC{novelcolor}11.88 & 75.07 & \CC{novelcolor}35.98 & 74.46 & 26.91 & 48.56 & 26.00 & 5.60 & 23.05 & \CC{novelcolor}21.01 & 39.96 & 35.59 \\
         & \ourmethod \citep{riz2023novel} & \CC{novelcolor}\textbf{29.35} & 71.35 & 28.70 & 12.21 & \CC{novelcolor}3.94 & 78.24 & \CC{novelcolor}\textbf{56.78} & 74.21 & 18.29 & 38.88 & 23.31 & 13.74 & 23.51 & \CC{novelcolor}30.02 & 38.24 & 36.35 \\
         & \newmethod (Ours) & \CC{novelcolor}16.29 & 71.37 & 30.01 & 19.75 & \CC{novelcolor}\textbf{24.90} & 77.14 & \CC{novelcolor}54.96 & 73.36 & 15.76 & 38.43 & 22.28 & 15.74 & 23.59 & \CC{novelcolor}\textbf{32.05} & 38.72 & 37.18 \\
        \midrule
        
        \multirow{2}{*}{POSS-$3^2$} & EUMS\dag \citep{zhao2022novel} & 40.14 & 69.45 & 27.67 & 13.50 & 34.86 & 76.03 & 54.66 & 75.59 & \CC{novelcolor}5.27 & 39.22 & \CC{novelcolor}7.79 & 8.52 & \CC{novelcolor}11.85 & \CC{novelcolor}8.31 & 43.96 & 35.74 \\
         & \ourmethod \citep{riz2023novel} & 37.16 & 71.81 & 29.74 & 14.64 & 28.38 & 77.53 & 52.09 & 73.00 & \CC{novelcolor}\textbf{11.51} & 47.11 & \CC{novelcolor}0.54 & 10.20 & \CC{novelcolor}14.79 & \CC{novelcolor}8.95 & 44.17 & 36.04 \\
         & \newmethod (Ours) & 38.37 & 72.45 & 27.96 & 14.47 & 26.19 & 78.08 & 54.73 & 74.31 & \CC{novelcolor}9.99 & 48.25 & \CC{novelcolor}\textbf{22.98} & 10.16 & \CC{novelcolor}\textbf{17.71} & \CC{novelcolor}\textbf{16.89} & 44.50 & 38.13 \\
        \midrule
        
        \multirow{2}{*}{POSS-$3^3$} & EUMS\dag \citep{zhao2022novel} & 41.17 & 70.68 & 28.08 & \CC{novelcolor}4.34 & 38.27 & 76.66 & 38.29 & 75.35 & 25.76 & \CC{novelcolor}34.34 & 28.31 & \CC{novelcolor}0.36 & 24.40 & \CC{novelcolor}13.01 & 44.70 & 37.38 \\
         & \ourmethod \citep{riz2023novel} & 38.55 & 70.36 & 30.91 & \CC{novelcolor}0.00 & 29.38 & 76.50 & 55.98 & 71.84 & 17.03 & \CC{novelcolor}31.87 & 26.15 & \CC{novelcolor}0.95 & 22.57 & \CC{novelcolor}10.94 & 43.93 & 36.32 \\
         & \newmethod (Ours) & 39.40 & 70.33 & 30.03 & \CC{novelcolor}\textbf{9.10} & 26.84 & 77.64 & 54.32 & 72.54 & 16.02 & \CC{novelcolor}\textbf{49.89} & 28.13 & \CC{novelcolor}\textbf{1.31} & 23.51 & \CC{novelcolor}\textbf{20.10} & 43.88 & 38.39 \\

        \bottomrule
        \addlinespace[2.5pt]
         \multicolumn{11}{c}{} & \multirow{3}{*}{Avg} & \multicolumn{3}{|l|}{EUMS\dag \citep{zhao2022novel}} & \CC{novelcolor}14.94  & 37.54 & 32.24 \\
         \multicolumn{11}{c}{} &  & \multicolumn{3}{|l|}{\ourmethod \citep{riz2023novel}} & \CC{novelcolor}21.40  & 38.23 & 34.52 \\
         \multicolumn{11}{c}{} &  & \multicolumn{3}{|l|}{\newmethod (Ours)} & \CC{novelcolor}\textbf{30.05}  & 38.21 & 36.82 \\
         \cmidrule[1pt]{12-18}
    \end{tabular}
    }
\end{table*}

We quantify the performance by using the mean Intersection over Union (mIoU), which is defined as the average IoU across the considered classes~\citep{behley2019semantickitti}.
We provide separate mIoU values for the base and novel classes.
We also report the overall mIoU computed across all the classes in the dataset for completeness.

\noindent \textbf{Implementation Details.}
We implement our network based on a MinkowskiUNet-34C network~\citep{choy20194d}.
Point-level features are extracted from the penultimate layer.
The segmentation heads $f_b$ and $f_n$ are implemented as linear layers, producing output logits for each point in the batched point clouds.
The projection head $f_s$ is a sequence of linear layer, batch norm, ReLU, and another linear layer.
The auxiliary network $f_a$ is the MinkowskiUNet-18 network presented in OpenScene~\citep{peng2023openscene}. We use the version distilled from nuScenes-OpenSeg for SemanticKITTI-$n^i$ and  SemanticPOSS-$n^i$, and the version distilled from ScanNet-OpenSeg for S3DIS-$n^i$.
We train our network for 10 epochs for SemanticKITTI-$n^i$ and SemanticPOSS-$n^i$, and for 50 epochs for S3DIS-$n^i$.
We use the SGD optimizer, with momentum 0.9 and weight decay 0.0001. 
Our learning rate scheduler consists of linear warm-up and cosine annealing, with $lr_{max} = 10^{-2}$ and $lr_{min} = 10^{-5}$. 
We train with batch size equal to 4. 
We employ five segmentation heads, that are used in synergy with an equal number of over-clustering heads, with $o = 3$.
In $\phi$, we set $p=0.5$ for SemanticKITTI-$n^i$, and $p=0.3$ for SemanticPOSS-$n^i$ and S3DIS-$n^i$.
We set $\gamma=3.0$ for SemanticKITTI-$n^i$, and $\gamma=7.0$ for SemanticPOSS-$n^i$ and S3DIS.
We adapted the implementation of the Sinkhorn-Knopp algorithm~\citep{cuturi2013sinkhorn} from the code provided by \citep{caron2020unsupervised}, with the introduction of the queue and an in-place normalisation steps.
Similarly to \citep{caron2020unsupervised}, we set $n_{sk\_iters}=3$, while we adopt a linear decay for $\epsilon$, with $\epsilon_{start}=0.3, \epsilon_{end}=0.05$.

\begin{table*}[t]
    \centering
    \caption{Novel class discovery results on SemanticKITTI.
    \newmethod outperforms EUMS$^\dag$ and \ourmethod on all four splits. 
    Full supervision: model trained with labels for base and novel classes. OpenScene$^\star$: reference described in Sec.~\ref{sec:zero-shot} (``\textit{n} Syn'' indicates the number \textit{n} of synonyms used to build the ensembles). EUMS$^\dag$: baseline described in Sec.~\ref{sec:adaptation_Zhao}. Highlighted values are the novel classes in each split.}
    \vspace{-.2cm}
    \label{tab:results_semantickitti}
    \tabcolsep 3pt
    \resizebox{\textwidth}{!}{%
    \begin{tabular}{l|l|ccccccccccccccccccc|ccc}
        \toprule
        \multirow{2}{*}{\textbf{Split}} & \multirow{2}{*}{\textbf{Model}} & \multirow{2}{*}{\rotatebox{45}{\textbf{bi.cle}}} & \multirow{2}{*}{\rotatebox{45}{\textbf{b.clst}}} & \multirow{2}{*}{\rotatebox{45}{\textbf{build.}}} & \multirow{2}{*}{\rotatebox{45}{\textbf{car}}} & \multirow{2}{*}{\rotatebox{45}{\textbf{fence}}} & \multirow{2}{*}{\rotatebox{45}{\textbf{mt.cle}}} & \multirow{2}{*}{\rotatebox{45}{\textbf{m.clst}}} & \multirow{2}{*}{\rotatebox{45}{\textbf{oth-g.}}} & \multirow{2}{*}{\rotatebox{45}{\textbf{oth-v.}}} & \multirow{2}{*}{\rotatebox{45}{\textbf{park.}}} & \multirow{2}{*}{\rotatebox{45}{\textbf{pers.}}} & \multirow{2}{*}{\rotatebox{45}{\textbf{pole}}} & \multirow{2}{*}{\rotatebox{45}{\textbf{road}}} & \multirow{2}{*}{\rotatebox{45}{\textbf{sidew.}}} & \multirow{2}{*}{\rotatebox{45}{\textbf{terra.}}} & \multirow{2}{*}{\rotatebox{45}{\textbf{traff.}}} & \multirow{2}{*}{\rotatebox{45}{\textbf{truck}}} & \multirow{2}{*}{\rotatebox{45}{\textbf{trunk}}} & \multirow{2}{*}{\rotatebox{45}{\textbf{veget.}}} & \multicolumn{3}{c}{\textbf{mIoU}}\\
         &  &  &  &  &  &  &  &  &  &  &  &  &  &  &  &  &  &  &  &  & \textbf{Novel} & \textbf{Base} & \textbf{All}\\
        \midrule
        
         & Full supervision & 6.30 & 39.50 & 85.40 & 90.00 & 23.20 & 20.30 & 5.70 & 3.90 & 18.00 & 28.90 & 31.00 & 40.60 & 90.90 & 74.60 & 62.10 & 20.50 & 62.90 & 46.20 & 83.90 & - & - & 43.89 \\
         \midrule
         
         & OpenScene$^\star$ 1 Syn. & 0.00 & 5.20 & 40.59 & 55.57 & 8.12 & 11.22 & 0.50 & 0.05 & 4.26 & 0.10 & 17.16 & 4.05 & 62.89 & 34.74 & 0.00 & 0.04 & 41.39 & 0.31 & 61.25 & - & - & 18.29 \\
         & OpenScene$^\star$ 3 Syn. & 0.04 & 0.00 & 42.56 & 44.53 & 8.01 & 8.55 & 5.25 & 0.09 & 0.27 & 0.25 & 23.80 & 6.19 & 45.57 & 36.71 & 11.48 & 0.01 & 41.30 & 5.48 & 72.45 & - & - & 18.55 \\
         & OpenScene$^\star$ 5 Syn. & 0.18 & 1.53 & 48.51 & 32.89 & 7.99 & 9.29 & 4.34 & 0.00 & 0.06 & 0.22 & 21.19 & 8.08 & 38.20 & 35.62 & 34.18 & 1.36 & 41.20 & 6.82 & 74.60 & - & - & 19.28 \\
        \midrule
        
        \multirow{2}{*}{KITTI-$5^0$} & EUMS\dag \citep{zhao2022novel} & 5.28 & 39.96 & \CC{novelcolor}15.77 & 79.20 & 9.03 & 16.89 & 2.52 & 0.07 & 11.39 & 14.40 & 12.67 & 29.17 & \CC{novelcolor}42.58 & \CC{novelcolor}26.10 & \CC{novelcolor}0.05 & 10.30 & 47.37 & 37.92 & \CC{novelcolor}38.35 & \CC{novelcolor}24.57 & 21.08 & 23.11 \\
          & \ourmethod \citep{riz2023novel} & 5.59 & 47.76 & \CC{novelcolor}52.68 & 82.60 & 13.76 & 25.55 & 1.36 & 1.66 & 14.52 & 19.80 & 25.86 & 32.12 & \CC{novelcolor}\textbf{56.74} & \CC{novelcolor}8.08 & \CC{novelcolor}23.84 & 14.28 & 49.41 & 36.18 & \CC{novelcolor}\textbf{44.17} & \CC{novelcolor}37.10 & 24.70 & 29.62 \\
         & \newmethod (Ours) & 6.64 & 43.88 & \CC{novelcolor}\textbf{71.95} & 83.34 & 13.63 & 24.74 & 2.47 & 2.40 & 15.12 & 18.67 & 24.61 & 31.60 & \CC{novelcolor}49.47 & \CC{novelcolor}\textbf{43.15} & \CC{novelcolor}\textbf{27.36} & 15.68 & 42.12 & 38.52 & \CC{novelcolor}37.46 & \CC{novelcolor}\textbf{45.88} & 25.96 & 31.20 \\
        \midrule
        
        \multirow{2}{*}{KITTI-$5^1$} & EUMS\dag \citep{zhao2022novel} & 7.53 & 42.41 & 79.97 & \CC{novelcolor}\textbf{76.77} & \CC{novelcolor}8.62 & 19.58 & 1.39 & \CC{novelcolor}\textbf{0.57} & 12.03 & \CC{novelcolor}14.14 & 13.95 & 40.74 & 86.32 & 66.45 & 56.29 & 11.97 & 44.79 & \CC{novelcolor}20.94 & 72.40 & \CC{novelcolor}24.21 & 37.06 & 35.62 \\
         & \ourmethod \citep{riz2023novel} & 7.36 & 51.23 & 84.53 & \CC{novelcolor}50.87 & \CC{novelcolor}7.27 & 28.93 & 1.76 & \CC{novelcolor}0.00 & 22.20 & \CC{novelcolor}\textbf{19.39} & 30.42 & 37.61 & 90.07 & 72.18 & 60.75 & 16.78 & 57.34 & \CC{novelcolor}\textbf{49.25} & 85.12 & \CC{novelcolor}25.36 & 43.09 & 40.69 \\
         & \newmethod (Ours) & 7.58 & 43.48 & 85.12 & \CC{novelcolor}68.70 & \CC{novelcolor}\textbf{18.98} & 24.42 & 3.48 & \CC{novelcolor}0.00 & 23.86 & \CC{novelcolor}19.09 & 27.00 & 36.50 & 89.30 & 71.92 & 61.99 & 17.16 & 55.85 & \CC{novelcolor}29.42 & 84.37 & \CC{novelcolor}\textbf{27.24} & 45.15 & 40.43 \\
        \midrule
        
        \multirow{2}{*}{KITTI-$5^2$} & EUMS\dag \citep{zhao2022novel} & 8.26 & 50.78 & 82.98 & 88.05 & 17.88 & \CC{novelcolor}2.75 & 2.32 & 0.17 & \CC{novelcolor}3.16 & 25.40 & 24.98 & \CC{novelcolor}20.20 & 88.30 & 71.04 & 57.85 & \CC{novelcolor}8.63 & \CC{novelcolor}\textbf{27.16} & 38.36 & 76.95 & \CC{novelcolor}12.38 & 42.22 & 36.59 \\
         & \ourmethod \citep{riz2023novel} & 6.72 & 49.24 & 86.36 & 90.79 & 23.68 & \CC{novelcolor}2.69 & 0.58 & 1.87 & \CC{novelcolor}\textbf{15.46} & 29.48 & 27.92 & \CC{novelcolor}\textbf{36.39} & 90.26 & 73.39 & 61.21 & \CC{novelcolor}17.83 & \CC{novelcolor}10.32 & 46.16 & 84.29 & \CC{novelcolor}16.54 & 44.80 & 39.72 \\
         & \newmethod (Ours) & 6.79 & 48.30 & 86.08 & 89.88 & 22.20 & \CC{novelcolor}\textbf{9.27} & 0.56 & 3.55 & \CC{novelcolor}10.51 & 28.35 & 27.10 & \CC{novelcolor}23.81 & 90.64 & 73.79 & 61.93 & \CC{novelcolor}\textbf{22.31} & \CC{novelcolor}22.11 & 46.06 & 83.76 & \CC{novelcolor}\textbf{17.60} & 47.85 & 39.84 \\
        \midrule
        
        \multirow{2}{*}{KITTI-$4^3$} & EUMS\dag \citep{zhao2022novel} & \CC{novelcolor}3.95 & \CC{novelcolor}2.47 & 80.10 & 87.21 & 16.81 & 14.02 & \CC{novelcolor}\textbf{14.98} & 0.31 & 14.13 & 20.77 & \CC{novelcolor}6.80 & 37.59 & 86.79 & 66.50 & 55.26 & 16.20 & 40.62 & 38.37 & 76.15 & \CC{novelcolor}7.05 & 43.39 & 35.74 \\
         & \ourmethod \citep{riz2023novel} & \CC{novelcolor}2.32 & \CC{novelcolor}27.83 & 86.04 & 89.89 & 23.06 & 24.47 & \CC{novelcolor}2.92 & 3.06 & 18.19 & 30.09 & \CC{novelcolor}\textbf{16.32} & 39.90 & 90.65 & 73.51 & 61.04 & 17.40 & 49.76 & 44.01 & 83.18 & \CC{novelcolor}12.35 & 48.95 & 41.24 \\
         & \newmethod (Ours) & \CC{novelcolor}\textbf{4.65} & \CC{novelcolor}\textbf{31.51} & 84.55 & 88.65 & 22.81 & 23.28 & \CC{novelcolor}8.23 & 2.62 & 17.89 & 28.69 & \CC{novelcolor}15.05 & 38.26 & 89.71 & 72.48 & 60.76 & 16.14 & 43.34 & 45.70 & 82.87 & \CC{novelcolor}\textbf{14.86} & 47.85 & 40.91 \\
         
        \bottomrule
        \addlinespace[2.5pt]
        \multicolumn{16}{c}{} & \multirow{3}{*}{Avg} & \multicolumn{4}{|l|}{EUMS\dag \citep{zhao2022novel}} & \CC{novelcolor}17.05  & 35.94 & 32.76 \\
        \multicolumn{16}{c}{} & & \multicolumn{4}{|l|}{\ourmethod \citep{riz2023novel}} & \CC{novelcolor}22.84  & 42.39 & 37.73 \\
        \multicolumn{16}{c}{} & & \multicolumn{4}{|l|}{\newmethod (Ours)} & \CC{novelcolor}\textbf{26.39}  & 41.69 & 38.10 \\
        \cmidrule[1pt]{17-24} 

    \end{tabular}
    }
\end{table*}
\begin{table*}[t]
    \centering
    \caption{Novel class discovery results on S3DIS.
    \newmethod outperforms EUMS$^\dag$ on all the four splits and \ourmethod on three out of the four splits.
    Full supervision: model trained with annotations for base and novel classes. OpenScene$^\star$: reference described in Sec.~\ref{sec:zero-shot} (``\textit{n} Syn'' indicates the number \textit{n} of synonyms used to build the ensembles). EUMS$^\dag$: baseline described in Sec.~\ref{sec:adaptation_Zhao}. Highlighted values are the novel classes in each split.}
    \vspace{-.2cm}
    \label{tab:results_s3dis}
    \tabcolsep 6pt
    \resizebox{\textwidth}{!}{%
    \begin{tabular}{l|l|ccccccccccccc|ccc}
        \toprule
        \multirow{2}{*}{\textbf{Split}} & \multirow{2}{*}{\textbf{Model}} & \multirow{2}{*}{\rotatebox{45}{\textbf{beam}}} & \multirow{2}{*}{\rotatebox{45}{\textbf{board}}} & \multirow{2}{*}{\rotatebox{45}{\textbf{book.}}} & \multirow{2}{*}{\rotatebox{45}{\textbf{ceil.}}} & \multirow{2}{*}{\rotatebox{45}{\textbf{chair}}} & \multirow{2}{*}{\rotatebox{45}{\textbf{clutt.}}} & \multirow{2}{*}{\rotatebox{45}{\textbf{col.}}} & \multirow{2}{*}{\rotatebox{45}{\textbf{door}}} & \multirow{2}{*}{\rotatebox{45}{\textbf{floor}}} & \multirow{2}{*}{\rotatebox{45}{\textbf{sofa}}} & \multirow{2}{*}{\rotatebox{45}{\textbf{table}}} & \multirow{2}{*}{\rotatebox{45}{\textbf{wall}}} & \multirow{2}{*}{\rotatebox{45}{\textbf{wind.}}} &  \multicolumn{3}{c}{\textbf{mIoU}} \\
         &  &  &  &  &  &  &  &  &  &  &  &  &  &  & \textbf{Novel} & \textbf{Base} & \textbf{All}\\
        \midrule
        
         & Full supervision & 0.05 & 13.46 & 58.45 & 75.88 & 74.72 & 35.08 & 22.55 & 39.62 & 91.29 & 21.31 & 67.55 & 68.27 & 9.33 & - & - & 44.43 \\
         \midrule
        
         & OpenScene$^\star$ 1 Syn. & 0.00 & 0.00 & 42.36 & 72.78 & 56.25 & 10.81 & 0.00 & 47.17 & 85.53 & 45.48 & 42.31 & 59.24 & 15.95 & - & - & 36.76 \\
        \midrule
        
        \multirow{2}{*}{S3DIS-$4^0$} & EUMS\dag \citep{zhao2022novel} & 0.02 & 12.15 & 42.27 & \CC{novelcolor}41.65 & 59.30 & \CC{novelcolor}10.08 & 19.85 & 24.29 & \CC{novelcolor}0.23 & 26.99 & 43.50 & \CC{novelcolor}3.56 & 6.03 & \CC{novelcolor}13.88 & 26.04 & 22.30 \\
         & \ourmethod \citep{riz2023novel} & 0.04 & 8.05 & 52.79 & \CC{novelcolor}48.00 & 67.39 & \CC{novelcolor}\textbf{20.03} & 25.98 & 36.83 & \CC{novelcolor}0.00 & 38.12 & 63.05 & \CC{novelcolor}36.28 & 6.48 & \CC{novelcolor}26.08 & 33.19 & 31.00 \\
         & \newmethod  (Ours) & 0.55 & 0.12 & 49.86 & \CC{novelcolor}\textbf{81.01} & 72.82 & \CC{novelcolor}9.99 & 28.48 & 35.48 & \CC{novelcolor}\textbf{94.39} & 43.50 & 64.37 & \CC{novelcolor}\textbf{38.37} & 2.91 & \CC{novelcolor}\textbf{55.94} & 33.12 & 40.14 \\
        \midrule
        
        \multirow{2}{*}{S3DIS-$3^1$} & EUMS\dag \citep{zhao2022novel} & 0.17 & 14.06 & 34.90 & 72.63 & \CC{novelcolor}2.67 & 20.11 & 19.70 & \CC{novelcolor}6.74 & 87.56 & 26.05 & \CC{novelcolor}19.66 & 63.58 & 3.66 & \CC{novelcolor}9.69 & 34.24 & 28.58 \\
         & \ourmethod \citep{riz2023novel} & 0.00 & 8.14 & 54.44 & 78.49 & \CC{novelcolor}22.49 & 37.03 & 27.81 & \CC{novelcolor}19.07 & 94.28 & 55.17 & 49.69\CC{novelcolor} & 64.34 & 11.00 & \CC{novelcolor}30.41 & 43.07 & 40.15 \\
         & \newmethod  (Ours) & 0.00 & 8.81 & 53.78 & 81.77 & \CC{novelcolor}\textbf{58.14} & 36.61 & 27.62 & \CC{novelcolor}\textbf{42.43} & 94.29 & 57.45 & \CC{novelcolor}\textbf{59.90} & 63.52 & 10.98 & \CC{novelcolor}\textbf{53.49} & 43.49 & 45.79 \\
        \midrule
        
        \multirow{2}{*}{S3DIS-$3^2$} & EUMS\dag \citep{zhao2022novel} & \CC{novelcolor}0.03 & 3.90 & \CC{novelcolor}27.36 & 76.43 & 68.21 & 23.99 & \CC{novelcolor}2.18 & 25.07 & 91.55 & 34.68 & 64.51 & 63.55 & 0.99 & \CC{novelcolor}9.86 & 45.29 & 37.11 \\
         & \ourmethod \citep{riz2023novel} & \CC{novelcolor}0.52 & 12.06 & \CC{novelcolor}\textbf{33.89} & 73.28 & 75.53 & 33.35 & \CC{novelcolor}6.56 & 30.24 & 92.99 & 50.51 & 68.36 & 64.99 & 6.94 & \CC{novelcolor}13.66 & 50.83 & 42.25 \\
         & \newmethod  (Ours) & \CC{novelcolor}\textbf{0.93} & 6.29 & \CC{novelcolor}33.34 & 79.59 & 76.69 & 36.74 & \CC{novelcolor}\textbf{12.26} & 32.57 & 95.91 & 46.14 & 67.65 & 62.92 & 5.37 & \CC{novelcolor}\textbf{15.51} & 50.99 & 42.80 \\
        \midrule
        
        \multirow{2}{*}{S3DIS-$3^3$} & EUMS\dag \citep{zhao2022novel} & 0.02 & \CC{novelcolor}5.32 & 56.68 & 77.26 & 72.70 & 36.71 & 28.52 & 45.24 & 93.64 & \CC{novelcolor}4.41 & 69.21 & 59.09 & \CC{novelcolor}2.52 & \CC{novelcolor}4.08 & 59.91 & 42.41 \\
         & \ourmethod \citep{riz2023novel} & 0.12 & \CC{novelcolor}0.29 & 54.54 & 79.44 & 78.01 & 38.07 & 27.68 & 39.04 & 95.50 & \CC{novelcolor}\textbf{30.25} & 67.77 & 68.18 & \CC{novelcolor}\textbf{8.44} & \CC{novelcolor}\textbf{12.99} & 54.83 & 45.18 \\
         & \newmethod  (Ours) & 0.00 & \CC{novelcolor}\textbf{7.26} & 56.55 & 82.90 & 76.56 & 36.82 & 25.87 & 44.71 & 96.38 & \CC{novelcolor}20.23 & 65.61 & 66.91 & \CC{novelcolor}6.24 & \CC{novelcolor}11.24 & 55.23 & 45.08 \\

        \bottomrule
        \addlinespace[2.5pt]
         \multicolumn{10}{c}{} & \multirow{4}{*}{Avg} & \multicolumn{4}{|l|}{EUMS\dag \citep{zhao2022novel}} & \CC{novelcolor}9.38  & 39.87 & 32.60\\
         \multicolumn{10}{c}{} &  & \multicolumn{4}{|l|}{\ourmethod \citep{riz2023novel}} & \CC{novelcolor}20.79 & 45.48 & 39.65\\
         \multicolumn{10}{c}{} &  & \multicolumn{4}{|l|}{\newmethod  (Ours)} & \CC{novelcolor}\textbf{34.05} & 45.71 & 43.45\\
         \cmidrule[1pt]{11-18}
         \cmidrule[1pt]{11-18}
    \end{tabular}
    }
\end{table*}

\subsection{Quantitative analysis}
We evaluate \newmethod on both outdoor LiDAR datasets (SemanticPOSS~\citep{pan2020semanticposs} and SemanticKITTI~\citep{behley2019semantickitti}) and indoor RGB-D datasets (S3DIS~\citep{armeni20163d}). For each setting, we report the upper bound \textit{Full supervision} obtained by supervised training over both base and novel classes. We name with \textit{OpenScene$^\star$ $n$ Syn.} the zero-shot results achieved by OpenScene as described in Sec.~\ref{sec:zero-shot} using \textit{n} synonyms when building the ensembles. This baseline is our competitor for the performance achieved on novel classes. 
EUMS$^\dag$~(Sec.\ref{sec:adaptation_Zhao}) and \ourmethod~\citep{riz2023novel} are the NCD approaches that we directly compare against \newmethod.

\noindent
\textbf{Outdoor datasets.}
Tab.~\ref{tab:results_poss} and Tab.~\ref{tab:results_semantickitti} report the segmentation results on SemanticPOSS and SemanticKITTI, respectively.\\
On SemanticPOSS, \newmethod achieves $30.05$ IoU on novel classes, improving of $+15.11$ IoU over EUMS$^\dag$ and $+8.65$ IoU over \ourmethod (Tab.~\ref{tab:results_poss}).
\newmethod outperforms the other methods on all the four dataset splits and on all the classes, except for \textit{bike}, \textit{person}, and \textit{pole}, where \ourmethod achieves better results. We attribute the significant decline in performance observed in \newmethod for the \textit{bike} class to the alignment procedure with the auxiliary zero-shot model, since the auxiliary zero-shot network exhibits notably poor results on this particular class (0.06 IoU). We consider the decrease in performance for the other two classes (i.e. \textit{person} and \textit{pole)} as simple fluctuations that may happen when \newmethod organizes its feature space differently from the one of \ourmethod.
\newmethod improves over the reference \textit{OpenScene$^\star$} baseline of $+7.00$ IoU, outperforming it on all classes, apart for \textit{car}, \textit{plant} and \textit{trashcan}.
Interestingly, the \textit{OpenScene$^\star$} setting outperforms even the \textit{Full supervision} upper bound on the \textit{car} class.\\
On SemanticKITTI, \newmethod achieves $26.39$ IoU on novel classes, improving of $+9.34$ IoU over EUMS$^\dag$ and $+3.55$ IoU over \ourmethod (Tab.~\ref{tab:results_semantickitti}).
\newmethod outperforms all the compared approaches on all the SemanticKITTI splits, showing a large improvement on novel classes, \textit{e.g.}, \textit{building} and \textit{sidewalk}. 
Again, \newmethod improves over the reference \textit{OpenScene$^\star$} baseline of $+7.11$ IoU, outperforming it on 14 out of 19 classes, surpassing it with a large margin in the classes \textit{bicyclist}, \textit{car}, and \textit{traffic-sign}.
Interestingly, \newmethod outperforms the \textit{Full supervision} upper bound on the \textit{traffic-sign} class, with an improvement of $+1.81$ IoU.

\begin{figure*}[ht]
\centering
    \begin{tabular}{ccccc}
    \raggedright
        \begin{overpic}[width=0.2\textwidth]{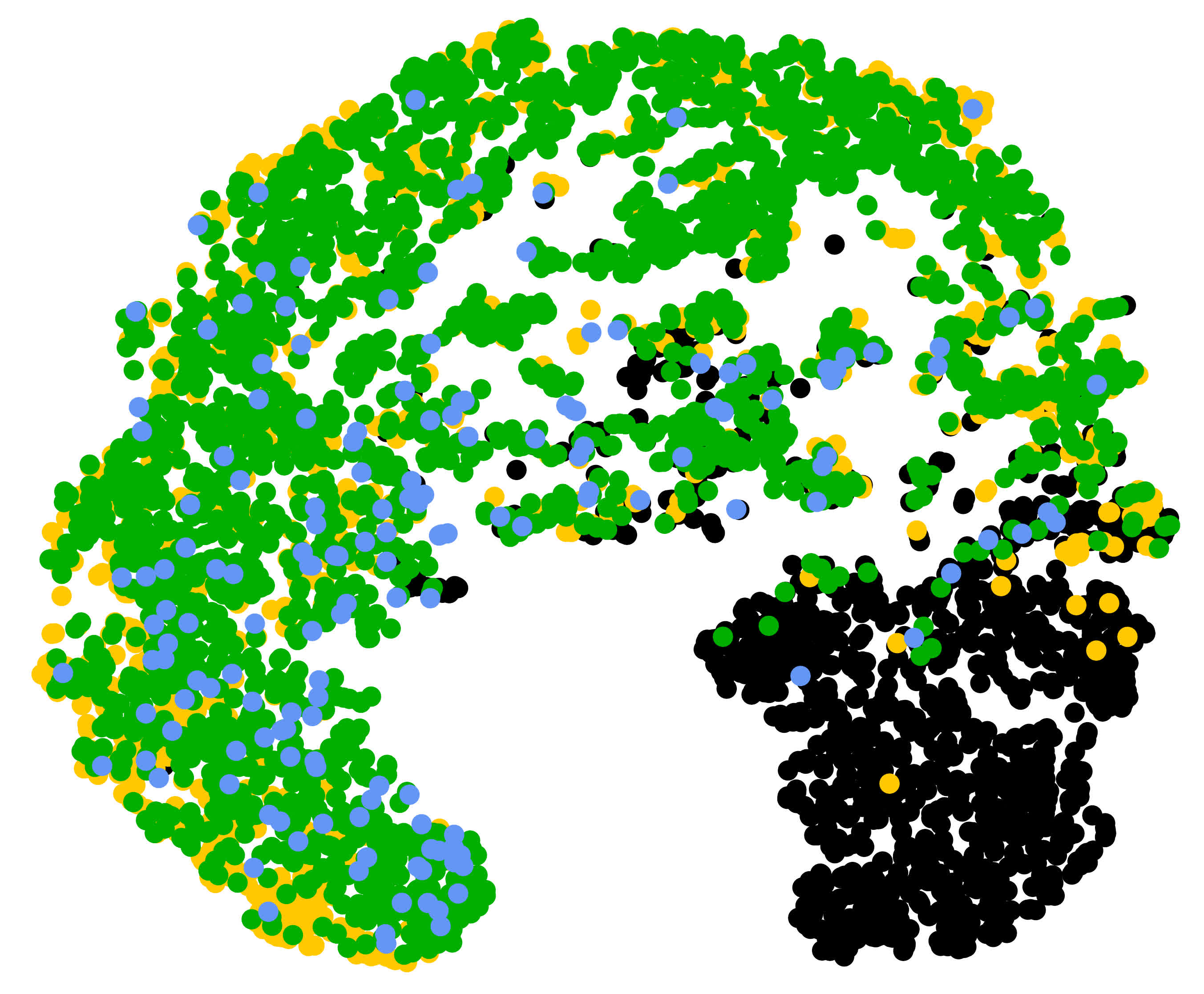}
            \put(0,85){\color{black}\footnotesize \textbf{\ourmethod~\citep{riz2023novel}}}
            \put(-10, 30){\small\rotatebox{90}{\color{black}\footnotesize \textbf{POSS-$4^0$}}}
        \end{overpic}
        & 
        \begin{overpic}[width=0.2\textwidth]{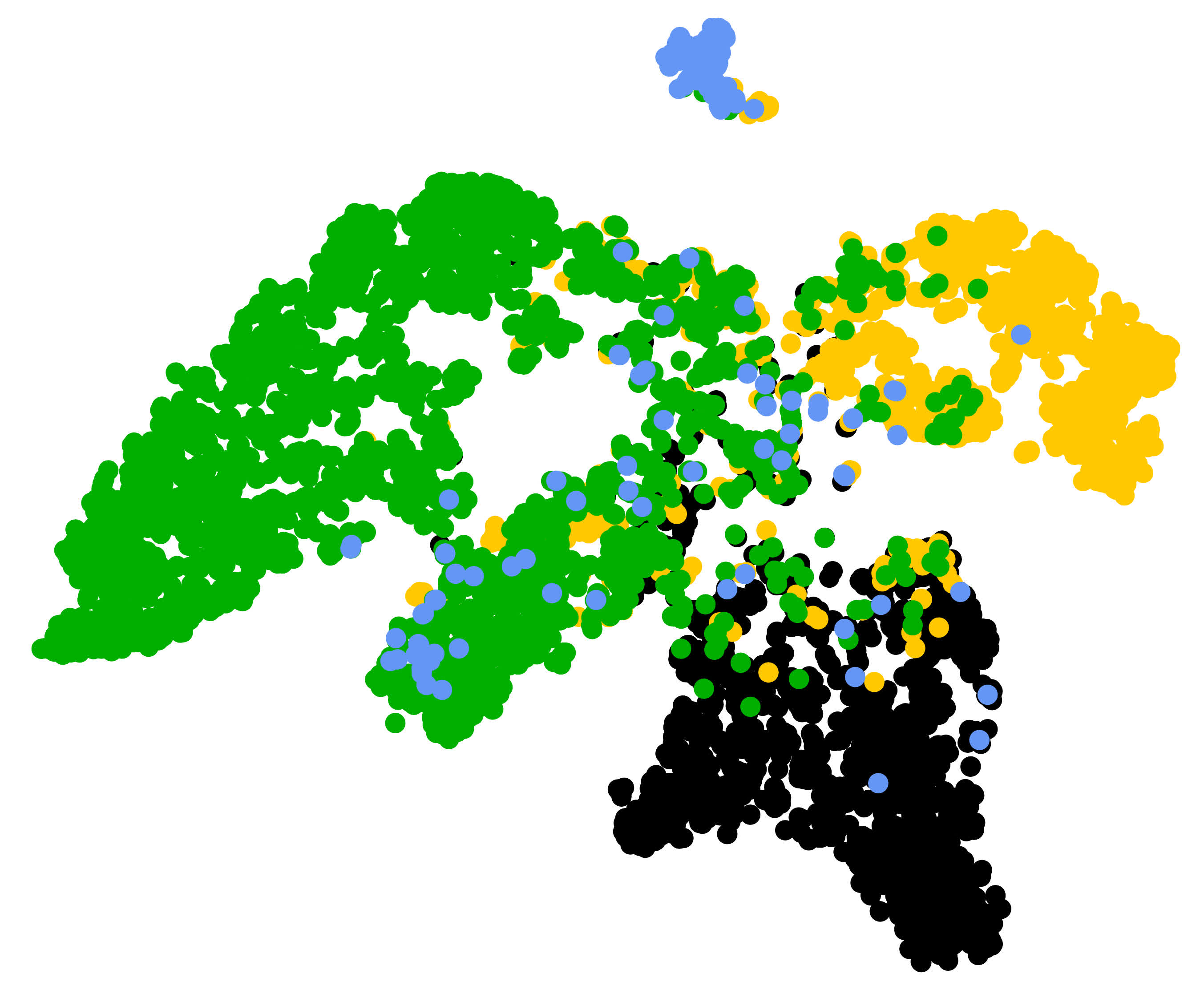}
            \put(25,85){\color{black}\footnotesize \textbf{\newmethod~(Ours)}}
        \end{overpic}
        &
        \hfill
        &
        \begin{overpic}[width=0.2\textwidth]{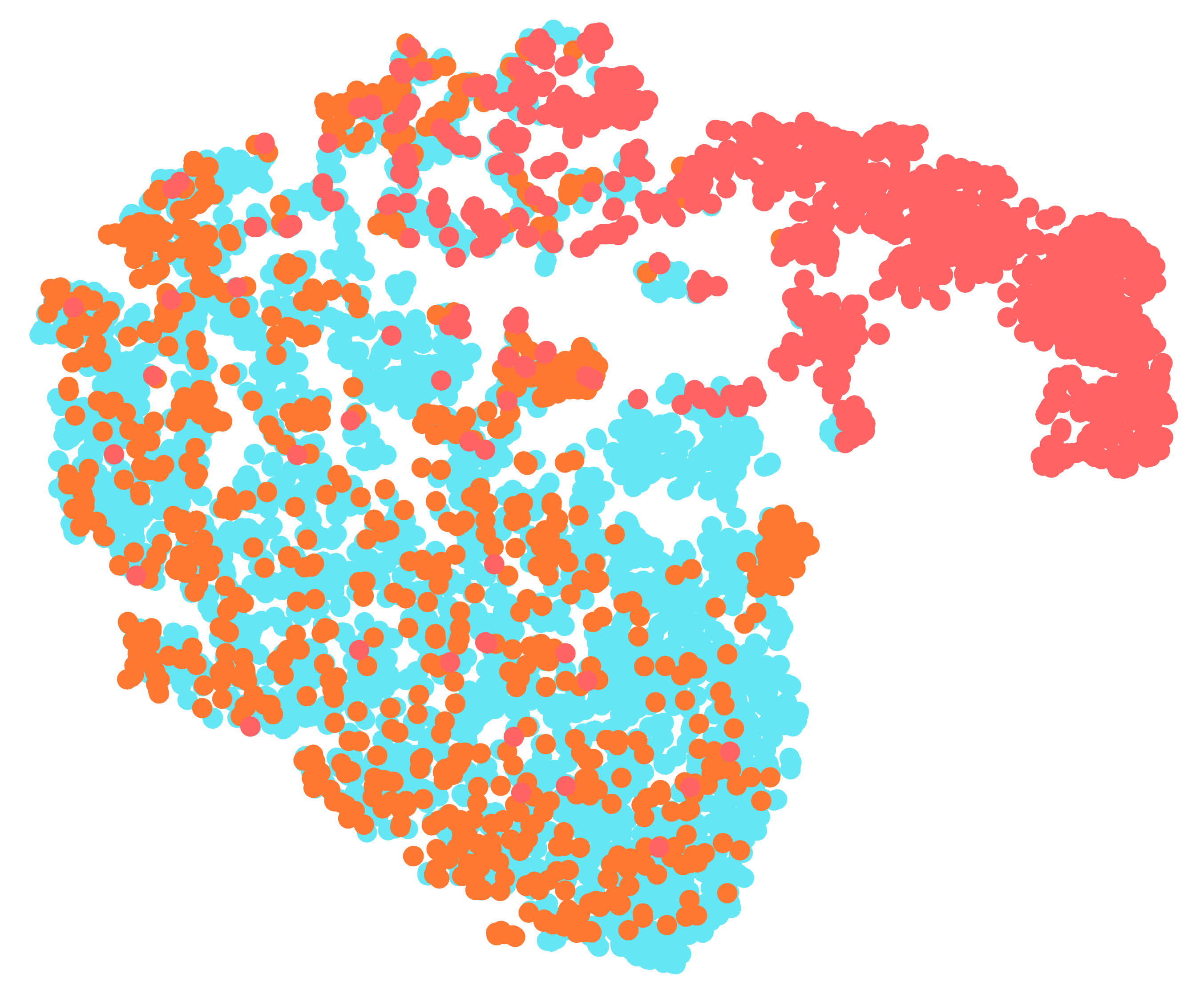}
            \put(0,85){\color{black}\footnotesize \textbf{\ourmethod~\citep{riz2023novel}}}
            \put(-10, 30){\small\rotatebox{90}{\color{black}\footnotesize \textbf{POSS-$3^1$}}}
        \end{overpic}
        & 
        \begin{overpic}[width=0.2\textwidth]{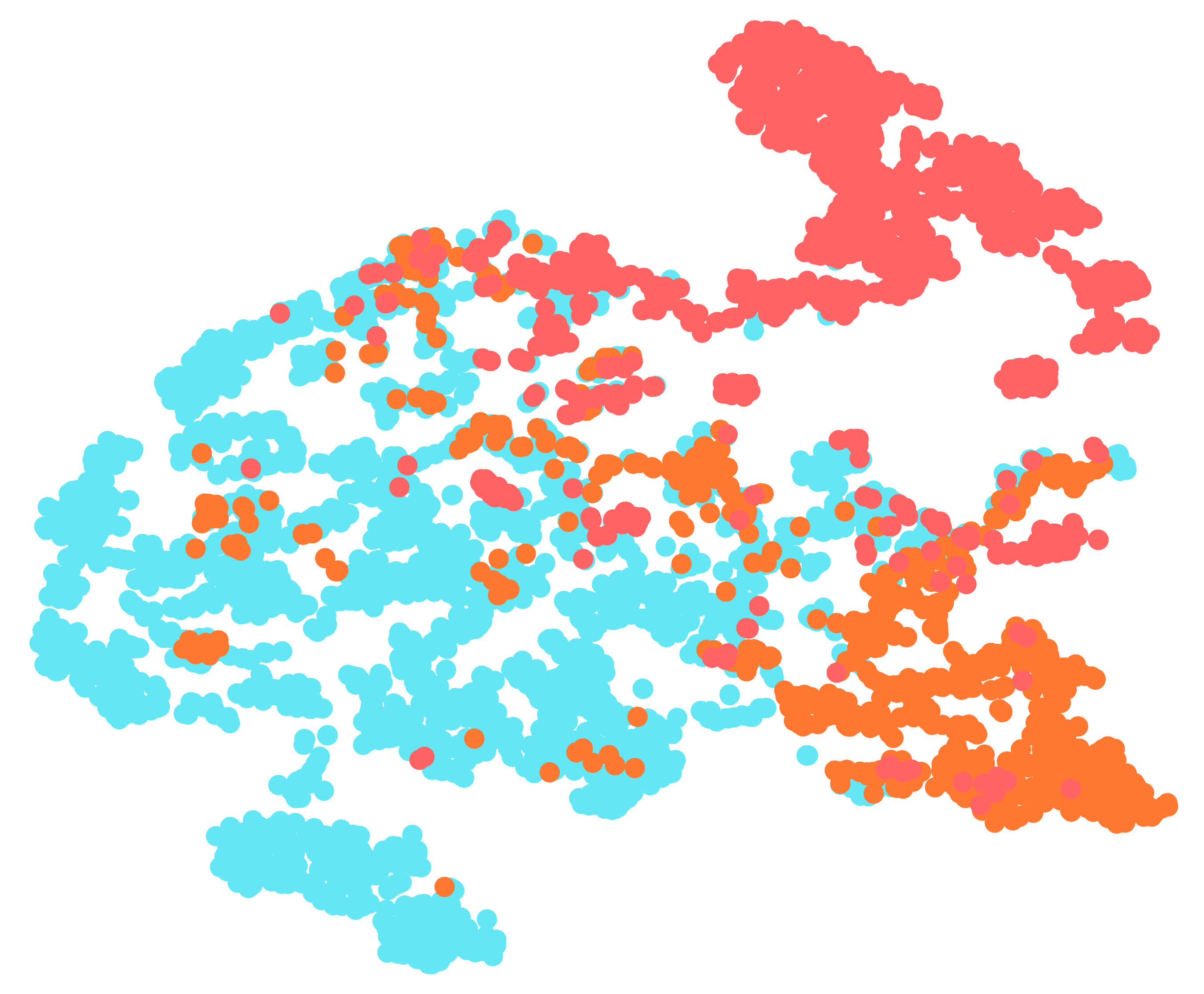}
            \put(25,85){\color{black}\footnotesize \textbf{\newmethod~(Ours)}}
        \end{overpic}
        \\
        \multicolumn{5}{c}{
        \begin{overpic}[width=0.99\textwidth]{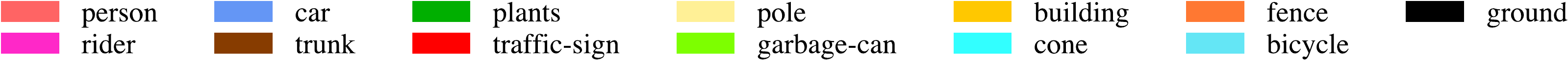}
        \end{overpic}}
        \\
        \begin{overpic}[width=0.2\textwidth]{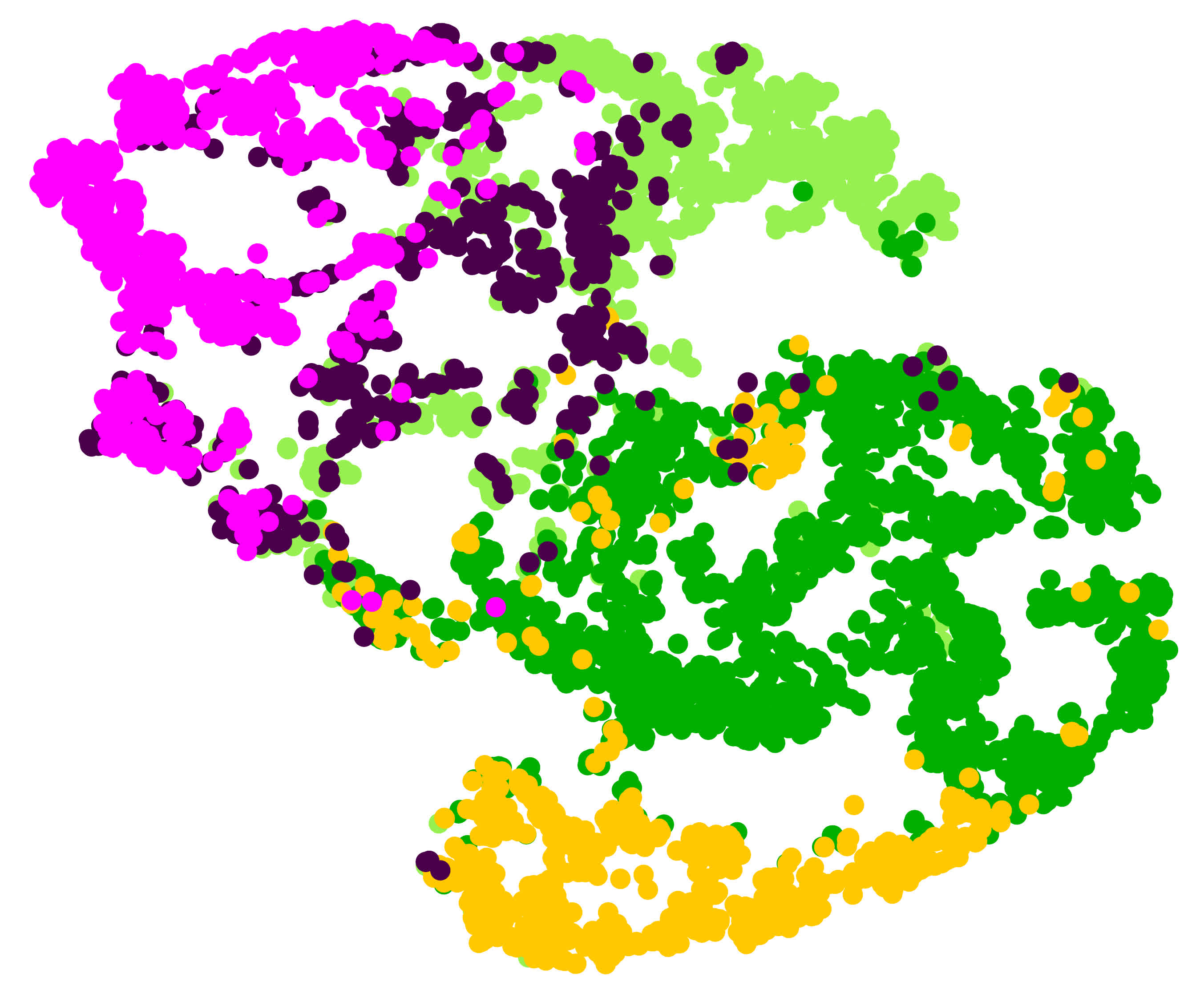}
            \put(-10, 30){\small\rotatebox{90}{\color{black}\footnotesize \textbf{KITTI-$5^0$}}}
        \end{overpic}
        & 
        \begin{overpic}[width=0.2\textwidth]{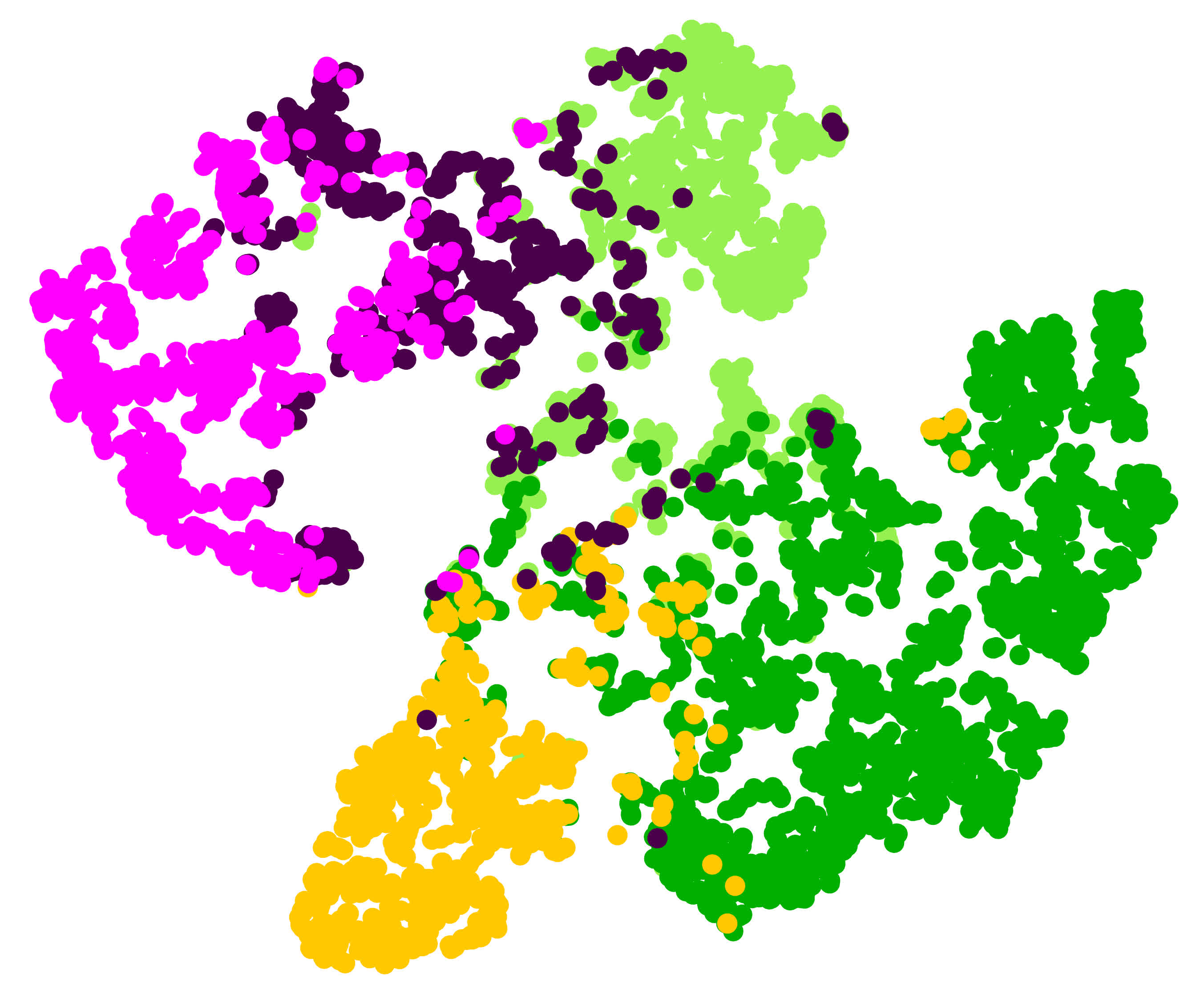}
        \end{overpic}
        &
        \hfill
        &
        \begin{overpic}[width=0.2\textwidth]{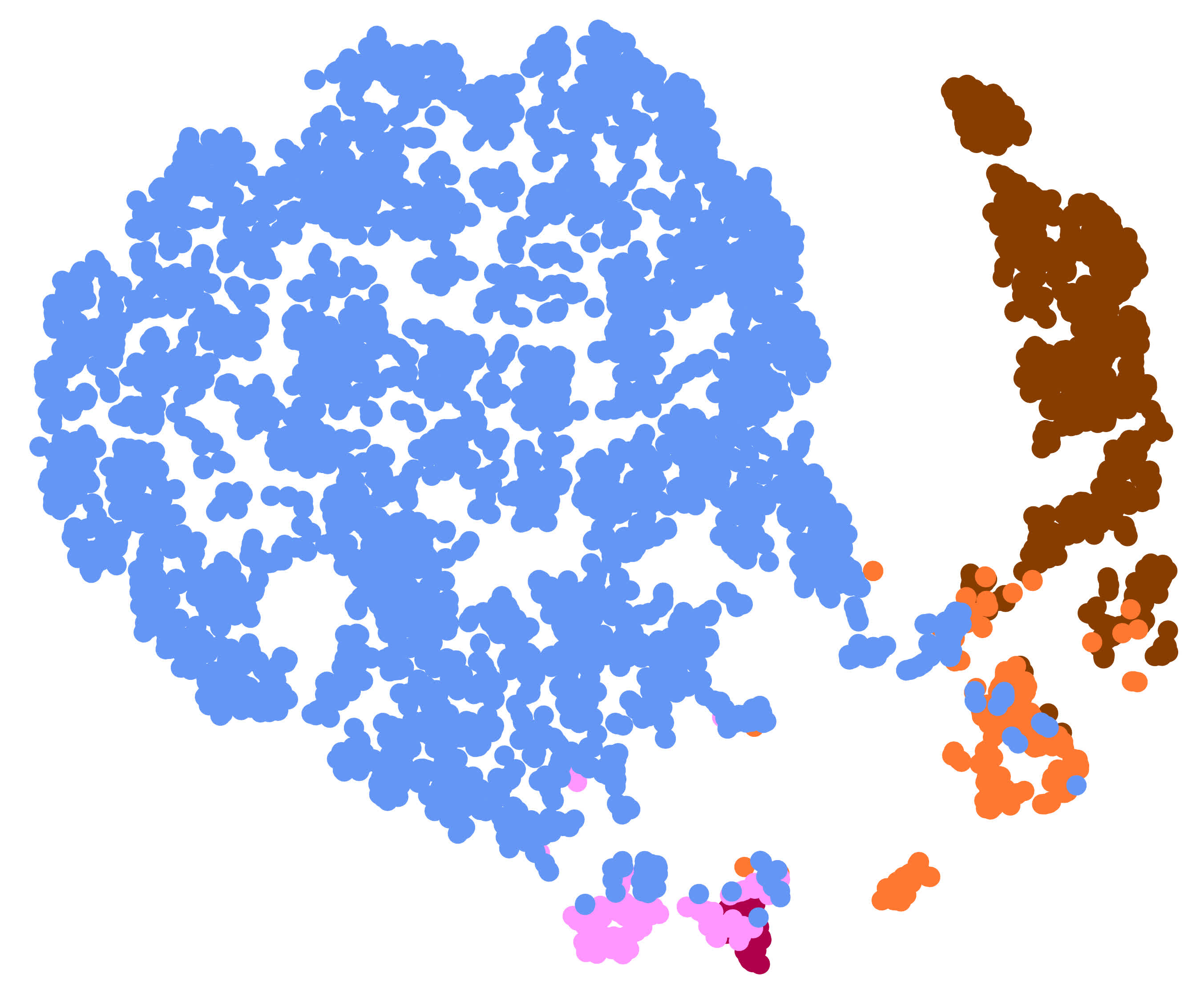}
            \put(-10, 30){\small\rotatebox{90}{\color{black}\footnotesize \textbf{KITTI-$5^1$}}}
        \end{overpic}
        & 
        \begin{overpic}[width=0.2\textwidth]{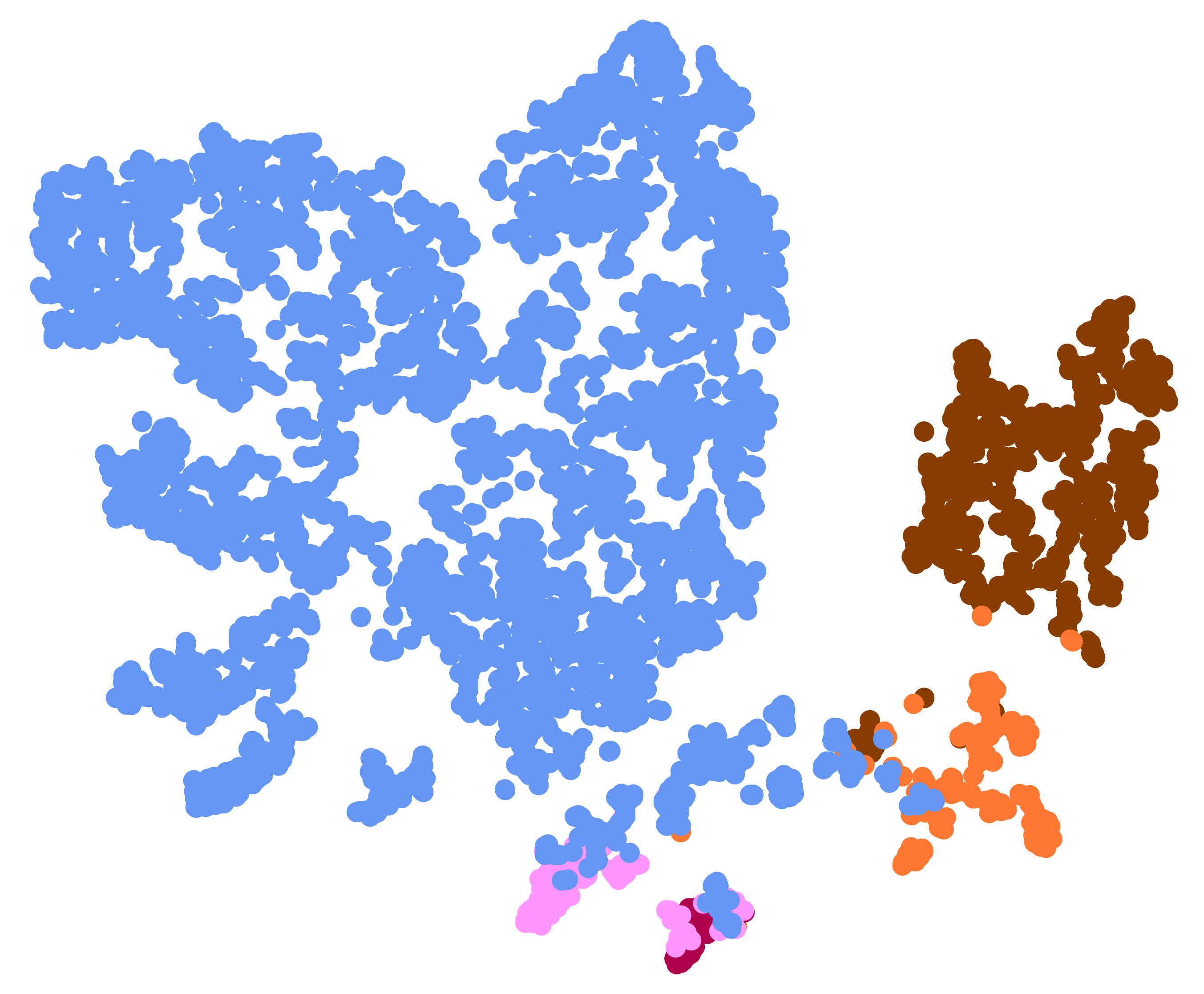}
        \end{overpic}
        \\
        \multicolumn{5}{c}{
        \begin{overpic}[width=0.99\textwidth]{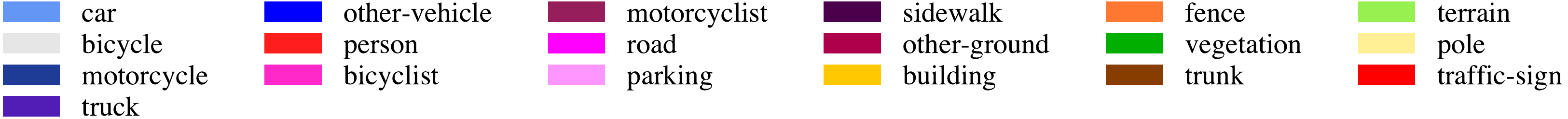}
        \end{overpic}}
        \\
        \begin{overpic}[width=0.2\textwidth]{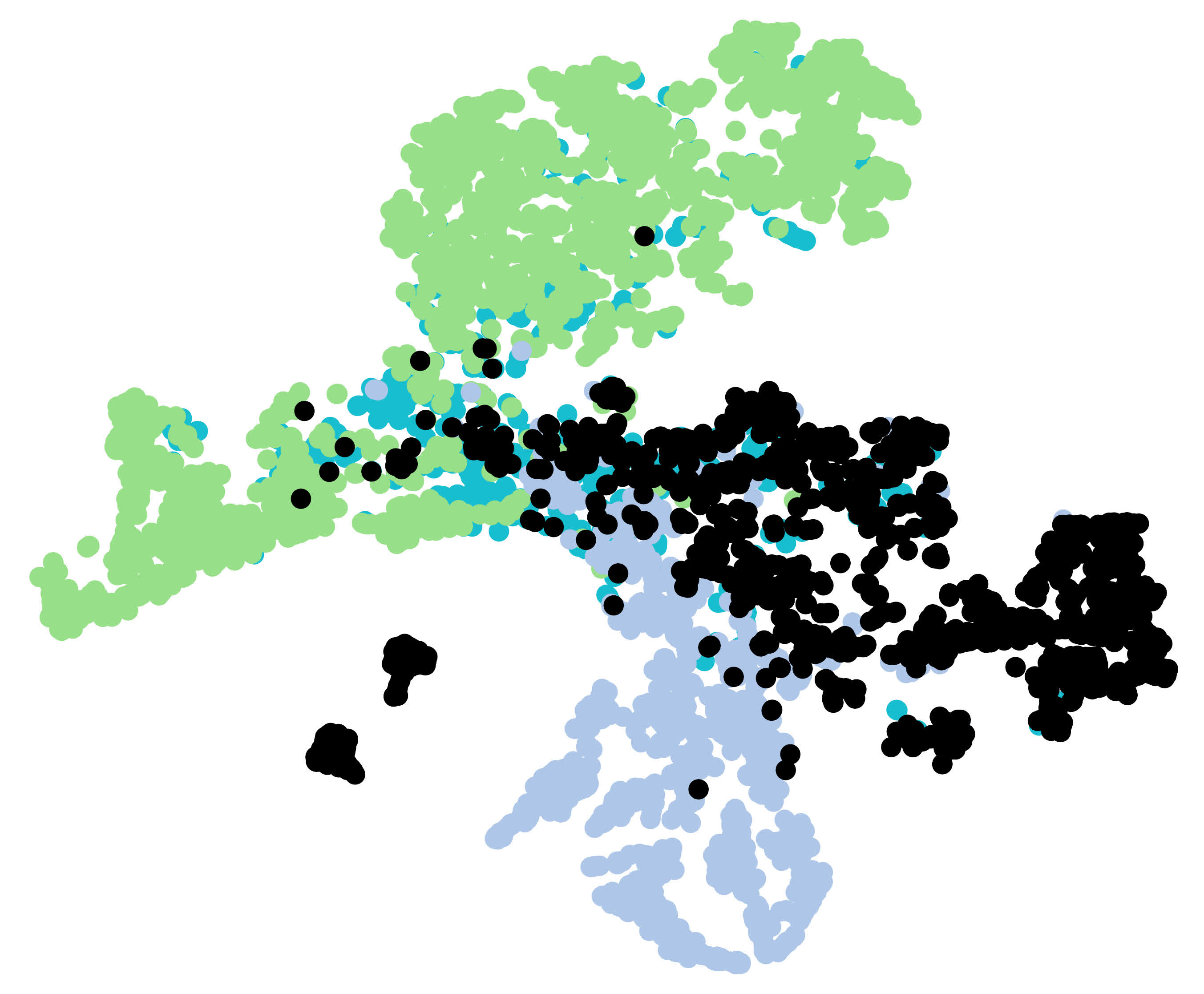}
            \put(-10, 30){\small\rotatebox{90}{\color{black}\footnotesize \textbf{S3DIS-$4^0$}}}
        \end{overpic}
        & 
        \begin{overpic}[width=0.2\textwidth]{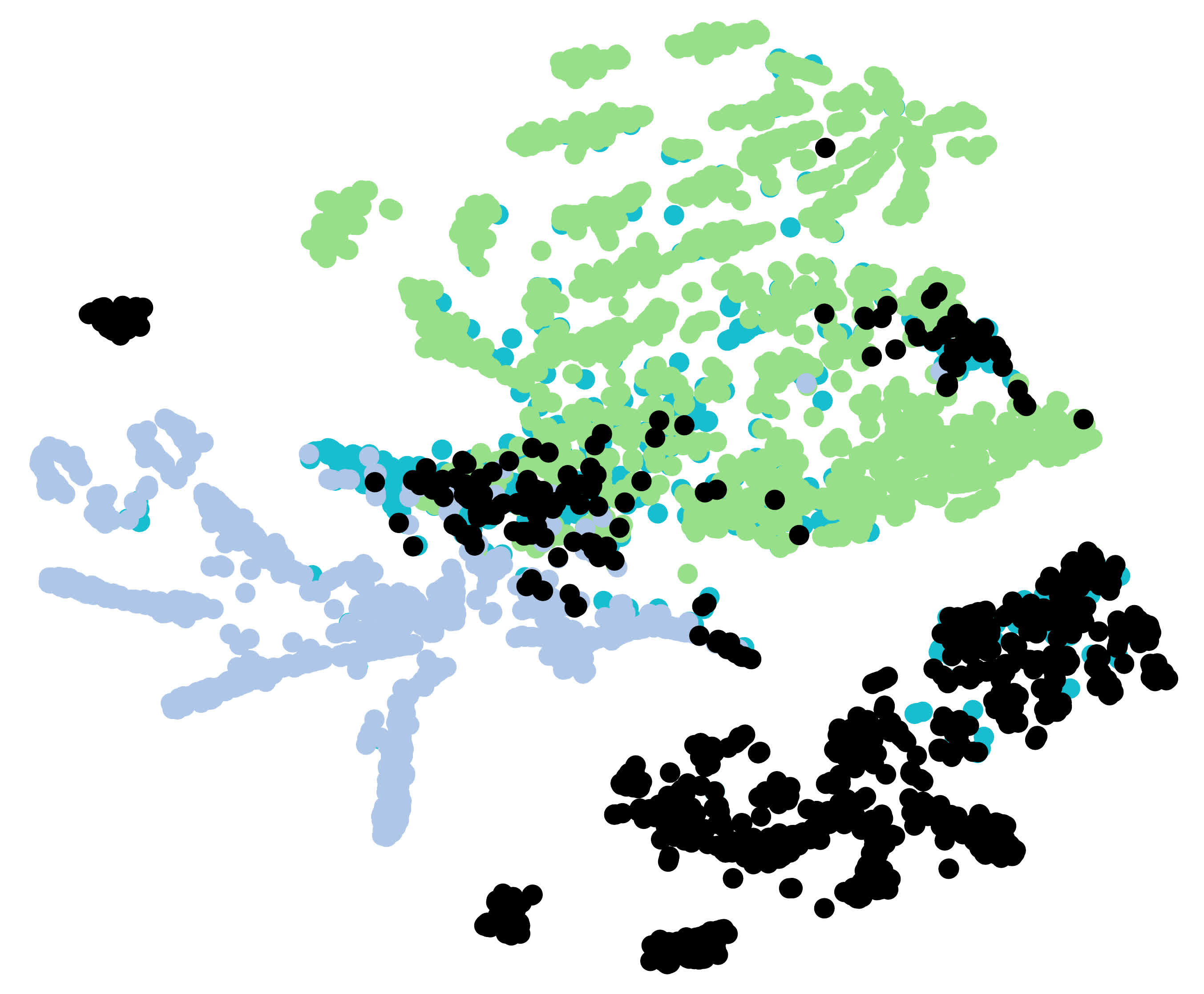}
        \end{overpic}
        &
        \hfill
        &
        \begin{overpic}[width=0.2\textwidth]{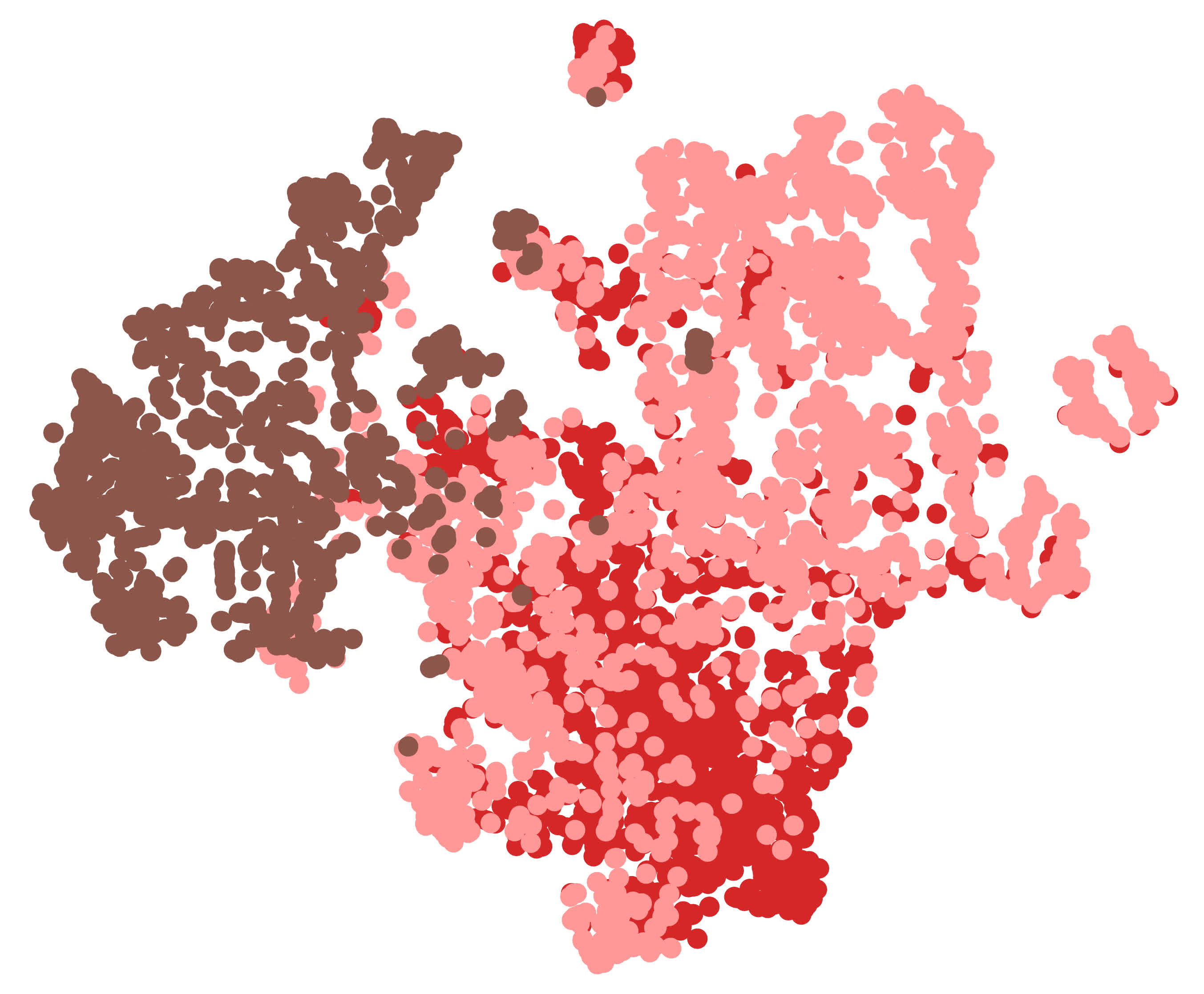}
            \put(-10, 30){\small\rotatebox{90}{\color{black}\footnotesize \textbf{S3DIS-$3^1$}}}
        \end{overpic}
        & 
        \begin{overpic}[width=0.2\textwidth]{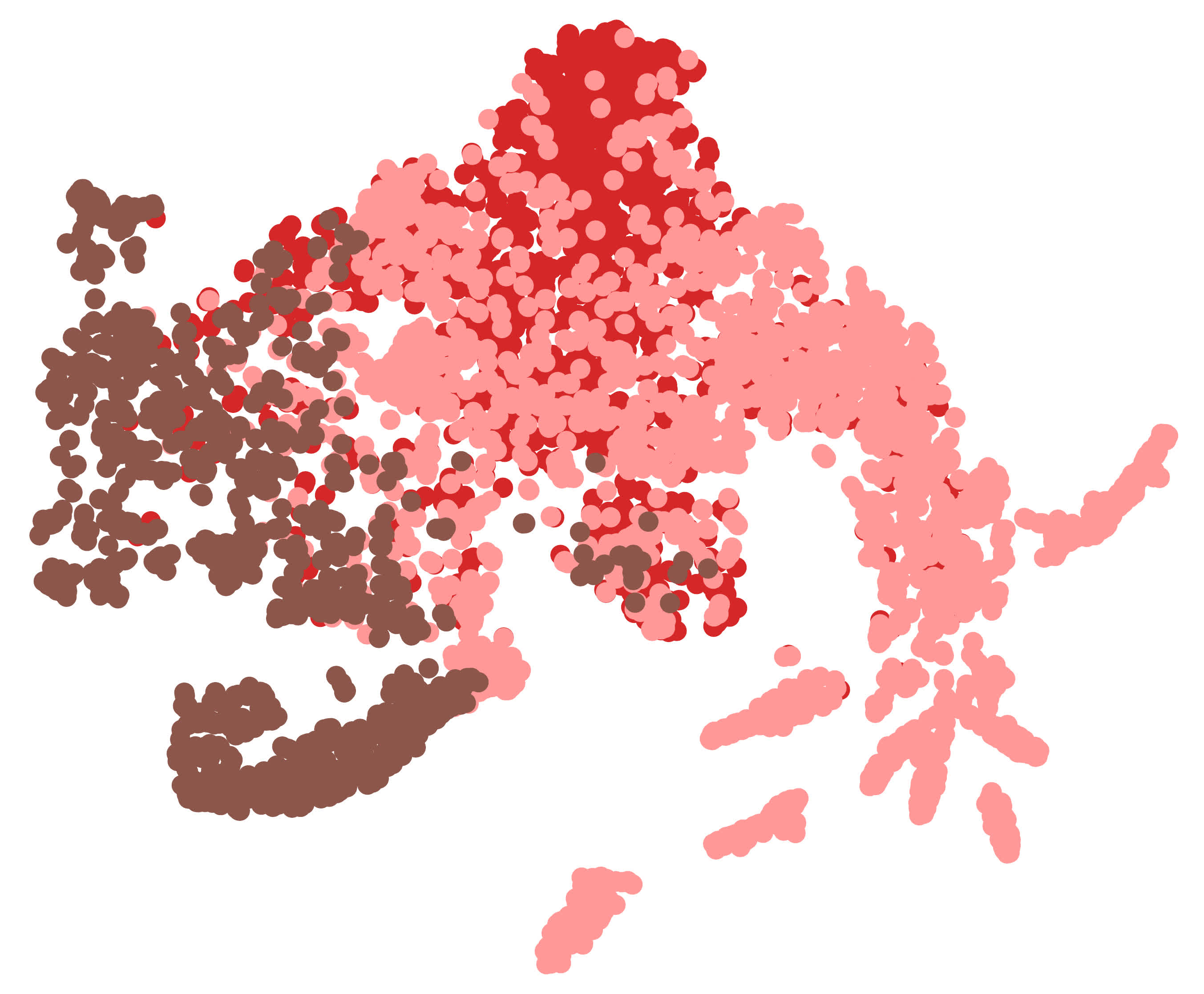}
        \end{overpic}
        \\
        \multicolumn{5}{c}{
        \begin{overpic}[width=0.99\textwidth]{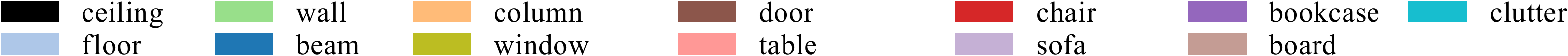}
        \end{overpic}}
    \end{tabular}
    \caption{t-SNE dimensionality reduction of the embedding space output by the feature extractors $f_\xi$ of \ourmethod and \newmethod for the novel points in different splits of our datasets. Compared to \ourmethod, \newmethod is able to better organize its embedding space, better grouping features of the novel classes in more compact and separated clusters.}
    \label{fig:t-SNE}
\end{figure*}

\noindent
\textbf{Indoor dataset.}
Tab.~\ref{tab:results_s3dis} reports the results on the indoor S3DIS dataset.
In this settings, \newmethod achieves $34.05$ IoU on novel classes, improving of $+24.67$ IoU over EUMS$^\dag$ and $+13.26$ IoU over \ourmethod.
\newmethod outperforms by a large margin EUMS$^\dag$ on all four splits. Compared to \ourmethod, it improves on three out of four splits, with the remarkably large margins of $+29.86$ IoU and $+23.08$ IoU on S3DIS-$4^0$ and S3DIS-$3^1$, respectively.
Considering the average performance over base and novel classes, \newmethod notably surpasses the results obtained by \textit{Full supervision} on two splits (S3DIS-$3^1$ and S3DIS-$3^3$), with $43.45$ IoU in average over the four splits (only $-0.98$ as compared to \textit{Full supervision}).

\begin{figure*}[t]
\centering
    \setlength\tabcolsep{2.5pt}
    \begin{tabular}{cccc}
    \raggedright
        \begin{overpic}[width=0.24\textwidth]{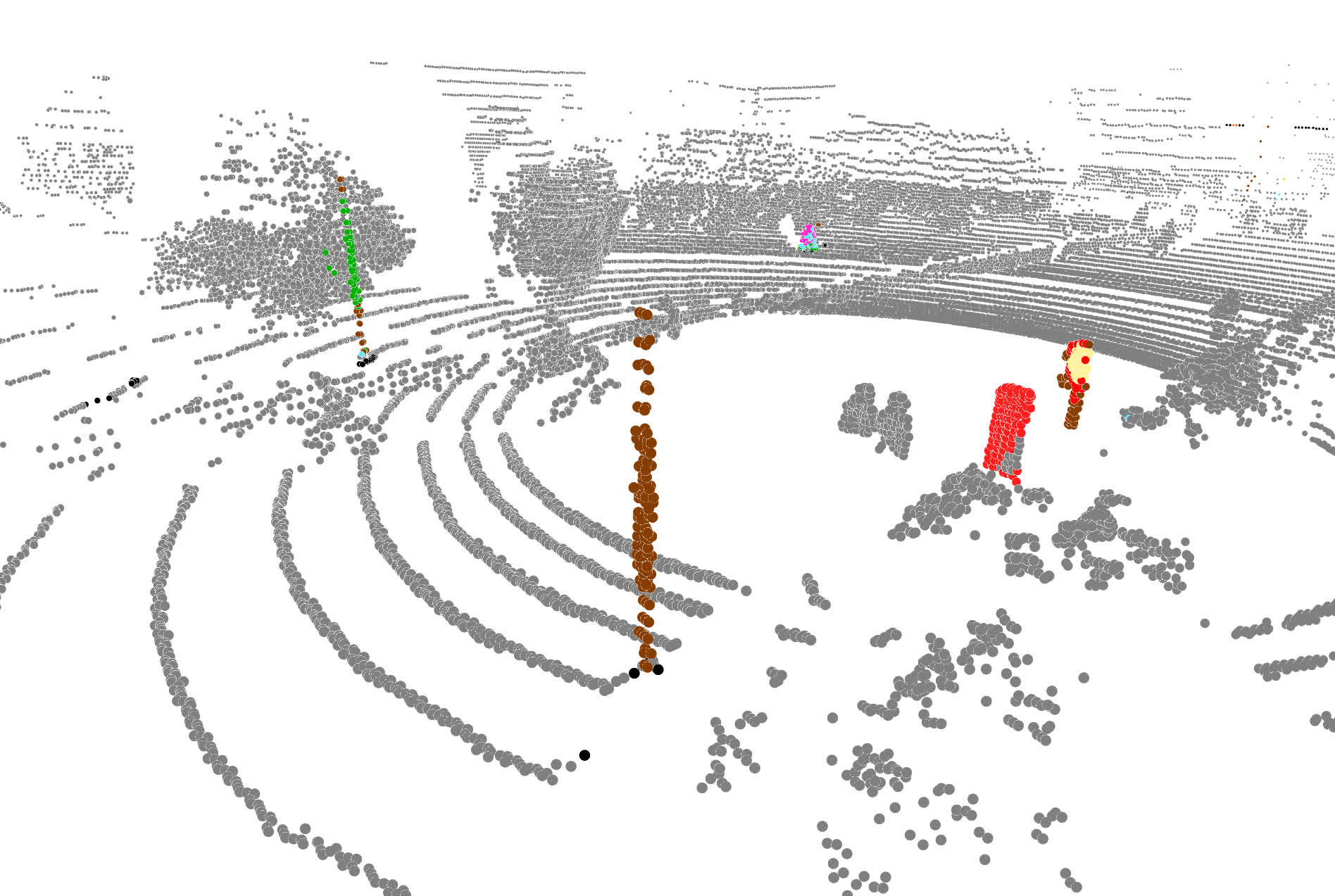}
        \put(0, 70){\color{black}\footnotesize \textbf{EUMS$^\dag$~\citep{zhao2022novel}}}
        \put(112,70){\color{black}\footnotesize \textbf{\ourmethod~\citep{riz2023novel}}}
        \put(235,70){\color{black}\footnotesize \textbf{\newmethod (Ours)}}
        \put(360,70){\color{black}\footnotesize \textbf{GT}}
        \put(-10, 20){\small\rotatebox{90}{\color{black}\footnotesize \textbf{POSS-$3^2$}}}
        \end{overpic} &  
        \begin{overpic}[width=0.24\textwidth]{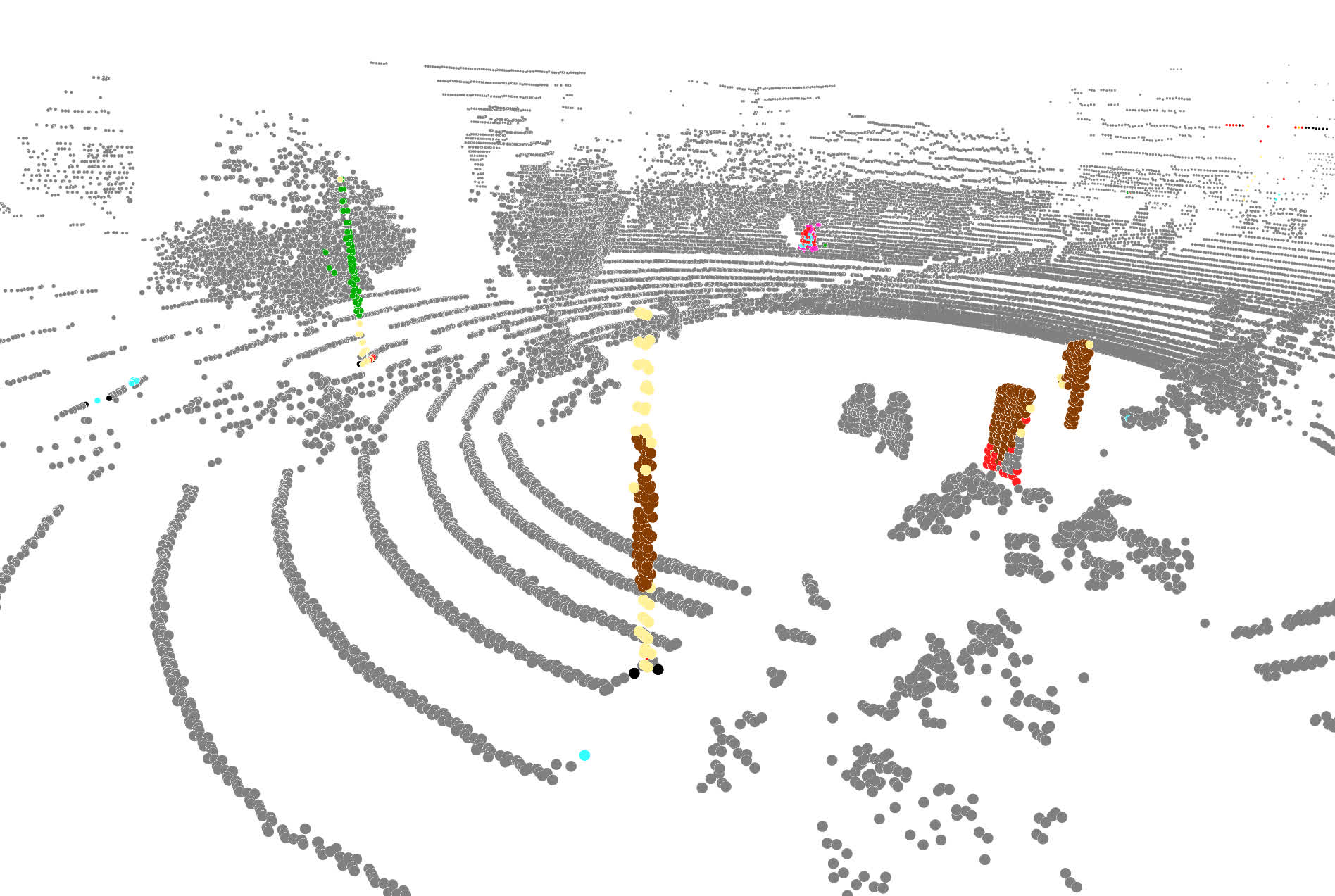}
        \end{overpic} &
        \begin{overpic}[width=0.24\textwidth]{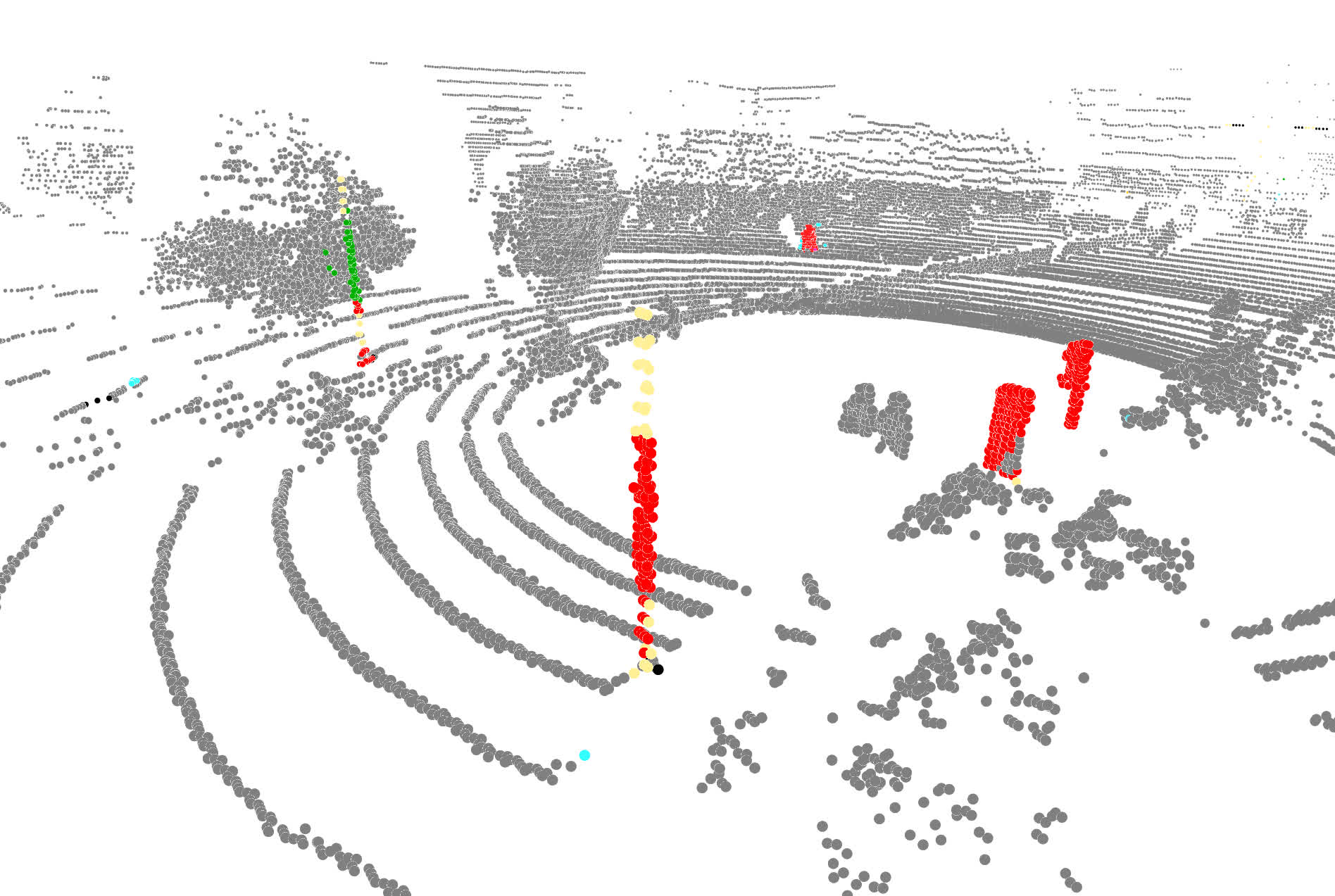}
        \end{overpic} &
        \begin{overpic}[width=0.24\textwidth]{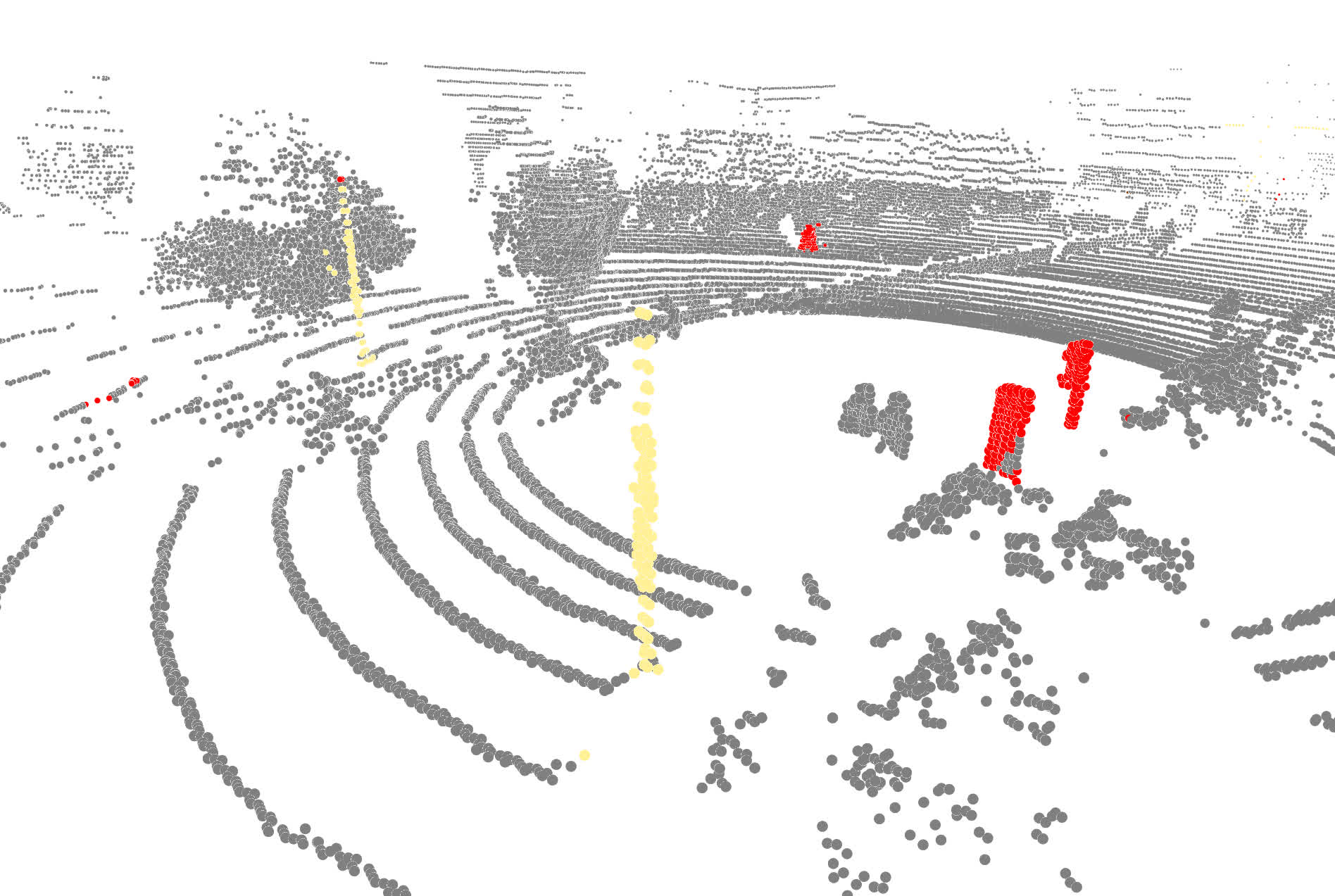}
        \end{overpic}
        \\
        \multicolumn{4}{c}{
        \begin{overpic}[width=0.99\textwidth]{_images_qualitative_poss_legend_poss.pdf}
        \end{overpic}}
        \\
        \begin{overpic}[width=0.24\textwidth]{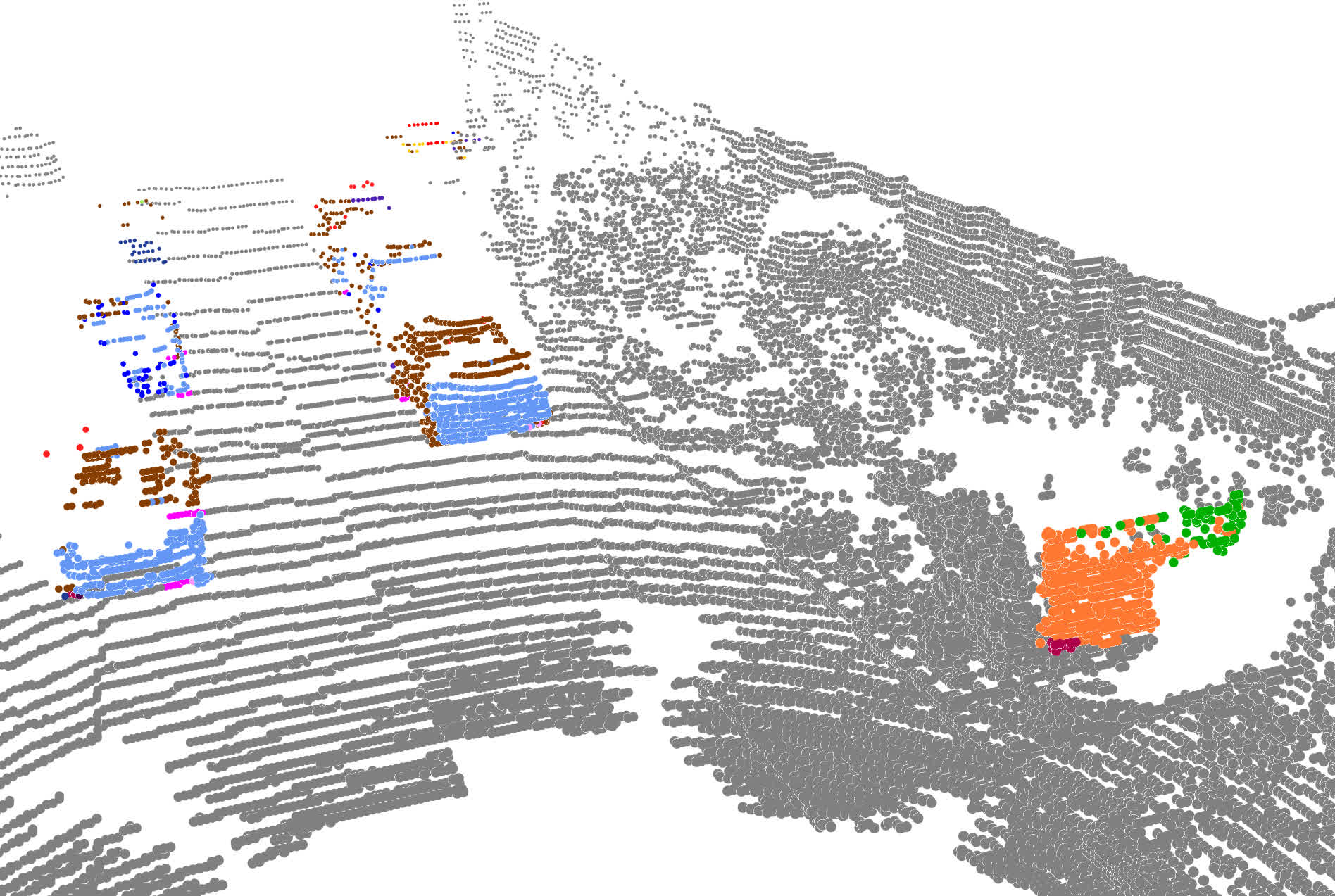}
        \put(-10, 20){\small\rotatebox{90}{\color{black}\footnotesize \textbf{KITTI-$5^1$}}}
        \end{overpic} &  
        \begin{overpic}[width=0.24\textwidth]{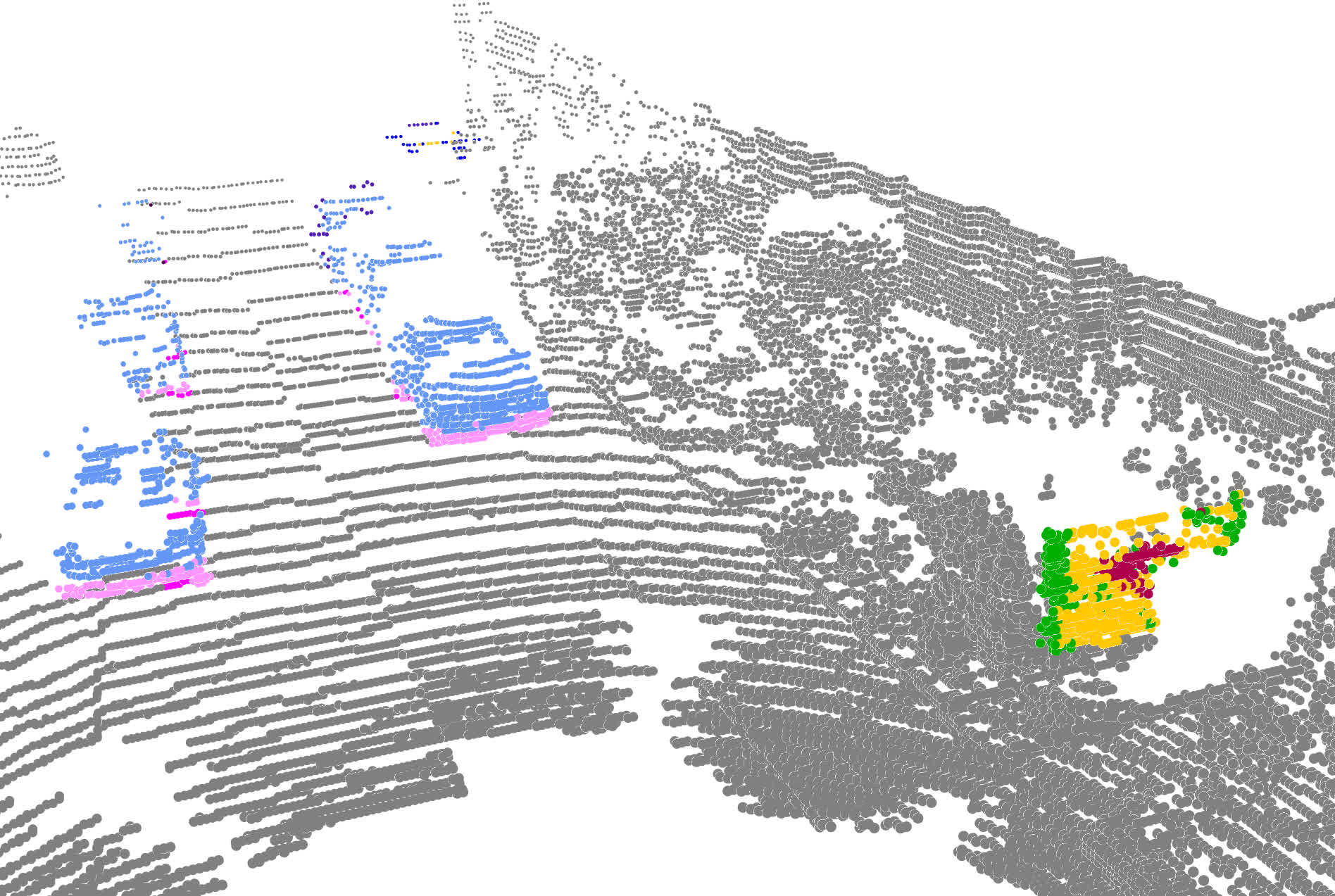}
        \end{overpic} &
        \begin{overpic}[width=0.24\textwidth]{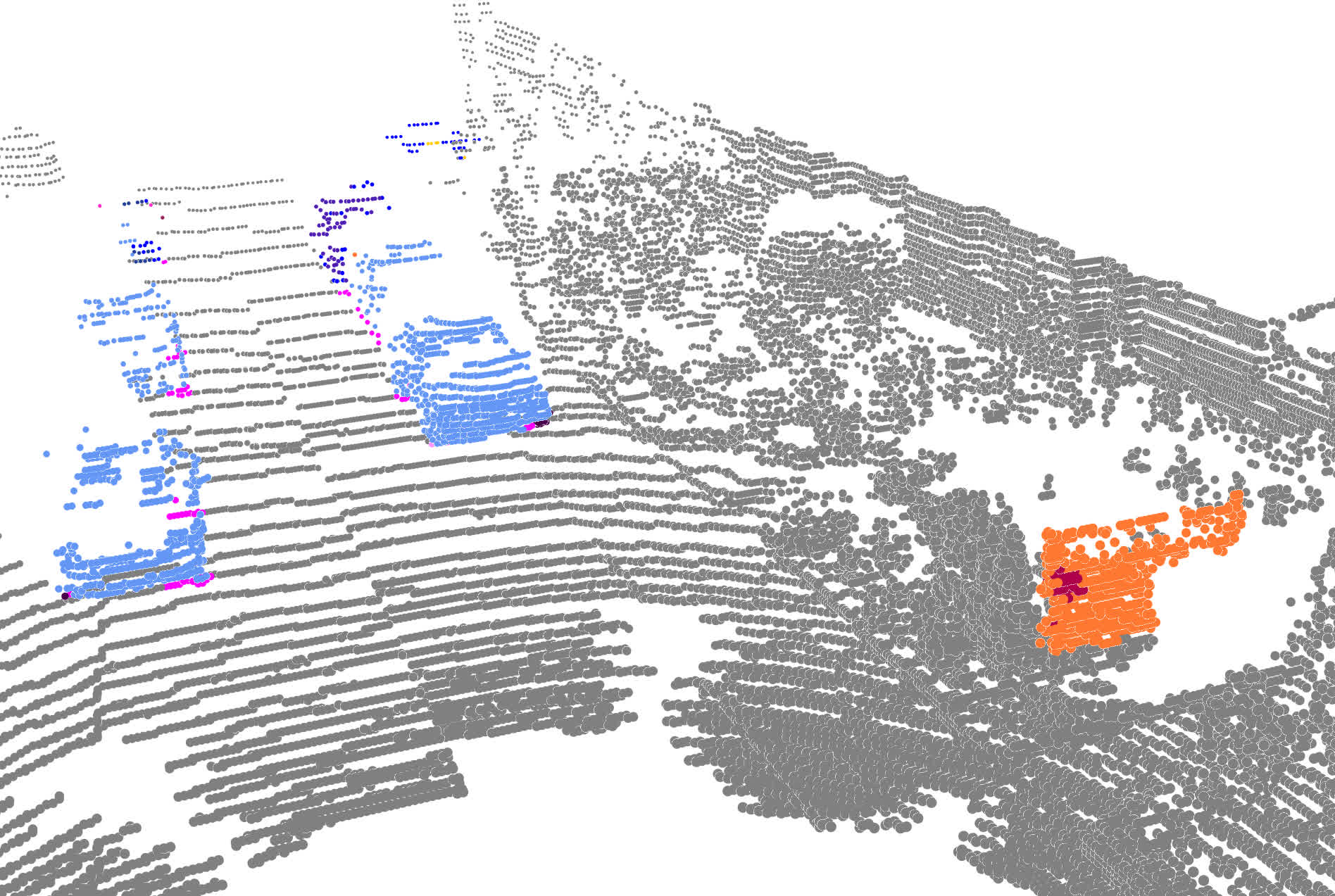}
        \end{overpic} &
        \begin{overpic}[width=0.24\textwidth]{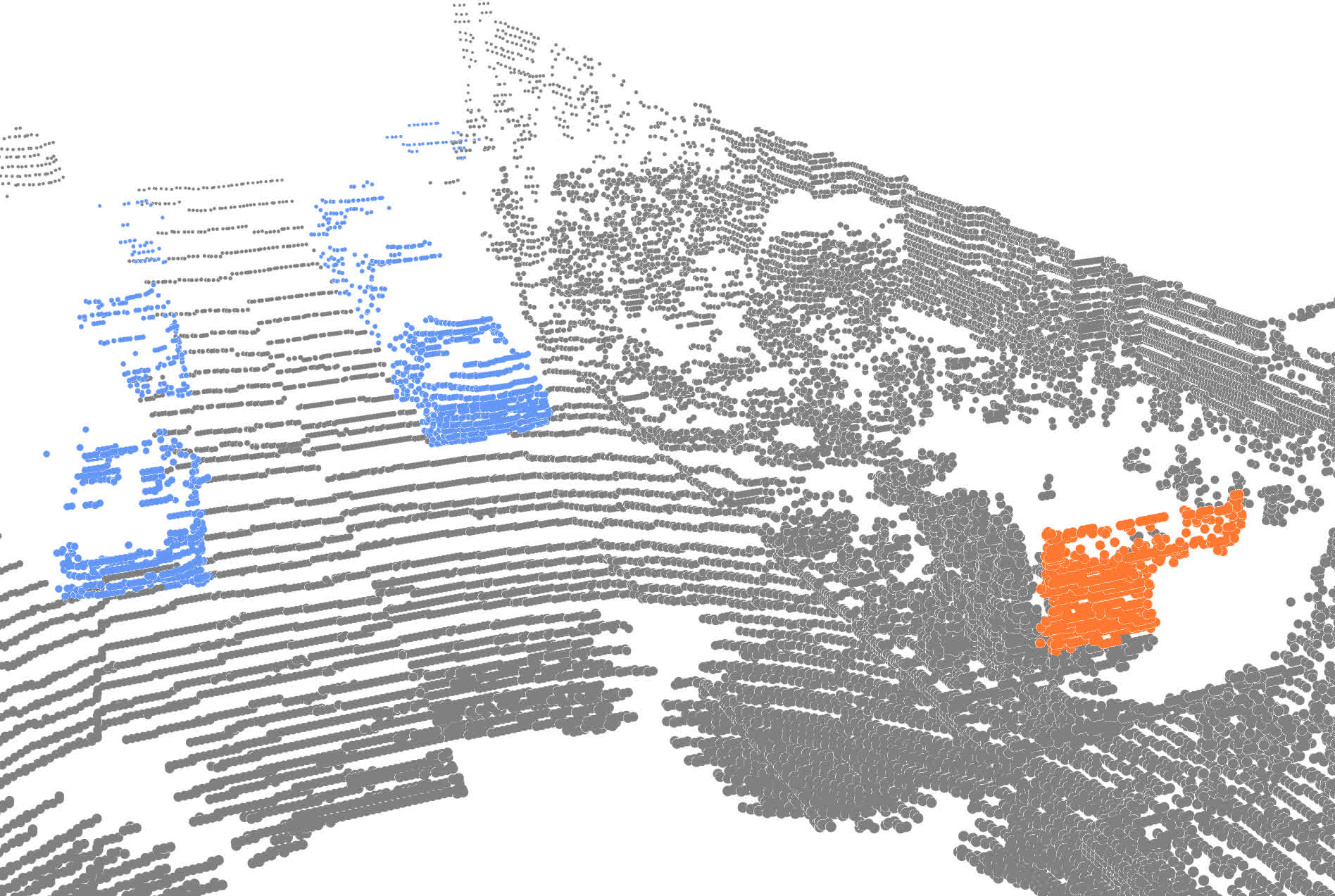}
        \end{overpic}
        \\
        \multicolumn{4}{c}{
        \begin{overpic}[width=0.99\textwidth]{_images_qualitative_kitti_legend_kitti.pdf}
        \end{overpic}}
        \\
        \begin{overpic}[width=0.24\textwidth]{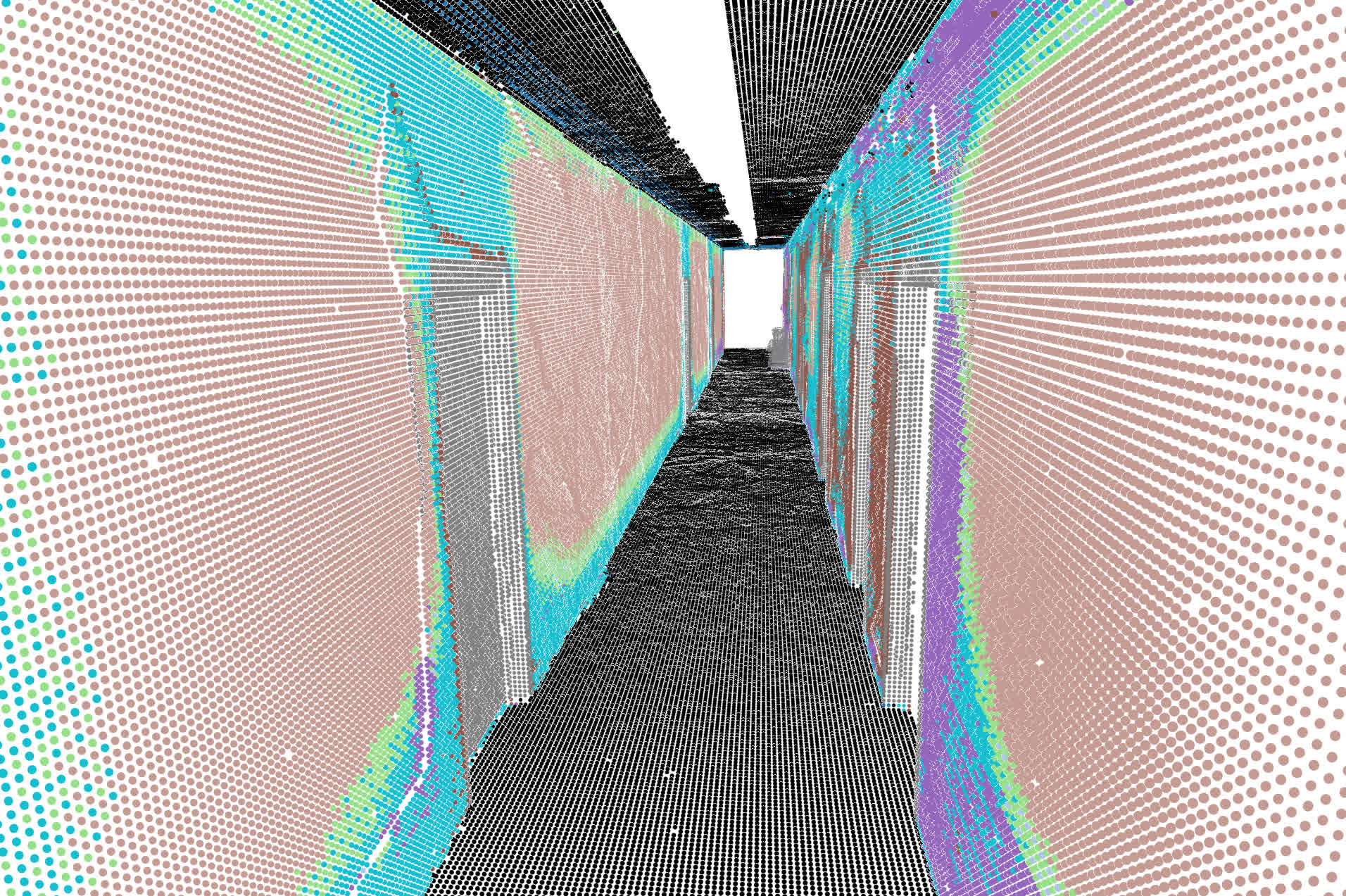}
        \put(-10, 20){\small\rotatebox{90}{\color{black}\footnotesize \textbf{S3DIS-$4^0$}}}
        \end{overpic} &  
        \begin{overpic}[width=0.24\textwidth]{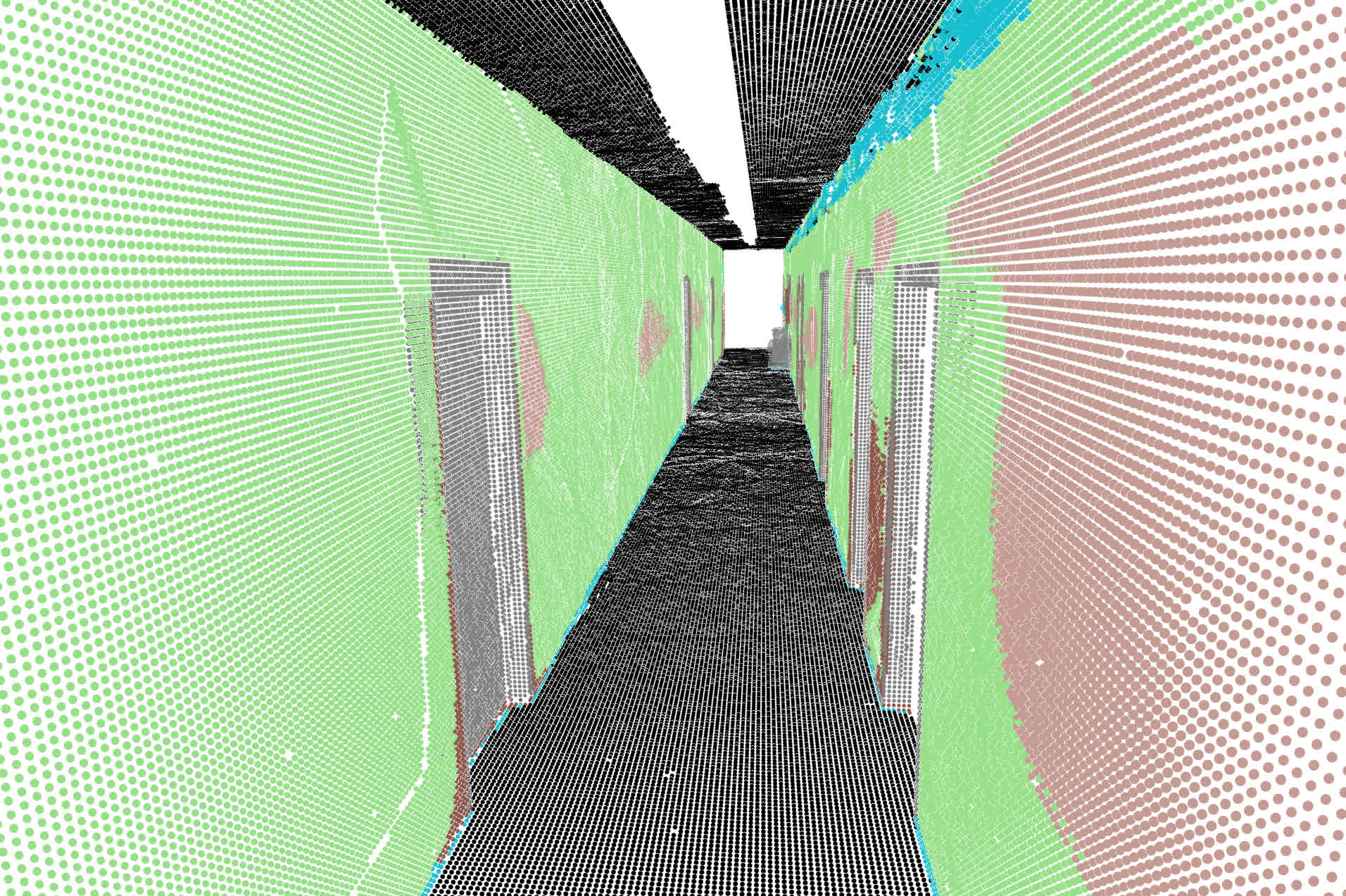}
        \end{overpic} &
        \begin{overpic}[width=0.24\textwidth]{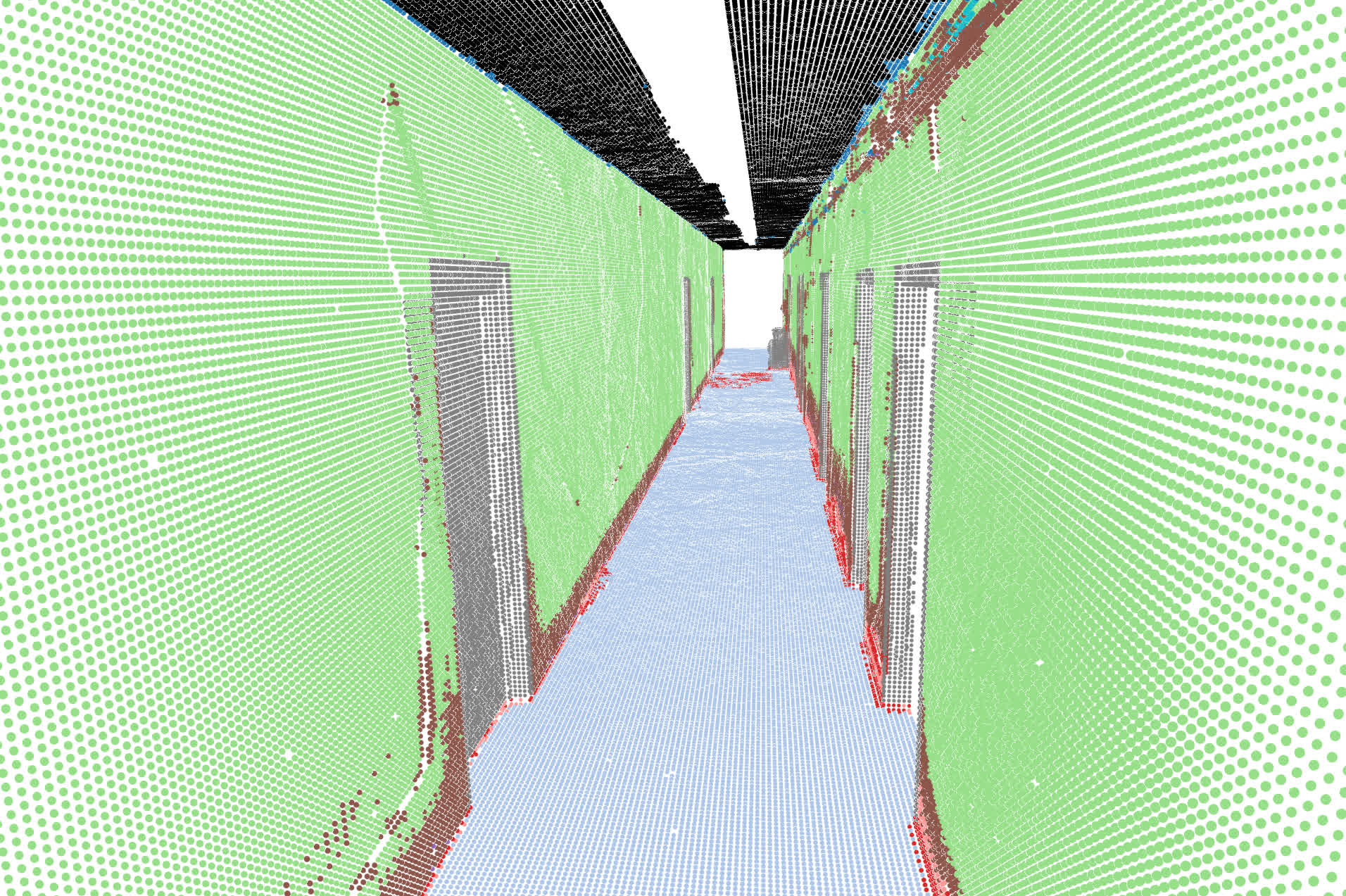}
        \end{overpic} &
        \begin{overpic}[width=0.24\textwidth]{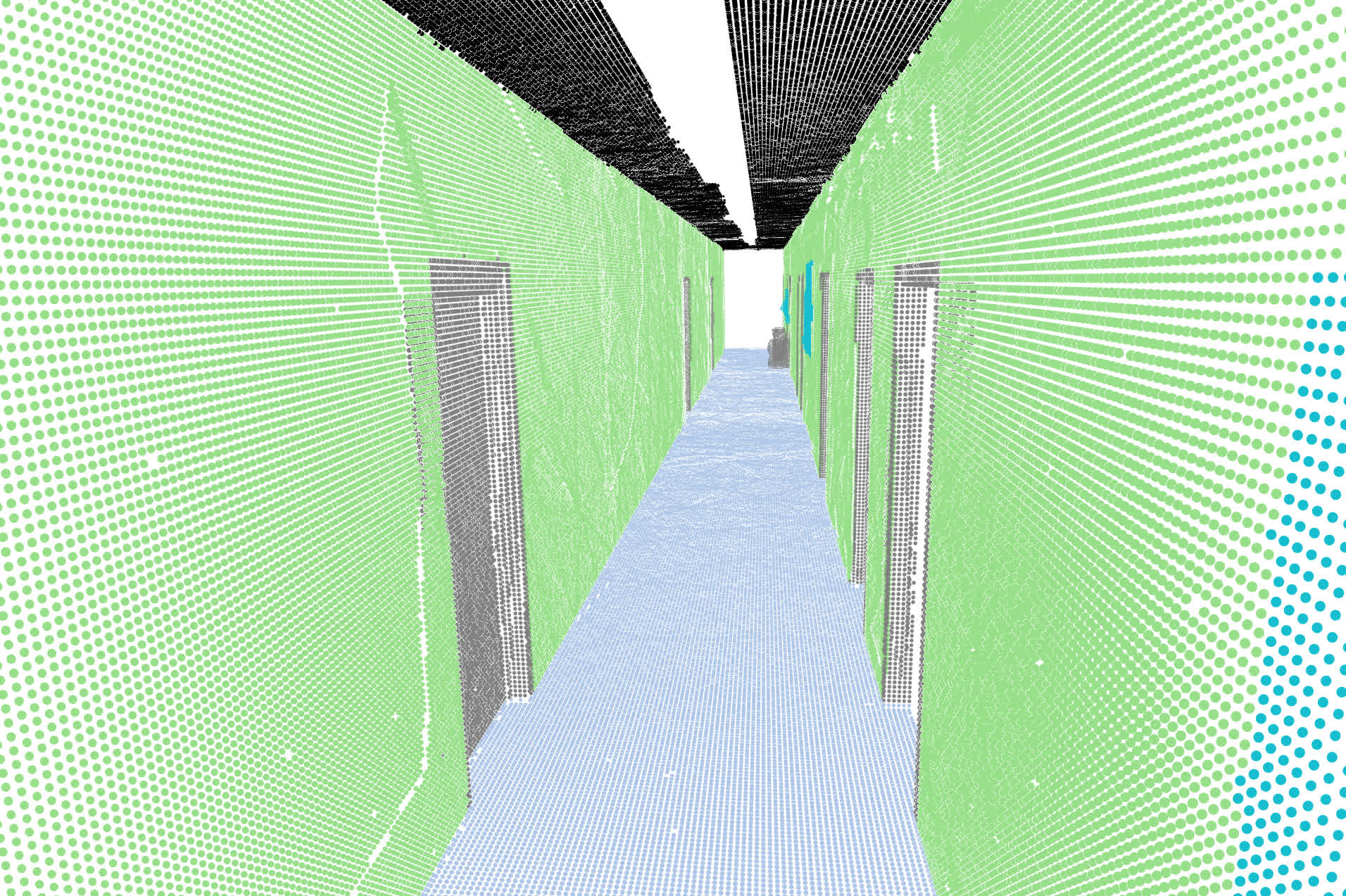}
        \end{overpic}
        \\
        \multicolumn{4}{c}{
        \begin{overpic}[width=0.99\textwidth]{_images_qualitative_s3dis_legend_s3dis.pdf}
        \end{overpic}}
    \end{tabular}
    \caption{Qualitative comparisons on SemanticPOSS (top), SemanticKITTI (centre) and S3DIS (bottom). 
    We report results on novel classes.
    EUMS$^\dagger$ fails in recognising novel objects with mixed and noisy predictions, e.g.~the \textit{car} class in KITTI-$5^1$ or the \textit{wall} class in S3DIS-$4^0$. 
    \ourmethod shows better segmentation performance on the novel classes, but it still misses a complete knowledge over the meaning of classes, e.g. it mixes \textit{trunk} and \textit{pole} classes in POSS-$3^2$ or \textit{ceiling} and \textit{floor} in S3DIS-$4^0$.
    \newmethod demonstrates superior performances on all three datasets, proving a better understanding of the scene.}
    \label{fig:qualitative}
\end{figure*}

\noindent
\textbf{Discussion.}
\newmethod consistently outperforms the compared baselines across most of the splits within the three datasets.
While the superiority of \newmethod over other NCD methods is empirically clear, understanding the underlying factors contributing to this improvement is essential.
Compared to EUMS$^\dag$, \newmethod achieves superior performance, thanks to online pseudo-labelling (shared with \ourmethod). This enables precise refinement and adaptation of the model's predictions and leads to better results. 
Compared to \ourmethod, \newmethod incorporates an alignment procedure that injects unsupervised semantic knowledge into our architecture. To assess the impact of the semantically-aligned branch, we compare the tSNE~\citep{van2008visualizing} dimensionality reduction of the embedding spaces of \ourmethod and \newmethod, as shown in Fig.~\ref{fig:t-SNE}.
We randomly selected eight point clouds from the validation set of each dataset and processed them through the feature extractors of both \ourmethod and \newmethod, resulting in point-wise features. From these, we retained $5000$ random novel points along with their respective features, applied t-SNE reduction, and visualized the points with colors corresponding to their ground-truth labels.
As illustrated in Fig.~\ref{fig:t-SNE}, \newmethod exhibits a more refined organization of the feature space, characterized by compact and well-separated class clusters.  This contributes to the observed performance enhancement between \newmethod and \ourmethod. 
\newmethod also significantly improves over the \textit{OpenScene$^\star$} baseline on two out of three datasets, highlighting that relying solely on the zero-shot capabilities of the auxiliary network is not enough for performance.
In contrast, \newmethod adeptly integrates the advantages of online pseudo-labelling and semantic alignment, showcasing its ability in making these components work in synergy for superior performance in point cloud semantic segmentation.\\
In our opinion, the performance of the auxiliary zero-shot network fluctuates significantly based on what we deem as class representation disparities between the distillation and testing datasets; it performs relatively well for well-represented classes but it encounters substantial challenges in accurately identifying rare classes during testing.

\noindent \textbf{Computational time.}
\newmethod shows a drastic reduction in the computational time when compared to EUMS$^\dag$.
Firstly, EUMS$^\dag$ requires a pre-training step and a fine-tuning step, i.e.~30 training epochs in total. 
Then, EUMS$^\dag$ requires a large amount of memory (up to 200 GB memory for KITTI-$5^0$) to store the data required for clustering, taking several hours (50 hrs) to complete the training procedure. 
Differently, \newmethod achieves superior performance with 10 training epochs, by using less memory (10 GB max) and a lower computational time (up to 25 hrs for KITTI-$5^0$).
We run these tests using one GPU Tesla A40-48GB.

\subsection{Qualitative analysis}
Fig.~\ref{fig:qualitative} depicts segmentation results obtained with \newmethod, \ourmethod and EUMS$^\dag$ across SemanticPOSS, SemaniticKITTI, and S3DIS datasets.
EUMS$^\dag$ faces substantial challenges when it comes to identifying novel objects within scenes, resulting in noisy and mixed labels for these categories. In KITTI-$5^1$, EUMS$^\dag$ inaccurately labels portions of \textit{car} objects as \textit{trunk}, and in S3DIS-$4^1$, parts of the \textit{wall} are mislabeled as \textit{table} and \textit{clutter}.
\ourmethod demonstrates improved semantic segmentation capabilities, giving in output more coherent labels and less noisy predictions. However, there are cases in which \ourmethod exhibits a limited understanding of the scenes. For instance, in POSS-$3^2$, \textit{trunk} is mixed with \textit{traffic-sign} and \textit{pole}. Moreover, in S3DIS-$4^0$, \ourmethod fails in differentiating between the \textit{ceiling} and the \textit{floor} class, likely due to their similar geometric structure.
In contrast, \newmethod shows enhanced segmentation and scene understanding capabilities. For example, it proficiently identifies \textit{traffic-sign} and \textit{pole} classes in POSS-$3^2$ and accurately segments \textit{fence} and \textit{car} in KITTI-$5^1$. Notably, \newmethod properly distinguishes the \textit{ceiling} from the \textit{floor} class in S3DIS-$4^0$, thanks to the semantical knowledge acquired through the semantic alignment procedure detailed in Sec.~\ref{sec:semantic_distillation}.

\section{Ablation studies}\label{sec:ablation}
We thoroughly evaluate \newmethod on SemanticPOSS, conducting an analysis of its core components and exploring how variations in its training parameters affect its performance. Namely, we analyse \newmethod behaviour when changing the value of the percentile $p$ and the semantic alignment loss weighting factor $\gamma$, to provide a comprehensive understanding of the efficacy of different part of our architecture.

We also evaluate \newmethod on S3DIS, analysing its stability across different runs, by conducting the same experiments multiple rounds and checking the mean and standard deviation of the obtained results.
Lastly, we assess the importance of acquisition sensor and distillation dataset in the comparison between \textit{OpenScene$^\star$} and \newmethod.

\begin{figure}[t]
\centering
    \begin{tabular}{@{}c@{}c}
        \begin{overpic}[width=0.48\columnwidth]{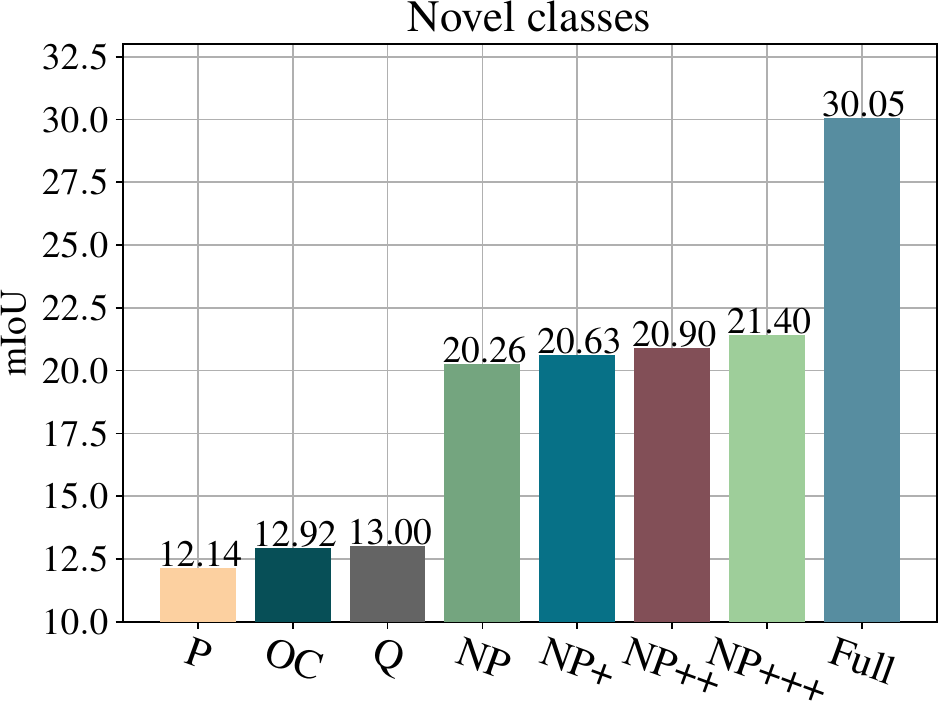}
        \end{overpic}& 
        \begin{overpic}[width=0.52\columnwidth]{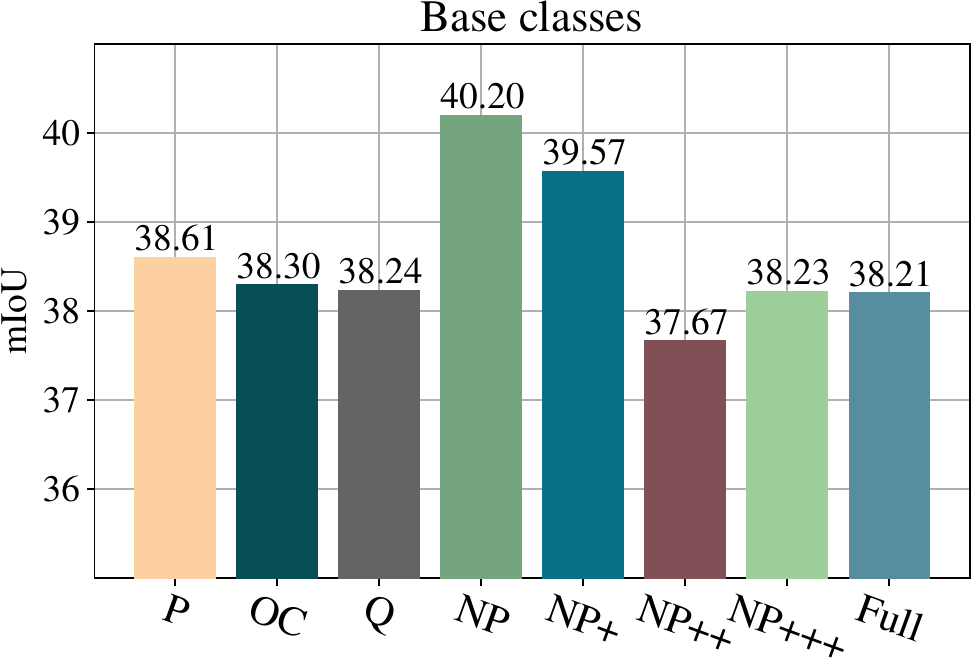}
        \end{overpic}
    \end{tabular}
    \vspace{-.3cm}
    \caption{Ablation study with different components and initialisation strategies on SemanticPOSS. 
    In $\mathsf{P}$, $\mathsf{OC}$ and $\mathsf{Q}$, we initialise the model after base pre-training, and use different configurations of the over-clustering heads and of our queue balancing. 
    In $\mathsf{NP}$ $\mathsf{NP}$+, $\mathsf{NP}$++ and $\mathsf{NP}$+++, we begin with $\mathsf{Q}$, we avoid pre-training, and we use $\phi$ and $\tau_c$ incrementally.
    In $\mathsf{Full}$, we add the semantic alignment head.
    See Sec.~\ref{sec:ablation} for definition of methods.}
    \label{fig:ablation_components}
\end{figure}
\noindent\textbf{Method components.}
Fig.~\ref{fig:ablation_components} shows the performance on novel and base classes of eight versions of \newmethod.
The first three versions use the pre-trained model on the base classes, while the last five versions use the model trained from scratch. 
Each version is defined as follows:
\begin{itemize}
    \item $\mathsf{P}$: we use a pre-trained model, and we remove $Z_q$, $\tau_c$ and the over-clustering heads.
    \item $\mathsf{OC}$: $\mathsf{P}$ + over-clustering heads.
    \item $\mathsf{Q}$: $\mathsf{OC}$ + $Z_q$, i.e.~our queue without uncertainty-aware filtering.
    \item $\mathsf{NP}$: $\mathsf{Q}$ without pre-training.
    \item $\mathsf{NP}$+: $\mathsf{NP}$ + our selection function $\phi$ on the queue.
    \item $\mathsf{NP}$++: $\mathsf{NP}$ + $\tau_c$ on the features used to derive the pseudo-labels.
    \item $\mathsf{NP}$+++: $\mathsf{NP}$ with $\tau_c$ and $\phi$, without the semantic alignment branch.
     \item $\mathsf{Full}$: \newmethod with all the components activated.
\end{itemize}
Pre-trained approaches generally underperform their trained-from-scratch counterparts on the novel classes.
This is visible in the low performance of $\mathsf{P}$, $\mathsf{OC}$ and $\mathsf{Q}$.
We have a significant improvement when pre-training is not used ($\mathsf{NP}$), i.e.~we achieve $20.26$ mIoU.
We can see that the queue both with and without pre-training is helpful.
When we add the feature selection for the queue and for the training, i.e.~$\mathsf{NP}$+ and $\mathsf{NP}$++, we have improvements, i.e.~$20.63$ mIoU and $20.90$ mIoU, respectively.
With $\mathsf{NP}$+++ we observe a further increase in performance, reaching $21.40$ mIoU.
The best performance is achieved with $\mathsf{Full}$, with an mIoU of $30.05$.
Although we can observe variations on the performance of the base classes, their information is retained by the network when we discover the novel categories.

\begin{table}[t]
    \centering
    \caption{Ablation study showing how different values of $p$ affect the performance on SemanticPOSS. 
    The lower $p$ is, the less severe the selection of the features, resulting in better performance for POSS-$4^0$. Differently, POSS-$3^3$ benefits from an higher value of $p$, which leads to a more vigorous filtering of the features. POSS-$3^1$ and POSS-$3^2$ show the best performance with $p=0.5$.}
    \vspace{-.2cm}
    \label{tab:components_ablations}
    \begin{tabular}{lccccc}
        \toprule
        \multirow{2}{*}{Split} & \multicolumn{5}{c}{Percentile $p$}\\
        & 0.1 & 0.3 & 0.5 & 0.7 & 0.9 \\
        \midrule
        POSS-$4^0$ & 30.81 & \textbf{35.70} & 28.77 & 30.93 & 26.69 \\
        POSS-$3^1$ & 28.33 & 30.02 & \textbf{30.43} & 23.32 & 18.91 \\
        POSS-$3^2$ & 8.07 & 8.95 & \textbf{10.32} & 10.25 & 7.76 \\
        POSS-$3^3$ & 10.55 & 10.94 & 11.69 & \textbf{14.38} & 13.42 \\
        \midrule
        Avg. & 19.44 & \textbf{21.40} & 20.30 & 19.72 & 16.70 \\
        \bottomrule
    \end{tabular}
\end{table}
\noindent \textbf{Percentile analysis.}
We study the behaviour of the percentile $p$ in our selection function $\phi$, when we apply it to the features both for pseudo-labelling and for the class-balanced queue $\mathtt{Z}_q$.
Tab.~\ref{tab:components_ablations} reports the results on each split of SemanticPOSS.
For each split, we can observe that the performance depends on the number of points and difficulty of the novel classes. In POSS$-4^0$ and POSS-$3^1$, lower values of $p$ result in less severe selection. 
We believe this is related to the class distribution within these splits.
This is in line with what observed in Tab.~\ref{tab:results_poss}.
In POSS$-3^2$ and POSS-$3^3$, we notice a different behaviour, a higher value of $p$ provides better results.
We relate this to the difficulty of the novel classes in these splits whose noisy pseudo-labels can benefit from a more rigorous selection of the features.

\begin{table*}[ht]
    \centering
    \tabcolsep 12pt
    \caption{Ablation study reporting mean and standard deviation obtained by running the same experiment $N=10$ times. We report results on novel and base classes, together with the performance on all classes. $\min_\sigma$ and $\max_\sigma$ represent the lowest and the highest standard deviation showcased for the different groups of classes, respectively.}
    \vspace{-.2cm}
    \label{tab:stability}
    \resizebox{\textwidth}{!}{%
    \begin{tabular}{l|ccc|ccc|ccc|ccc}
        \toprule
        & \multicolumn{3}{c|}{S3DIS-$4^0$} & \multicolumn{3}{c|}{S3DIS-$3^1$} & \multicolumn{3}{c|}{S3DIS-$3^2$} & \multicolumn{3}{c}{S3DIS-$3^3$}\\
        & novel & base & all & novel & base & all & novel & base & all & novel & base & all \\
        \midrule
        $\mu$ & 55.71 & 30.82 & 38.48 & 52.28 & 41.48 & 43.97 & 18.27 & 50.92 & 43.38 & 10.90 & 54.80 & 44.67 \\
        $\sigma$ & 0.53 & 0.76 & 0.53 & 0.95 & 0.42 & 0.51 & 1.29 & 0.48 & 0.57 & 1.08 & 0.39 & 0.45 \\
        \midrule
        $\min_\sigma$ & 0.14 & 0.03 & 0.03 & 0.95 & 0.04 & 0.04 & 0.20 & 0.29 & 0.20 & 0.62 & 0.12 & 0.12 \\
        $\max_\sigma$ & 2.24 & 2.47 & 2.47 & 1.82 & 3.89 & 3.89 & 3.65 & 3.10 & 3.65 & 3.96 & 2.37 & 3.96 \\
        \bottomrule
    \end{tabular}
    }
\end{table*}

\noindent \textbf{Loss weighting analysis.}
\begin{figure}[t]
    \centering
    \includegraphics[width=0.95\linewidth]{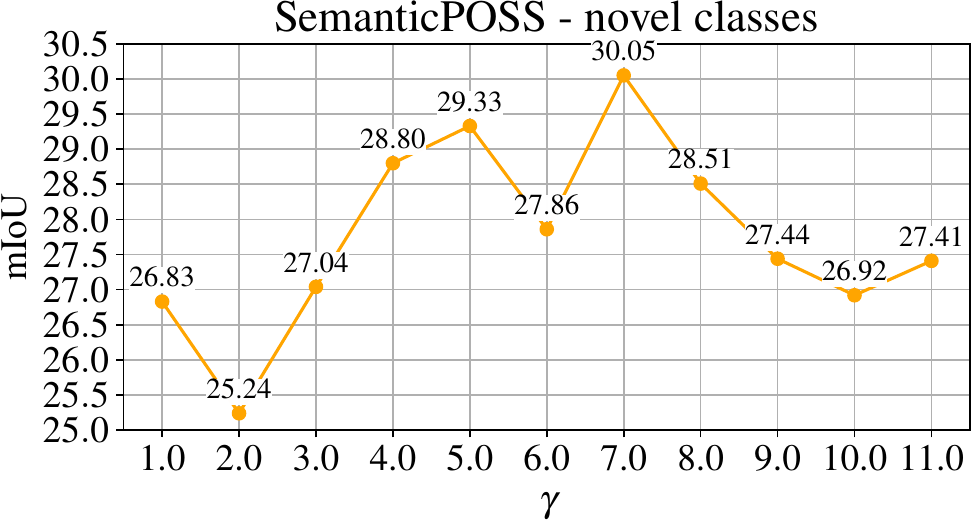}
    \caption{Ablation study analysing the difference in performance when changing the value of $\gamma$, the weighting factor of $\ell_\text{A}$. Reported performance are in terms of average mIoU of novel classes in the four splits of SemanticPOSS.}
    \label{fig:gamma_ablation}
\end{figure}
We examine the behavior of \newmethod when adjusting the value of $\gamma$, the weighting factor assigned to the alignment loss $\ell_\text{A}$. The results shown in Fig.~\ref{fig:gamma_ablation} depict the average performance across the four SemanticPOSS splits for novel classes as we vary $\gamma$. We observe that increasing the value of $\gamma$ corresponds to higher mIoU values for novel classes. The optimal performance is achieved when $\gamma$ is set to $7.0$. However, further increasing the value of $\gamma$ leads to a decrease in mIoU values. We attribute this trend to an imbalance between the alignment loss $\ell_\text{A}$ and the segmentation loss $\ell_\text{S}$. By assigning excessive weight to $\ell_\text{A}$, there is an indirect reduction in the significance of $\ell_\text{S}$, which in turn decreases the network's ability to converge to a good solution.

\noindent \textbf{\newmethod's stability.}
We assess the stability in \newmethod optimisation by running the same experiment $N=10$ times. 
Tab.~\ref{tab:stability} presents mean $\mu$ and standard deviation $\sigma$ for all the splits in S3DIS. 
The results show that \newmethod is generally stable across different runs of the same experiment, with an average standard deviation ($\sigma$) of $0.52$ across the four splits. 
In three out of four splits, the novel classes exhibit higher standard deviations compared to the base classes. 
This difference is likely attributed to the fact that the first group of (novel) classes is learned solely with the supervision of pseudo-labels and distillation, whereas the base classes benefit from the availability of labelled data.

\noindent \textbf{Distillation data.}
In Sec.~\ref{sec:experiments}, the \textit{OpenScene$^\star$} baseline is obtained by testing the OpenScene model on data that differs from the one seen during distillation: we use the ScanNet-OpenSeg model for S3DIS and nuScenes-OpenSeg for SemanticKITTI and SemanticPOSS. In Tabs.~\ref{tab:results_s3dis_distillation} \& \ref{tab:results_nuscenes} we report results obtained by testing the \textit{OpenScene$^\star$} baseline on data which is similar to the one seen during training.

Tab.~\ref{tab:results_s3dis_distillation} reports the results obtained on S3DIS using the Matterport-OpenSeg OpenScene model. The testing dataset (S3DIS) shares the same acquisition sensor as the distillation data (Matterport~\citep{chang2017matterport3d}) for the Matterport-OpenSeg model, i.e. the Matterport360 camera. This should reduce the domain gap between training and testing data, resulting in better results for \textit{OpenScene$^\star$}. Interestingly, on \textit{OpenScene$^\star$}, the use of the Matterport-OpenSeg model produces very similar results as ScanNet-OpenSeg, with $36.67$ average IoU in the first case and $36.76$ in the second case. However, when applied to NCD, the use of Matterport-OpenSeg results in worse average results on novel classes, dropping from $34.05$ IoU to $32.91$.

\begin{table*}[t]
    \centering
    \caption{Novel Class discovery results on S3DIS, with two different OpenScene models.
    OpenScene$^\star$: reference described in Sec.~\ref{sec:zero-shot} (``\textit{n} Syn'' indicates the number \textit{n} of synonyms used to build the ensembles, (\texttt{dataset}) indicates the distillation dataset). \newmethod (\texttt{dataset}) is the model with distillation from the OpenScene model trained on \texttt{dataset}. Highlighted values are the novel classes in each split.}
    \vspace{-.2cm}
    \label{tab:results_s3dis_distillation}
    \tabcolsep 6pt
    \resizebox{\textwidth}{!}{%
    \begin{tabular}{l|l|ccccccccccccc|ccc}
        \toprule
        \multirow{2}{*}{\textbf{Split}} & \multirow{2}{*}{\textbf{Model}} & \multirow{2}{*}{\rotatebox{45}{\textbf{beam}}} & \multirow{2}{*}{\rotatebox{45}{\textbf{board}}} & \multirow{2}{*}{\rotatebox{45}{\textbf{book.}}} & \multirow{2}{*}{\rotatebox{45}{\textbf{ceil.}}} & \multirow{2}{*}{\rotatebox{45}{\textbf{chair}}} & \multirow{2}{*}{\rotatebox{45}{\textbf{clutt.}}} & \multirow{2}{*}{\rotatebox{45}{\textbf{col.}}} & \multirow{2}{*}{\rotatebox{45}{\textbf{door}}} & \multirow{2}{*}{\rotatebox{45}{\textbf{floor}}} & \multirow{2}{*}{\rotatebox{45}{\textbf{sofa}}} & \multirow{2}{*}{\rotatebox{45}{\textbf{table}}} & \multirow{2}{*}{\rotatebox{45}{\textbf{wall}}} & \multirow{2}{*}{\rotatebox{45}{\textbf{wind.}}} &  \multicolumn{3}{c}{\textbf{mIoU}} \\
         &  &  &  &  &  &  &  &  &  &  &  &  &  &  & \textbf{Novel} & \textbf{Base} & \textbf{All}\\
        \midrule
        
         & OpenScene$^\star$ 1 Syn. (ScanNet)& 0.00 & 0.00 & 42.36 & 72.78 & 56.25 & 10.81 & 0.00 & 47.17 & 85.53 & 45.48 & 42.31 & 59.24 & 15.95 & - & - & 36.76 \\
         & OpenScene$^\star$ 1 Syn. (Matterport)& 0.00 & 0.00 & 41.60 & 72.78 & 56.45 & 10.82 & 0.00 & 47.27 & 85.38 & 45.18 & 42.38 & 59.08 & 15.76 & - & - & 36.67 \\
        \midrule
        
        \multirow{2}{*}{S3DIS-$4^0$} & \newmethod (ScanNet) (Ours) & 0.55 & 0.12 & 49.86 & \CC{novelcolor}81.01 & 72.82 & \CC{novelcolor}\textbf{9.99} & 28.48 & 35.48 & \CC{novelcolor}\textbf{94.39} & 43.50 & 64.37 & \CC{novelcolor}\textbf{38.37} & 2.91 & \CC{novelcolor}\textbf{55.94} & 33.12 & 40.14 \\
         & \newmethod (Matterport) (Ours) & 0.68 & 11.51 & 52.29 & \CC{novelcolor}\textbf{82.69} & 73.16 & \CC{novelcolor}4.56 & 23.36 & 24.62 & \CC{novelcolor}92.73 & 48.70 & 60.19 & \CC{novelcolor}36.55 & 7.59 & \CC{novelcolor}54.05 & 33.57 & 39.87 \\
        \midrule
        
        \multirow{2}{*}{S3DIS-$3^1$} & \newmethod (ScanNet) (Ours) & 0.00 & 8.81 & 53.78 & 81.77 & \CC{novelcolor}58.14 & 36.61 & 27.62 & \CC{novelcolor}\textbf{42.43} & 94.29 & 57.45 & \CC{novelcolor}\textbf{59.90} & 63.52 & 10.98 & \CC{novelcolor}\textbf{53.49} & 43.49 & 45.79 \\
         & \newmethod (Matterport) (Ours) & 0.00 & 10.85 & 54.99 & 83.95 & \CC{novelcolor}\textbf{60.51} & 35.29 & 29.50 & \CC{novelcolor}39.00 & 93.36 & 43.83 & \CC{novelcolor}56.77 & 62.06 & 13.93 & \CC{novelcolor}52.09 & 42.78 & 44.93 \\
        \midrule
        
        \multirow{2}{*}{S3DIS-$3^2$} & \newmethod (ScanNet) (Ours) & \CC{novelcolor}\textbf{0.93} & 6.29 & \CC{novelcolor}33.34 & 79.59 & 76.69 & 36.74 & \CC{novelcolor}\textbf{12.26} & 32.57 & 95.91 & 46.14 & 67.65 & 62.92 & 5.37 & \CC{novelcolor}15.51 & 50.99 & 42.80 \\
         & \newmethod (Matterport) (Ours) & \CC{novelcolor}0.00 & 7.32 & \CC{novelcolor}\textbf{47.34} & 71.68 & 79.25 & 32.68 & \CC{novelcolor}8.63 & 31.69 & 94.91 & 46.96 & 67.45 & 61.77 & 10.68 & \CC{novelcolor}\textbf{18.65} & 50.44 & 43.10 \\
        \midrule
        
        \multirow{2}{*}{S3DIS-$3^3$} & \newmethod (ScanNet) (Ours) & 0.00 & \CC{novelcolor}\textbf{7.26} & 56.55 & 82.90 & 76.56 & 36.82 & 25.87 & 44.71 & 96.38 & \CC{novelcolor}\textbf{20.23} & 65.61 & 66.91 & \CC{novelcolor}6.24 & \CC{novelcolor}\textbf{11.24} & 55.23 & 45.08 \\
         & \newmethod (Matterport) (Ours) & 0.00 & \CC{novelcolor}1.95 & 56.07 & 81.21 & 75.92 & 36.25 & 22.36 & 44.08 & 95.30 & \CC{novelcolor}11.76 & 68.30 & 66.40 & \CC{novelcolor}\textbf{6.80} & \CC{novelcolor}6.84 & 54.59 & 43.57 \\

        \bottomrule
        \addlinespace[2.5pt]
         \multicolumn{10}{c}{} & \multirow{2}{*}{Avg} & \multicolumn{4}{|l|}{\newmethod (ScanNet) (Ours)} & \CC{novelcolor}\textbf{34.05} & 45.71 & 43.45\\
         \multicolumn{10}{c}{} &  & \multicolumn{4}{|l|}{\newmethod (Matterport) (Ours)} & \CC{novelcolor}32.91 & 45.34 & 42.87\\
         \cmidrule[1pt]{11-18}
         \cmidrule[1pt]{11-18}
    \end{tabular}
    }
\end{table*}
\begin{table*}[th!]
    \centering
    \caption{Novel class discovery results on nuScenes. 
    OpenScene$^\star$: reference described in Sec.~\ref{sec:zero-shot} (``\textit{n} Syn'' indicates the number \textit{n} of synonyms used to build the ensembles). Highlighted values are the novel classes in each split.}
    \vspace{-.2cm}
    \label{tab:results_nuscenes}
    \tabcolsep 6pt
    \resizebox{\textwidth}{!}{%
    \begin{tabular}{l|l|cccccccccccccccc|ccc}
        \toprule
        \multirow{2}{*}{\textbf{Split}} & \multirow{2}{*}{\textbf{Model}} & \multirow{2}{*}{\rotatebox{45}{\textbf{barr.}}} & \multirow{2}{*}{\rotatebox{45}{\textbf{bi.cle}}} & \multirow{2}{*}{\rotatebox{45}{\textbf{bus}}} & \multirow{2}{*}{\rotatebox{45}{\textbf{car}}} & \multirow{2}{*}{\rotatebox{45}{\textbf{const.}}} & \multirow{2}{*}{\rotatebox{45}{\textbf{driv.s.}}} & \multirow{2}{*}{\rotatebox{45}{\textbf{manm.}}} & \multirow{2}{*}{\rotatebox{45}{\textbf{mt.cle}}} & \multirow{2}{*}{\rotatebox{45}{\textbf{oth-g.}}} & \multirow{2}{*}{\rotatebox{45}{\textbf{pede.}}} & \multirow{2}{*}{\rotatebox{45}{\textbf{sidew.}}} & \multirow{2}{*}{\rotatebox{45}{\textbf{terr.}}} & \multirow{2}{*}{\rotatebox{45}{\textbf{tr. c.}}} & \multirow{2}{*}{\rotatebox{45}{\textbf{trail.}}} & \multirow{2}{*}{\rotatebox{45}{\textbf{truck}}} & \multirow{2}{*}{\rotatebox{45}{\textbf{veget.}}}&  \multicolumn{3}{c}{\textbf{mIoU}} \\
         &  &  &  &  &  &  &  &  &  &  &  &  &  &  &  &  &  & \textbf{Novel} & \textbf{Base} & \textbf{All}\\
        \midrule

         &  OpenScene$^\star$ 1 Syn. & 9.92 & 0.00 & 41.98 & 68.61 & 17.11 & 79.29 & 31.92 & 20.48 & 0.03 & 55.71 & 22.28 & 0.00 & 11.62 & 7.73 & 46.25 & 44.92 & - & - & 28.68 \\
         &  OpenScene$^\star$ 3 Syn. & 9.92 & 0.00 & 41.90 & 62.20 & 16.85 & 74.97 & 55.00 & 20.27 & 0.06 & 42.34 & 32.06 & 11.96 & 9.86 & 13.09 & 41.16 & 77.31 & - & - & 31.81 \\
         &  OpenScene$^\star$ 5 Syn. & 11.50 & 0.22 & 40.98 & 55.98 & 17.49 & 76.58 & 54.84 & 21.21 & 0.13 & 36.57 & 29.55 & 29.27 & 5.13 & 14.06 & 32.10 & 74.41 & - & - & 31.25 \\
        \midrule
        
        nuScenes-$4^0$ & \newmethod (Ours) & 31.87 & 5.84 & 36.51 & 73.58 & 8.81 & \CC{novelcolor}30.08 & \CC{novelcolor}47.82 & 13.86 & 28.30 & 21.24 & 46.18 & \CC{novelcolor}14.47 & 7.74 & 17.50 & 49.94 & \CC{novelcolor}69.11 & \CC{novelcolor}40.37 & 28.45 & 31.43 \\
        \midrule
        
        nuScenes-$4^1$ & \newmethod (Ours) & \CC{novelcolor}22.62 & 9.15 & 43.80 & \CC{novelcolor}45.85 & 10.25 & 84.68 & 82.40 & 14.61 & 32.53 & 29.50 & \CC{novelcolor}38.88 & 53.81 & 10.26 & 22.29 & \CC{novelcolor}0.09 & 82.77 & \CC{novelcolor}26.86 & 39.67 & 36.47  \\
        \midrule
        
        nuScenes-$4^2$ & \newmethod (Ours) & 40.42 & 7.33 & \CC{novelcolor}15.71 & 79.03 & 10.80 & 84.43 & 80.72 & 15.28 & \CC{novelcolor}16.42 & \CC{novelcolor}13.62 & 52.27 & 51.68 & 9.40 & \CC{novelcolor}0.11 & 49.60 & 82.30 & \CC{novelcolor}11.46 & 46.94 & 38.07 \\
        \midrule
        
        nuScenes-$4^3$ & \newmethod (Ours) & 39.00 & \CC{novelcolor}1.31 & 41.76 & 78.93 & \CC{novelcolor}4.24 & 87.74 & 80.03 & \CC{novelcolor}4.65 & 34.94 & 29.51 & 53.31 & 53.98 & \CC{novelcolor}4.86 & 21.65 & 48.20 & 81.42 & \CC{novelcolor}3.76 & 54.21 & 41.60 \\

        \bottomrule
        \addlinespace[2.5pt]
         \multicolumn{14}{c}{} & Avg & \multicolumn{3}{|l|}{\newmethod (Ours)} & \CC{novelcolor}20.61 & 42.32 & 36.89 \\
         \cmidrule[1pt]{15-21}
    \end{tabular}
    }
\end{table*}

Tab.~\ref{tab:results_nuscenes} presents a comparative analysis between the baseline method \textit{OpenScene$^\star$} and \newmethod when applied to the nuScenes dataset \citep{caesar2020nuscenes, fong2022panopticnuscenes}. As detailed in Sec.~\ref{sec:experiments_sub}, we outline in Tab.~\ref{tab:nuScenes_folds} the dataset division into splits for the NCD setting. Notably, \textit{OpenScene$^\star$} achieves $31.81$ IoU in its optimal configuration, surpassing the average IoU performance of \newmethod across the four splits on novel classes ($20.61$). \newmethod only exhibits a superior performance in three specific classes: \textit{barrier}, \textit{bicycle}, and \textit{other-ground}. This experiment underscores that when both distillation and testing are conducted on the same dataset, \textit{OpenScene$^\star$} demonstrates a very good performance. However, employing distillation on one dataset and evaluating it on another dataset with certain domain gap (as in all our previous experiments) results in a poorer performance. This shows that the distilled open-vocabulary knowledge has limited generalisation capability when tested cross-dataset.

\begin{table}[t]
    \centering
    \caption{nuScenes splits, defined as nuScenes-$n^i$, where $n$ is the number of novel classes and $i$ is the split index.}
    \label{tab:nuScenes_folds}
    \vspace{-.2cm}
    \begin{tabular}{ll}
        \toprule
        Split & Novel Classes  \\
        \midrule
        nuScenes-$4^0$ & \textit{driveable s.}, \textit{manmade}, \textit{terrain}, \textit{veget.} \\
        nuScenes-$4^1$ & \textit{barrier}, \textit{car}, \textit{sidewalk}, \textit{truck} \\
        nuScenes-$4^2$ & \textit{bus}, \textit{other g.}, \textit{pedestrian}, \textit{trailer} \\
        nuScenes-$4^3$ & \textit{bicycle}, \textit{constr. v.}, \textit{motorc.}, \textit{traffic c.} \\
        \bottomrule
    \end{tabular}
\end{table}

\section{Conclusions}\label{sec:conclusions}

We explored the new problem of novel class discovery for 3D point cloud segmentation.
Firstly, we adapted the only NCD method for 2D image semantic segmentation to 3D point cloud data, and experimentally found that it has several limitations.
We discussed that extending 2D NCD approaches to 3D data (point clouds) is not trivial because the assumptions made for 2D data are not easily transferable to 3D.
Secondly, we presented \newmethod, an extension of our original \ourmethod method, that tackles NCD for point cloud segmentation by using online clustering, uncertainty quantification and semantic distillation through a foundation model. 
We showed that the zero-shot accuracy of such foundation model alone is not satisfactory and we proved that by using it in combination with our \newmethod we can achieve higher performance.
Lastly, we introduced a novel evaluation protocol to asses the performance of NCD in point cloud segmentation.
Experiments on three different segmentation dataset showed that \newmethod outperforms the compared baselines by a large margin.

\noindent \textbf{Limitations}
The first limitation of \newmethod is the prior knowledge on the number of novel classes $C_n$ to discover.
This could be a limitation when $C_n$ is not a known prior and novel classes appear in an incremental manner.
We believe that a solution may be to learn novel classes incrementally, as for example proposed by~\cite{roy2022class} in the 2D Novel Class Discovery literature.
Finally, \newmethod lacks a mechanism to prevent drift when the auxiliary network outputs inaccurate features for novel classes.
\newmethod may benefit the introduction of a filtering mechanism to avoid point features when the auxiliary network exhibits high uncertainty, as for example proposed by~\cite{saltori2022gipso}.

\bibliography{egbib}

\end{document}